\documentclass[10pt,twocolumn,letterpaper]{article}

\usepackage{iccv}
\usepackage{times}
\usepackage{epsfig}
\usepackage{graphicx}
\usepackage{amsmath}
\usepackage{amssymb}
\usepackage{xcolor}
\usepackage{capt-of}
\usepackage{multirow}
\usepackage{amsmath}
\usepackage[accsupp]{axessibility}
\usepackage[toc,page]{appendix}
\usepackage{soul}
\DeclareMathOperator*{\argmax}{arg\,max}

\def\eg{\emph{e.g}\onedot} 
\def\ie{\emph{i.e}\onedot}

\definecolor{ao}{rgb}{0.0, 0.42, 0.24}
\definecolor{amethyst}{rgb}{0.6, 0.4, 0.8}

\usepackage[pagebackref=true,breaklinks=true,letterpaper=true,colorlinks,bookmarks=false]{hyperref}

\iccvfinalcopy %

\def\iccvPaperID{4145} %

\ificcvfinal\fi

\newif\ifdraft
\draftfalse
\ifdraft

\newcommand{\dcc}[1]{{\color{red}[\textbf{DC:} #1]}}
\newcommand{\opc}[1]{{\color{blue}[\textbf{OP:} #1]}}
\newcommand{\hec}[1]{{\color{teal}[\textbf{HE:} #1]}}
\newcommand{\dgc}[1]{{\color{violet}[\textbf{DG:} #1]}}

\else
\newcommand{\dcc}[1]{}
\newcommand{\opc}[1]{}
\newcommand{\hec}[1]{}
\newcommand{\dgc}[1]{}
\newcommand{\iec}[1]{}

\fi

\begin{document}

\title{Localizing Object-level Shape Variations with Text-to-Image Diffusion Models}

\author{
        \vspace{8pt}
        Or Patashnik$^1$ 
        \hspace{6mm}  Daniel Garibi$^1$ 
        \hspace{6mm}  Idan Azuri$^2$ 
        \hspace{6mm}  Hadar Averbuch-Elor$^1$ 
        \hspace{6mm}  Daniel Cohen-Or$^1$ \\ \vspace{6pt}
$^1$Tel-Aviv University \hspace{12mm} $^2$Independent Researcher
\\ 
\small{\url{https://orpatashnik.github.io/local-prompt-mixing/}}
}

\maketitle
\ificcvfinal\thispagestyle{empty}\fi

\begin{abstract}
\vspace{-10pt}

Text-to-image models give rise to workflows which often begin with an exploration step, where users sift through a large collection of generated images. The global nature of the text-to-image generation process prevents users from narrowing their exploration to a particular object in the image. In this paper, we present a technique to generate a collection of images that depicts variations in the shape of a specific object, enabling an object-level shape exploration process. Creating plausible variations is challenging as it requires control over the shape of the generated object while respecting its semantics. A particular challenge when generating object variations is accurately localizing the manipulation applied over the object's shape. We introduce a prompt-mixing technique that switches between prompts along the denoising process to attain a variety of shape choices. To localize the image-space operation, we present two techniques that use the self-attention layers in conjunction with the cross-attention layers. Moreover, we show that these localization techniques are general and effective beyond the scope of generating object variations. Extensive results and comparisons demonstrate the effectiveness of our method in generating object variations, and the competence of our localization techniques. 

\vspace{-18pt}
\end{abstract}

\begin{figure}[t]
    \centering
    {\normalsize
    \includegraphics[width=1.0\linewidth]{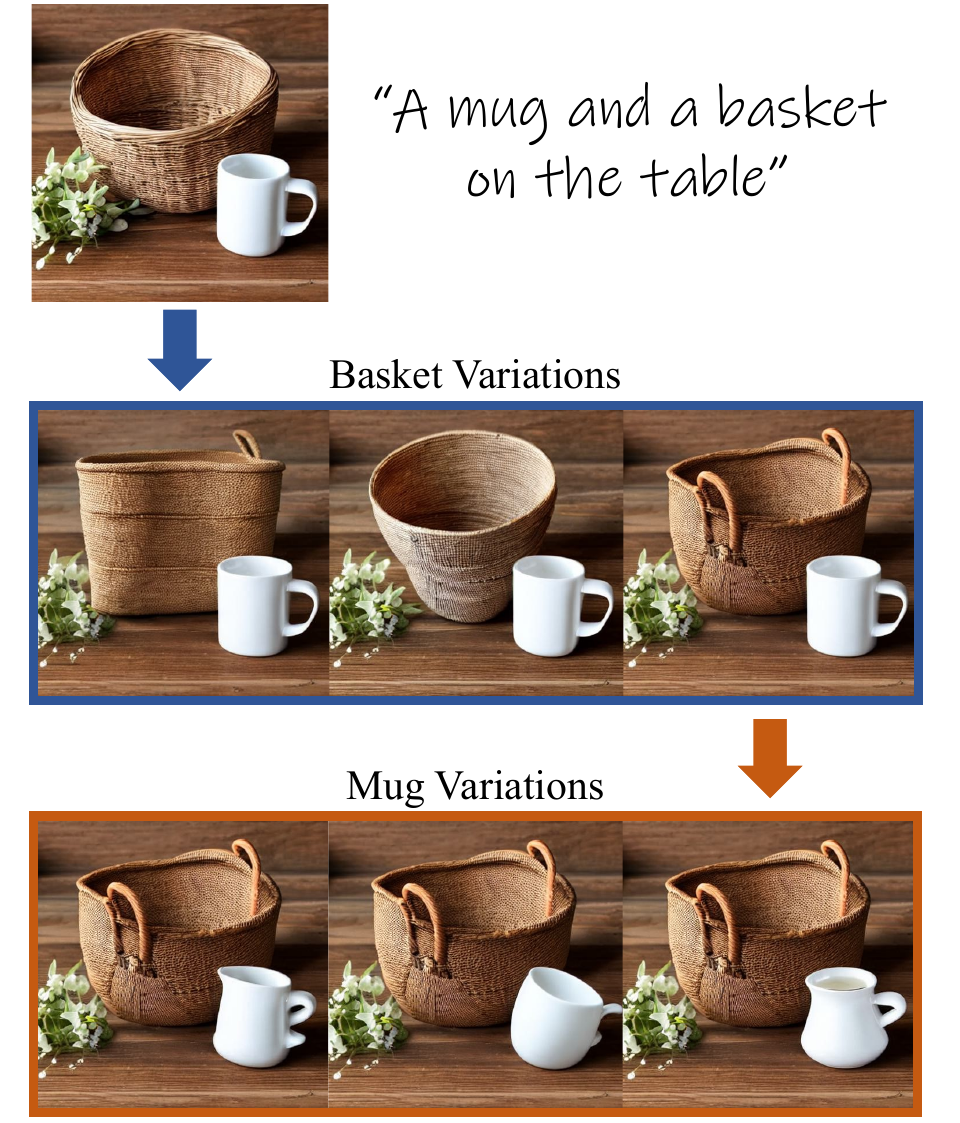}
    \caption{Our method generates shape variations of an object. Here, given the text prompt, we generate variations of the basket. After selecting a preferred basket, we generate variations of the mug. We develop general techniques for localizing modifications.}
    \vspace{-10pt}
    \label{fig:teaser}}
\end{figure}

\section{Introduction}

Text-to-image diffusion models have recently shown unprecedented image quality and diversity \cite{Ramesh2022HierarchicalTI, Saharia2022PhotorealisticTD, Rombach2021HighResolutionIS}, and have opened a new era in image synthesis. 
Still, the control over the generated image is limited, resulting in a tedious selection procedure, where users 
sample numerous initial seed noises, 
from which they can select a preferred one. 
Images generated from different initial noise with the same text prompt share semantics, but the shape, appearance, and location of the generated shapes may differ greatly.
The uncontrolled global changes realized by such a sampling process, do not allow users to interact with the generated image and narrow down their open-ended exploration process. In particular, the lack of object-level control of the user with the generated image hinders the user's ability to focus on refining specific objects during their exploration.

\begin{figure}
       \centering
        \setlength{\tabcolsep}{0.0pt}
        {\scriptsize
        \begin{tabular}{ccc cccc}
            \multicolumn{7}{c}{"A \emph{basket} with bananas"} \\
            \raisebox{6pt}{\rotatebox{90}{ Inpainting~\cite{Rombach2021HighResolutionIS}}} &
            { } &
            \includegraphics[width=0.235\linewidth]{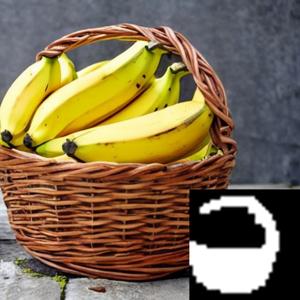} &
            {  } &
            \includegraphics[width=0.235\linewidth]{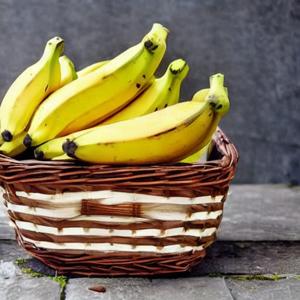} &
            \includegraphics[width=0.235\linewidth]{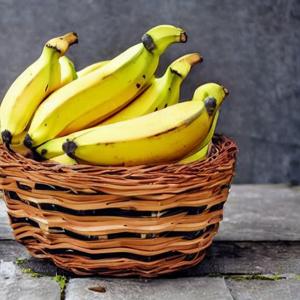} &
            \includegraphics[width=0.235\linewidth]{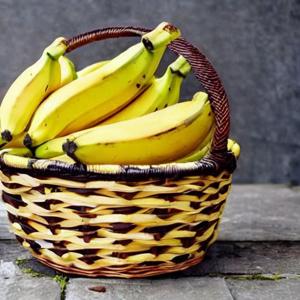} \\ 
            \raisebox{10pt}{\rotatebox{90}{ SDEdit~\cite{meng2022sdedit}}} &
            { } &
            \includegraphics[width=0.235\linewidth]{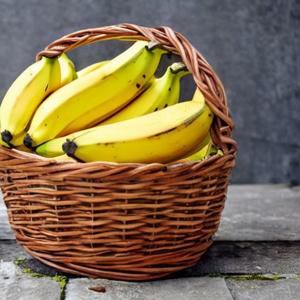} &
            {  } &
            \includegraphics[width=0.235\linewidth]{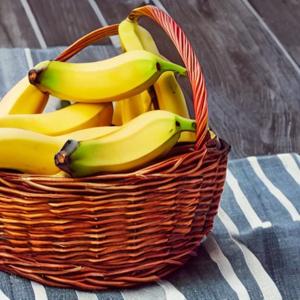} &
            \includegraphics[width=0.235\linewidth]{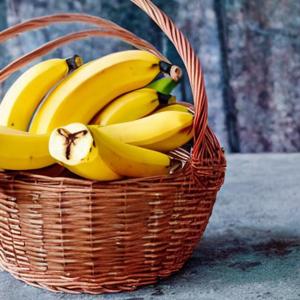} &
            \includegraphics[width=0.235\linewidth]{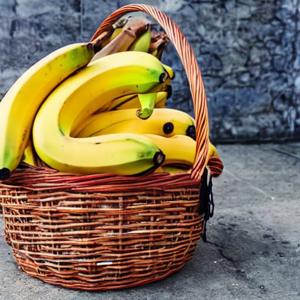} \\
            && Input & \multicolumn{4}{c}{different seeds samples}
        \end{tabular}
        }
    \vspace{1mm}
    \captionof{figure}{Inpainting and SDEdit struggle to achieve significant shape variations of the basket. Inpainting is restricted to the mask while SDEdit struggles at localizing changes, see the bananas.}
    \vspace{-12pt}
    \label{fig:inpainting}
\end{figure}

A possible approach for interacting with the generated image as a means to explore the shape and appearance of a specific object in the image is to use text-guided image inpainting~\cite{Rombach2021HighResolutionIS} or SDEdit~\cite{meng2022sdedit}. These methods mainly excel in changing the texture of an object and are therefore more suitable for textural exploration.
However, changing the shape of an object, particularly if other regions in the image should be preserved, is significantly more challenging, and these methods struggle to achieve that without affecting the entire image.
Figure~\ref{fig:inpainting} demonstrates the lack of shape variations with the above approaches.

In this paper, we deal with object-level shape exploration, where the user attains a gallery with variations of a specific object in the image, without having to provide any additional input. Specifically, acknowledging that previous works have struggled to perform geometric manipulations,
we focus on object-level shape variations, which are automatically generated and presented to the user, see Figure~\ref{fig:teaser}.

We introduce a prompt-mixing technique, where a mix of different prompts is used along the denoising process. This approach is built on the premise that
the denoising process is an innate coarse-to-fine synthesis,
which roughly consists of three stages. In the first stage, the general configuration or layout is drafted. In the second stage, the shapes of the objects are formed. Finally, in the third stage, their fine visual details are generated.

Common in text-based image editing techniques, a challenge arises when localizing the modification of the object.
Hence, we develop two novel means to localize edits.
First, to preserve the shapes of other objects, we inject a localized self-attention map from the original image into the newly generated image.
This injection leads to a rough alignment between the two images. Next, to further preserve appearance (\eg, image background),
we automatically extract labeled segmentation maps of both the original and generated images. 
The segmentation maps allow applying the edits locally in selected segments only. Finally, during the last denoising steps, we seamlessly blend all segments together.

We demonstrate that without any costly optimization process, our method offers the user the ability to generate object-level shape variations, while remaining faithful to the original image, either generated or real. Moreover, we show that our localization techniques are beneficial to not only object shape variations, but also to generic local image editing methods. We demonstrate the improved results achieved by integrating our localization techniques into existing image editing methods.
Extensive experiments are conducted to show that our approach can create more diverse results with larger shape changes and better content preservation compared to alternative methods.

\section{Related Work}

\subsection{Text-Guided Image Generation}
Text-to-image synthesis is a longstanding problem in computer vision and computer graphics. Early works were GAN-based~\cite{Xu2017AttnGANFT, Reed2016LearningWA, reed2016learning, Zhang2016StackGANTT} and were trained on small-scale datasets, typically of a single class. Recently, with the rapid progress in diffusion models~\cite{SohlDickstein2015DeepUL, Ho2020DenoisingDP, Dhariwal2021DiffusionMB}, auto-regressive models~\cite{Yu2022ScalingAM, Ding2021CogViewMT, Gafni2022MakeASceneST, Chang2022MaskGITMG}, and the availability of gigantic text-image datasets~\cite{Schuhmann2022LAION5BAO}, 
large-scale text-to-image models~\cite{Nichol2021GLIDETP, Rombach2021HighResolutionIS, Ramesh2021ZeroShotTG, Saharia2022PhotorealisticTD, Sauer2023ARXIV, balaji2022eDiff-I, kang2023gigagan} have lead to a huge leap in performance. Our work uses the publicly available Stable Diffusion model based on Latent Diffusion Models~\cite{Rombach2021HighResolutionIS}.

Large-scale text-to-image models allow the user to generate a gallery of images for a given text prompt.
The control over the generated image, however, is limited, with attributes such as image composition, object shape, color, and texture changing depending on the arbitrary randomly sampled initial noise.
Thus, recent works have introduced additional spatial conditions to the model such as segmentation maps~\cite{Avrahami2022SpaTextSR, bar2023multidiffusion}, bounding boxes~\cite{Li2023GLIGENOG, Rombach2021HighResolutionIS}, keypoints~\cite{Li2023GLIGENOG} and other visual conditions \cite{voynov2022sketch, zhang2023adding, lhhuang2023composer}. Such conditions offer spatial control, but no object-level control. Alternatively, to gain object-level control, numerous text-guided image editing methods have been recently developed. 

\subsection{Text-Guided Image Editing}
Generative models are a powerful tool for image editing~\cite{styleflow, shen2020interpreting, Patashnik_2021_ICCV, gal2021stylegannada, bar2022text2live}. With the increased performance of text-to-image diffusion models, many methods~\cite{meng2022sdedit, hertz2022prompt, Chefer2023AttendandExciteAS} have utilized them for text-guided image editing. A simple approach adds noise to the input image and then denoises it with a guiding prompt~\cite{avrahami2022blended, avrahami2022blended_latent, meng2022sdedit, Couairon2022DiffEditDS}. To localize the edit, a user-defined mask is required~\cite{Rombach2021HighResolutionIS}. Another approach manipulates internal representations of the model (\eg, attention maps) during the generation process~\cite{hertz2022prompt, pnpDiffusion2022, Parmar2023ZeroshotIT} to preserve the image layout. Other methods operate in the text encoder latent space~\cite{gal2022textual} and possibly fine-tune the generator~\cite{ruiz2022dreambooth, kawar2022imagic, Gal2023DesigningAE}, or train a model on image pairs~\cite{brooks2022instructpix2pix}. 

Recently, it has been shown that one can change an object's semantics by switching the text prompt along the denoising process~\cite{liew2022magicmix}. This method shares similarities with our method since they also mix prompts. However, their method is limited to appearance modifications as a means to change the semantics and explicitly preserve the object's shape. In contrast, our work focuses on object geometric modifications.
It should be noted that unlike all the editing methods above, generating object shape variations is not an editing task per-se, as it aims at generating multiple object-level variations for a given image while preserving the semantics of the object. 
Furthermore, our introduced editing localization techniques are complementary to the editing methods mentioned above as we later demonstrate.
\begin{figure*}
    \centering
    \includegraphics[width=\textwidth]{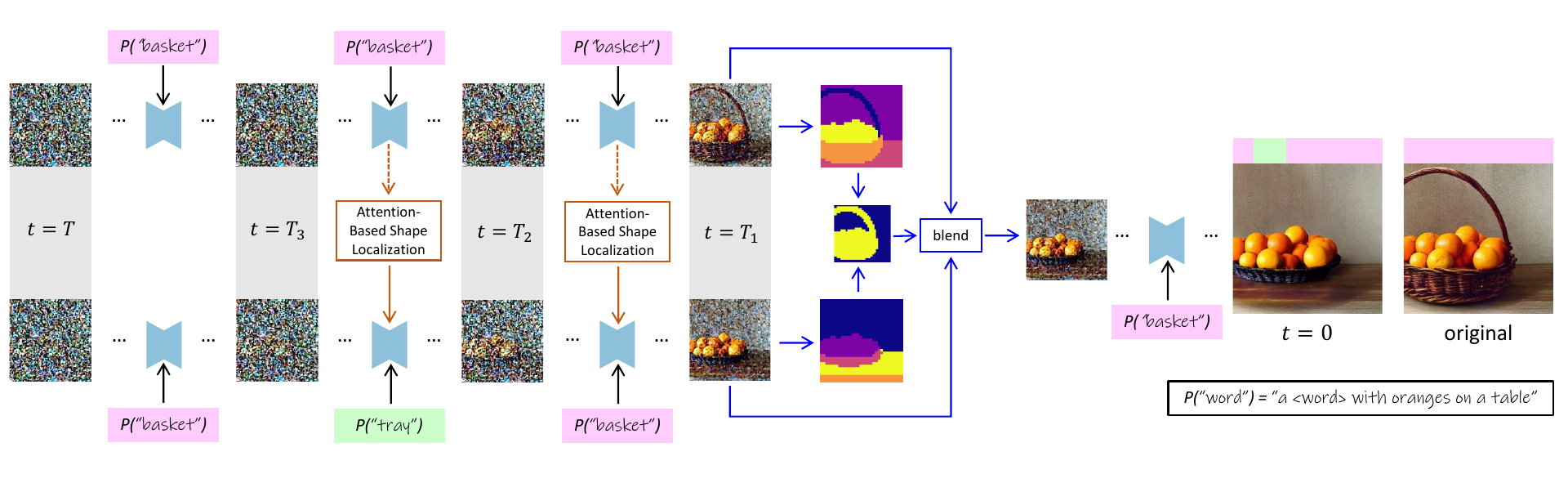}
    \caption{
    Given a reference image, and its corresponding denoising process, our full pipeline consists of three main building blocks.
    We perform Mix-and-Match in the timestamp intervals $[T, T_3], [T_3, T_2], [T_2, 0]$ using the prompt $P(w)$. For example, during the intervals $[T, T_3], [T_2, 0]$ we set $w=\text{``basket''}$, while during the interval $[T_3, T_2]$ we set $w=\text{``tray''}$. 
    During the denoising process, we apply our attention-based shape localization technique to preserve other objects' structures (here, ``table''). We do so by selectively injecting the self-attention map from the reference denoising process.
    At $t=T_1$, we apply controllable background preservation by segmenting the reference and the newly generated images, blend them, and proceed the denoising process.
    }
    \vspace{-12pt}
    \label{fig:overview}
\end{figure*}

\section{Preliminaries}

\paragraph{Latent Diffusion Models}
We demonstrate our method applied over the publicly available Stable Diffusion model which is built over the Latent Diffusion Models (LDM) architecture~\cite{Rombach2021HighResolutionIS}. 
In LDM, a diffusion model operates in the latent space of a pretrained autoencoder. 

The denoising network is implemented as a UNET and consists of self-attention layers followed by cross-attention layers.
At each timestep $t$, the noised spatial code $z_t$ is passed as input to the denoising network. The intermediate features of the network, denoted by $\phi(z_t)$, receive information from the self and cross-attention layers. The attention mechanism consists of three main components: Keys ($K$), Queries ($Q$), and Values ($V$). The Keys and the Queries together form an attention map, which is multiplied by the Values. In this work, we utilize the attention maps of both the self and cross-attention layers.
\vspace{-16pt}

\paragraph{Cross-Attention Layers in LDM}
Text guidance in LDM is performed using the cross-attention mechanism. 
Specifically, denoting the text encoding by $c$, $Q=f_Q(\phi(z_t))$, $K=f_K(c)$, and $V=f_V(c)$ are obtained using learned linear layers $f_Q, f_K, f_V$. 
Each token in the text prompt corresponds to an attention map formed by the Queries and the Keys, which is multiplied by each token's Values. Therefore, intuitively, the Keys and the Queries control the placement of each token, while the Values control its shape and appearance, as we later show in the supplementary materials. It is important to note, however, that these components are not fully disentangled.

By the definition of the cross-attention mechanism, it can be observed that the encoding of the text prompt is fed only to $f_K$ and $f_V$, and therefore the Keys and the Values are the only components affected directly by the text prompt.
\vspace{-16pt}

\paragraph{Self-Attention Layers in LDM}
Self-attention layers model the relation between each pixel to all the other pixels. In LDM, each such pixel corresponds to a patch in the final generated image. Previous works~\cite{hertz2022prompt, pnpDiffusion2022} have shown that self-attention strictly controls the image layout and objects' shapes in it, and is therefore useful to preserve the input image structure in image editing. In our work, we aim to modify the object of interest and preserve the remainder of the image. Thus, injecting the entire self-attention map does not allow for shape variations on the object of interest as was demonstrated in the above works.

\section{Prompt-Mixing} \label{sec:analysis}

To generate object variations, we propose a method that operates during inference time and does not require any optimization or model training. Given a text prompt $P$ and an object of interest, represented by a word $w$, we manipulate the denoising process to obtain object-level variations. 

The key enabler for generating shape variations for an object is our prompt-mixing technique. 
In prompt-mixing, different prompts are used in different time intervals of the denoising process. Specifically, we define three timestep intervals, $[T, T_3], [T_3, T_2], [T_2, 0]$, and use a different prompt in each interval to guide the denoising process. We denote by $P_{[t, t']}$ the prompt used in interval $[t, t']$. 
This technique is based upon insights related to the coarse-to-fine nature of the denoising process. These insights were also mentioned in \cite{balaji2022eDiff-I, Chefer2023AttendandExciteAS, voynov2022sketch, daras2022multiresolution, choi2021ilvr}, and we further analyze them in Section~\ref{sec:diff-stages}.

Note that common to image editing techniques, localizing the manipulated region is crucial for a successful result. In Section~\ref{sec:localization}, we introduce two generic techniques for achieving localized editing, and use them in our full pipeline for object variations illustrated in Figure~\ref{fig:overview}.

\subsection{Denoising Diffusion Process Stages} \label{sec:diff-stages}

\begin{figure}
    \centering
    \setlength{\tabcolsep}{0pt}
    {\scriptsize
    \begin{tabular}{ccc c c c c}
        \multicolumn{3}{c}{Single Prompt} &
        & 
        \multicolumn{3}{c}{Prompt-Mixing} \\
    
        \includegraphics[width=0.19\linewidth]{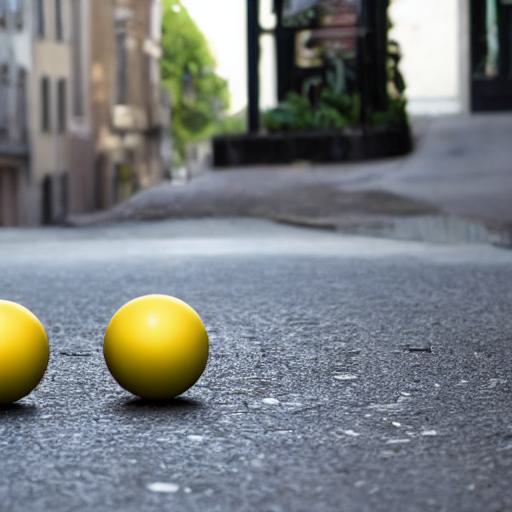} &
        \includegraphics[width=0.19\linewidth]{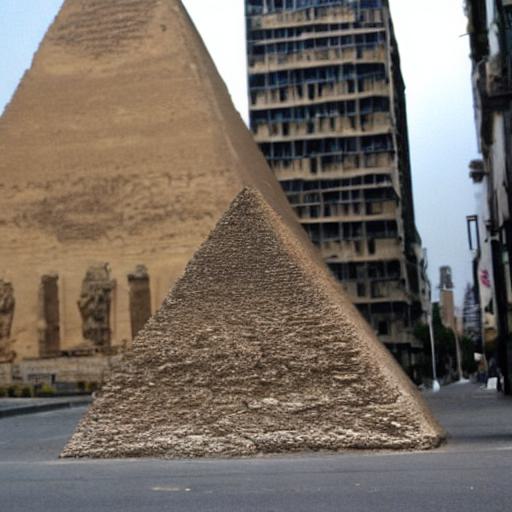} &
        \includegraphics[width=0.19\linewidth]{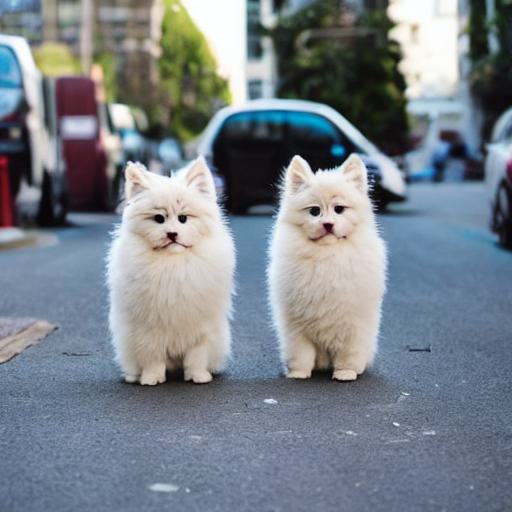} &
        { } &
        \includegraphics[width=0.19\linewidth]{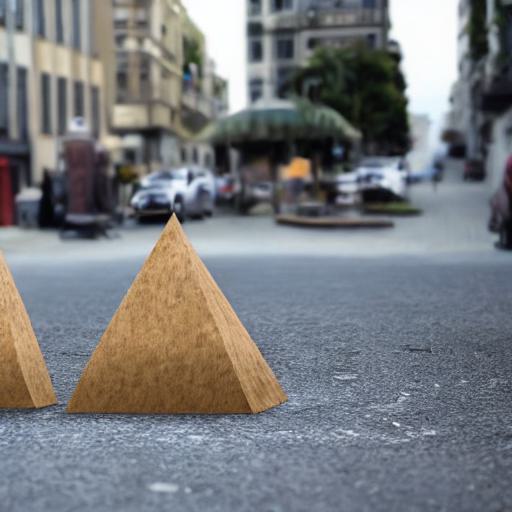} &
        { } &
        \includegraphics[width=0.19\linewidth]{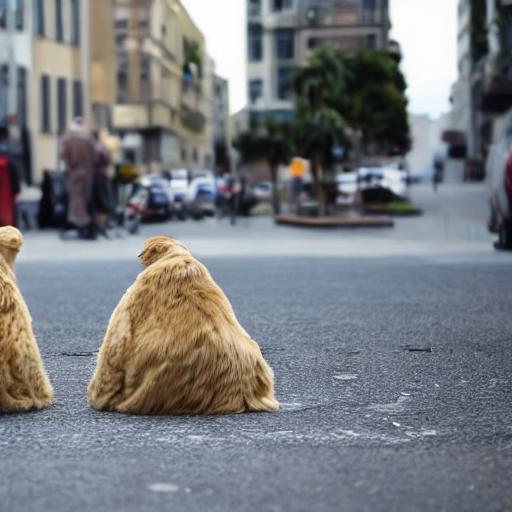} \\
        \includegraphics[width=0.19\linewidth]{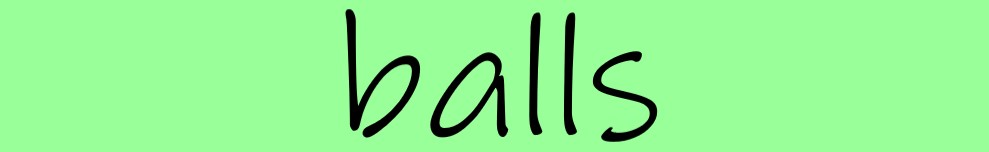} &
        \includegraphics[width=0.19\linewidth]{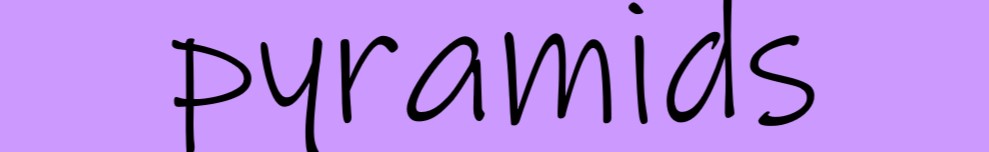} &
        \includegraphics[width=0.19\linewidth]{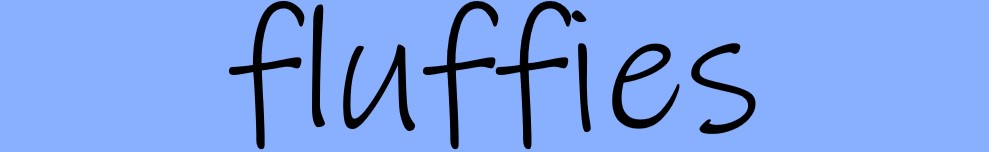} &
        { } &
        \includegraphics[width=0.19\linewidth]{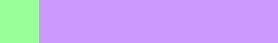} &
        {}&
        \includegraphics[width=0.19\linewidth]{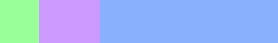} \\
        \includegraphics[width=0.19\linewidth]{images/analysis/balls.jpg} &
        \includegraphics[width=0.19\linewidth]{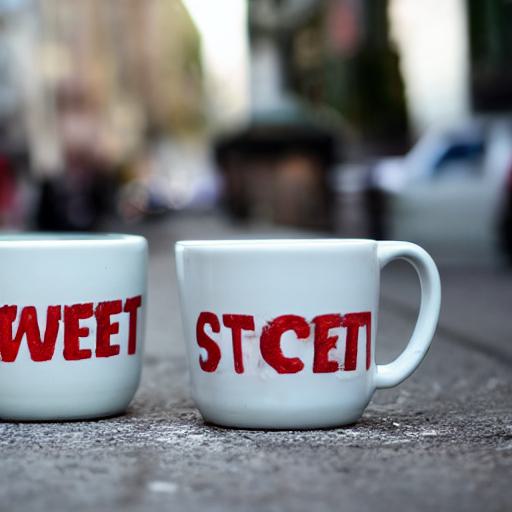} &
        \includegraphics[width=0.19\linewidth]{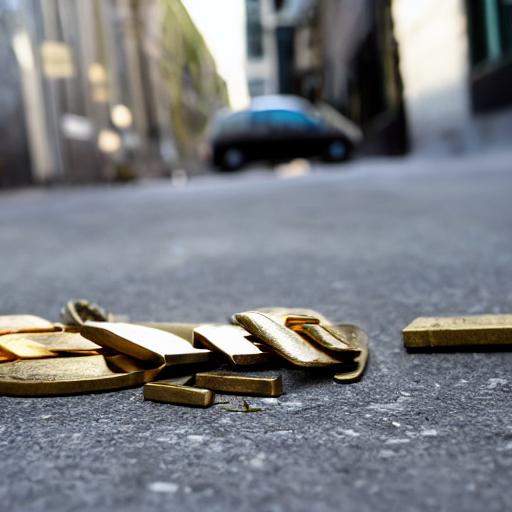} &
        { } &
        \includegraphics[width=0.19\linewidth]{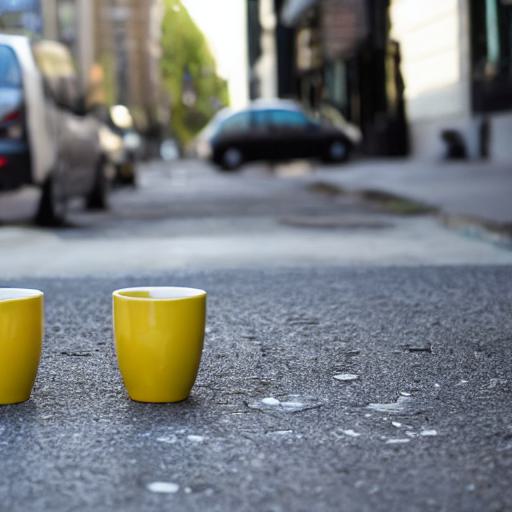} &
        {} &
        \includegraphics[width=0.19\linewidth]{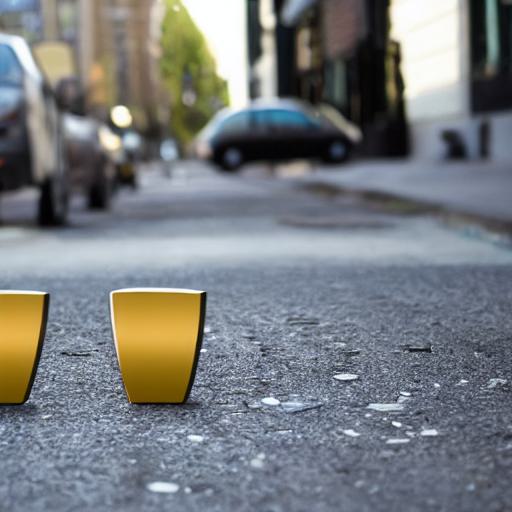} \\
        \includegraphics[width=0.19\linewidth]{images/analysis/balls_bar.jpg} &
        \includegraphics[width=0.19\linewidth]{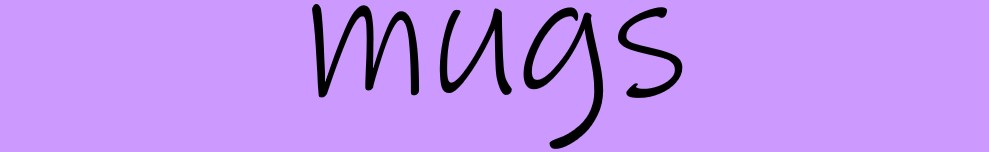} &
        \includegraphics[width=0.19\linewidth]{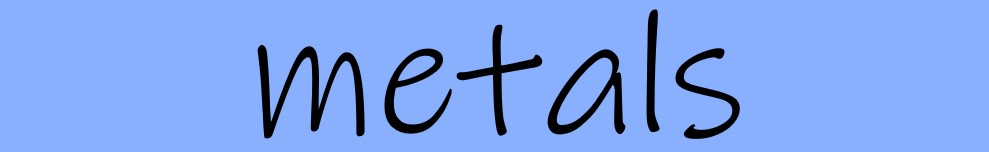} &
        { } &
        \includegraphics[width=0.19\linewidth]{images/analysis/balls_pyramids_bar.jpg} &
        {} &
        \includegraphics[width=0.19\linewidth]{images/analysis/balls_pyramids_fluffies_bar.jpg} 
    \end{tabular}
    }
    
    \caption{
    Prompt-mixing. For the prompt ``Two $\left<w\right>$ on the street'',
    the colored bars under the images on the right represent the corresponding word used along the denoising process.
    }
    \vspace{-12pt}
    \label{fig:analysis-stages}
\end{figure}

We analyze the timestep intervals defined above, to show the type of attributes in the image controlled by each interval. This analysis demonstrates the coarse-to-fine nature of the denoising process. We show the results of the analysis in Figure~\ref{fig:analysis-stages}.
In each row, the three leftmost images were generated 
using the prompt $P(w)= \enspace$``Two $\left<w\right>$ on the street'' along the entire denoising process, where $w$ represents a different word in each image. All images were generated using the same initial noise. 
For the two rightmost images in each row we apply prompt-mixing. Specifically,
we alter $w$ in the input prompt $P(w)$ in each time interval. 
As mentioned earlier, the input prompt is fed into the cross-attention layers, and directly affects the Keys and Values.
We use the altered prompt $P(w')$ to compute the Values, while using the original prompt $P(w)$ to compute the Keys. This design choice is explained in the supplementary materials.

In the fourth column of each row, we use $P_{[T, T_3]}({\color{ao}{w_1}})$ and $P_{[T_3, 0]}({\color{magenta}{w_2}})$.
As can be seen, we obtain an image containing {\color{magenta}{$w_2$}} (pyramids, mugs), with the layout and background of the images containing the balls ({\color{ao}{$w_1$}}).
In the fifth column, we use $P_{[T, T_3]}({\color{ao}{w_1}})$, $P_{[T_3, T_2]}({\color{magenta}{w_2}})$, and $P_{[T_2, 0]}({\color{blue}{w_3}})$.
As observed, we now obtain an image containing {\color{magenta}{$w_2$}} (pyramids, mugs), with the layout and background of the images containing the balls ({\color{ao}{$w_1$}}), and the fine visual details (\eg, texture) of {\color{blue}{$w_3$}} (fluffies, metals). 
We conclude that the first interval, $[T, T_3]$, controls the image layout, the second $[T_3, T_2]$ controls the shapes of the objects, and the third  $[T_2, 0]$ controls fine visual details (\eg, texture). We provide additional examples in the supplementary materials.

\subsection{Object Variations}

\paragraph{Mix-and-Match}

Let $P(w)$ denote the prompt of the original image, where $w$ denotes the word corresponding to the object of interest. 
To generate shape variations of the object of interest, we perform prompt-mixing
where $P_{[T, T_3]}(w) = P_{[T_2, 0]}(w)$. This is a special case of prompt-mixing, which we term \emph{Mix-and-Match}, since we perform mixing in the second interval, and match the prompts between the first and third intervals. Formally, our shape variations are achieved by using $P_{[T, T_3]}(w)$, $P_{[T_3, T_2]}(w')$, and $P_{[T_2, 0]}(w)$, where $w'$ is a ``proxy'' word (explained below).

Mix-and-Match allows keeping the original image layout, formed in the first interval, the shape of $w'$, set in the second interval, and the fine visual details of the original object represented by $w$ during the third interval.

\vspace{-12pt}
\paragraph{Proxy Words}

Here we describe our scheme for determining the proxy words. Intuitively, a proxy word represents a semantically close object to the object of interest, and whose shape is rather different.
Motivated by the use of CLIP~\cite{Radford2021LearningTV} for extracting text encodings in Stable Diffusion, we use CLIP's text space for finding the set of proxy words.

Given a word $w$, we seek to find the $k$ most similar tokens $\{w'_1, ..., w_k'\}$ to $w$. To this end, we consider all tokens $t$ of CLIP's tokenizer and embed each to the CLIP embedding space using prompts of the form $P_{\text{sim}}(t) =$ ``A photo of a $\left<t\right>$''. We then take the $k$ tokens with the smallest CLIP-space distances to the encoding of $P_\text{sim}(w)$, the prompt representing our object of interest. This gives us the $k$ tokens with the closest semantic meaning to $w$.
To take into account the input prompt context, we rank these $k$ tokens according to CLIP's distance between $P(t)$ and $P(w)$. Finally, we define the top $m$ tokens as proxy words. For each proxy word, we perform Mix-and-Match to obtain an image with a variation of the object of interest.

Note, that since we consider embeddings at the token level, some proxy words may not have semantic meaning alone. However, we observe that they still provide meaningful variations when used in our Mix-and-Match technique due to their close proximity in CLIP's embedding space.

\section{Edit Localization} \label{sec:localization}

As mentioned above, localizing the edit is especially challenging when changing the object's shape. To this end, we present two 
techniques that assist in localizing the edit from two different aspects. 
These localization techniques are crucial for successfully generating object shape variations. As we shall show, other known editing methods can also benefit from integrating them, leading to better localized manipulations.

\begin{figure}
    \includegraphics[width=\linewidth]{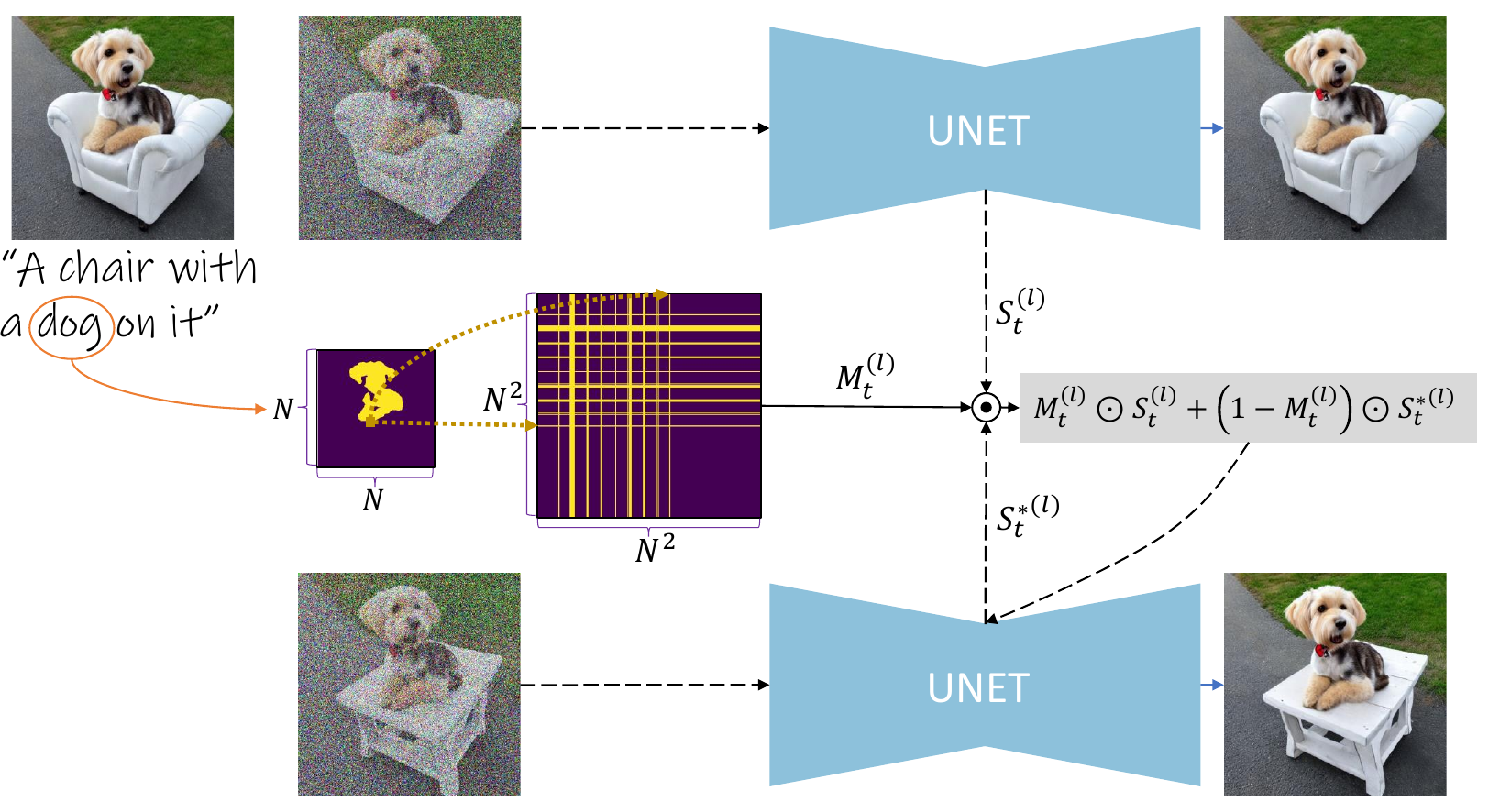}
    \vspace{1mm}
    \caption{Attention-based shape localization. Refer to Section~\ref{sec:attn-based-loc} for more details.}
    \vspace{-12pt}
    \label{fig:attn_localization}
\end{figure}

\subsection{Attention-Based Shape Localization} \label{sec:attn-based-loc}

To preserve the shapes of objects in the image, 
we introduce a shape localization technique based on injecting information from the self-attention maps of the source image into the self-attention maps of the generated image.
In the object variations pipeline, we apply this technique to objects that we aim to preserve. %
Injecting the full self-attention maps, even for a few steps, accurately preserves the structure of the original image, but at the same time prevents noticeable shape changes in the object we aim to change.

Our technique, depicted in Figure~\ref{fig:attn_localization}, revolves around a selective injection of self-attention maps. 
Consider a specific self-attention layer $l$ in the denoising network, which receives features of dimension $N\times N$, and the attention map formed by this layer, $S_t^{(l)}$, whose dimensions are $N^2 \times N^2$. The value $S_t^{(l)}[i, j]$ in the map indicates the extent to which pixel $j$ affects pixel $i$. In other words, row $i$ of the map shows the degree to which each pixel impacts pixel $i$, while column $j$ displays the degree to which pixel $j$ impacts all other pixels in the image. To preserve the shape of an object, we inject the rows and columns of the self-attention map that correspond to the pixels containing the object of interest.
Specifically, for a given denoising timestep $t$, the self-attention layer $l$,  and the self-attention map $S_t^{(l)}$, we define a corresponding mask $M_t^{(l)}$ by:
\vspace{-2pt}
\begin{equation}
    M_t^{(l)}[i, j] = 
    \begin{cases}
           1 &  i\in O_t^{(l)} \enspace\text{or}\enspace j \in O_t^{(l)}\\
           0 &\text{otherwise}, \\ 
         \end{cases}
\vspace{-2pt}
\end{equation}
where $O_t^{(l)}$ is the set of pixels corresponding to the object we aim to preserve. We explain later how we find $O_t^{(l)}$. After defining the mask $M_t^{(l)}$, the self-attention map in the newly generated image is changed to be: %
\vspace{-2pt}
\begin{equation}
    S_t^{*(l)} \leftarrow M_t^{(l)} \cdot S_t^{(l)} + (1 - M_t^{(l)}) S_t^{*(l)},
\end{equation}
where $S_t^{(l)}$ and $S_t^{*(l)}$ are the self-attention maps of the original and the newly generated images, respectively.
Additional mask controls are presented in the supplementary.

To find the pixels in which an object is located (\ie defining the set of pixels $O_t^{(l)}$), we leverage the cross-attention maps. These maps model the relations between each pixel in the image and each of the prompt's tokens. For an object we aim to preserve, we consider the cross-attention map of the corresponding token in the prompt. We then define the set $O_t^{(l)}$ of the object's pixels to be pixels with high activation in the cross-attention map by setting a fixed threshold.

\subsection{Controllable Background Preservation}

As shown in previous works~\cite{pnpDiffusion2022, hertz2022prompt}, self-attention injection preserves mainly structures.  
Therefore, we introduce a \emph{controllable background preservation} technique which preserves the appearance of the background and possibly some user-defined objects, specified by their corresponding nous in the input prompt. We give the user control to set the user-defined objects to be preserved since different images may require different configurations. 
For example, in Figure~\ref{fig:overview}, a user may want to preserve the oranges if the basket's size fits, while in other cases where the size of the basket is changed, it is desirable to change the oranges to properly fill a basket with a modified size.

To preserve the appearance of the desired regions, at $t=T_1$ we blend the original and the generated images, taking the changed regions (\eg, the object of interest) from the generated image and the unchanged regions (\eg, background) from the original image.
Next, we present our novel segmentation approach and describe the blending.

\vspace{-14pt}
\paragraph{Self-segmentation}
We perform the segmentation on noised latent codes and, as such, off-the-shelf semantic segmentation methods cannot be applied. Hence, we introduce a segmentation method that segments the image based on self-attention maps, and labels each segment by considering cross-attention maps. The method is based on the premise that internal features of a generative model encode the information needed for segmentation~\cite{Collins20, zhang21}.

\begin{figure}
    \centering
    \setlength{\tabcolsep}{1pt}
    {\scriptsize
    \begin{tabular}{c c c c}
        \includegraphics[width=0.235\linewidth]{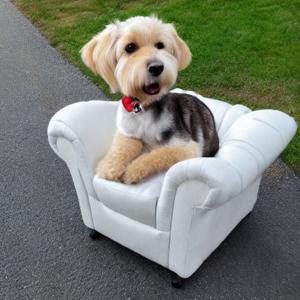} &
        \includegraphics[width=0.235\linewidth]{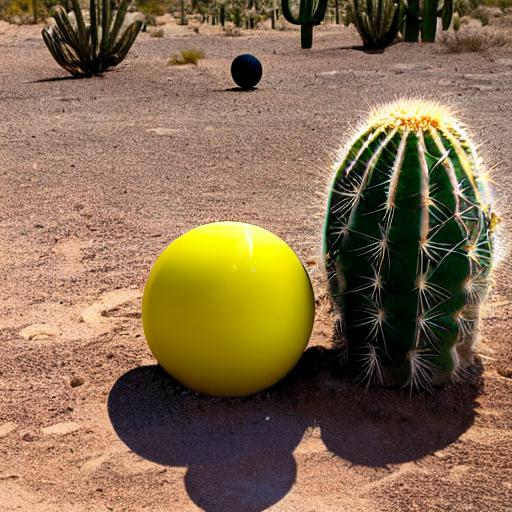} &
        \includegraphics[width=0.235\linewidth]{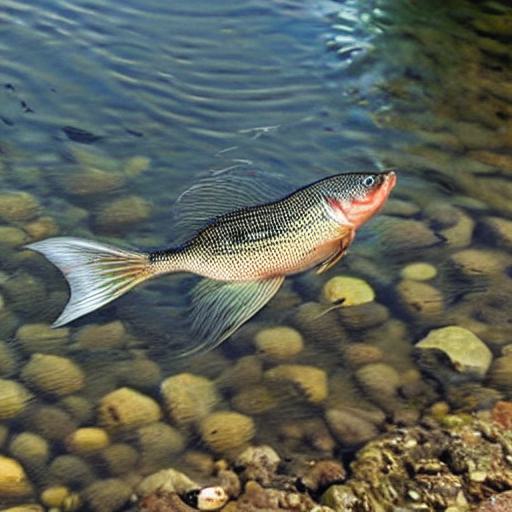} &
        \includegraphics[width=0.235\linewidth]{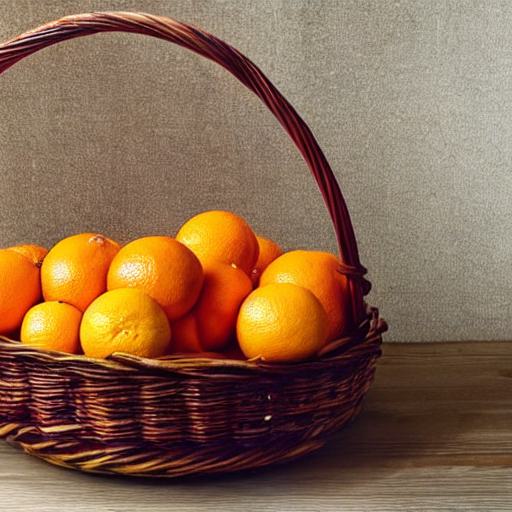} \\
        \includegraphics[width=0.235\linewidth]{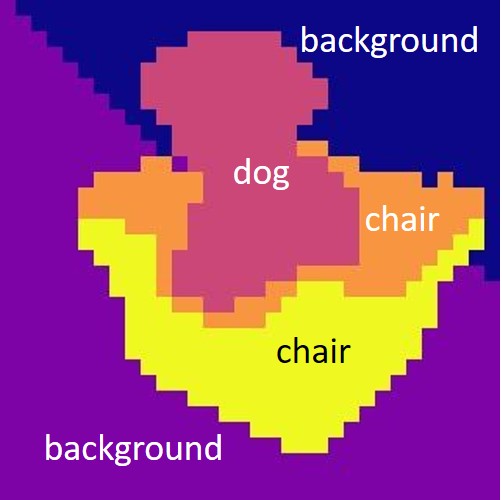} &
        \includegraphics[width=0.235\linewidth]{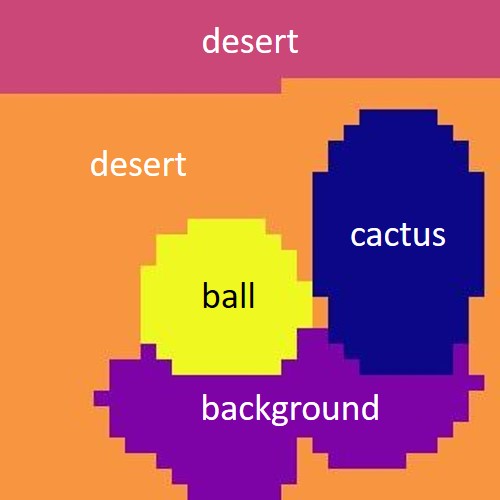} &
        \includegraphics[width=0.235\linewidth]{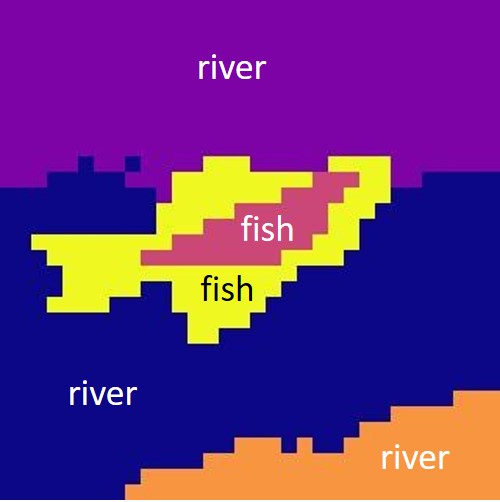} &
        \includegraphics[width=0.235\linewidth]{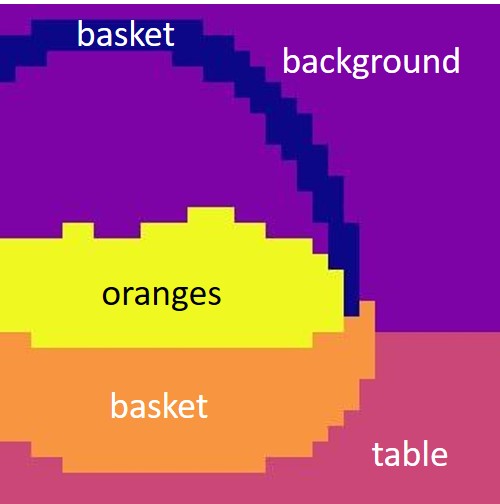} 
        
    \end{tabular}
    }
    \vspace{2mm}
    \caption{
    Segmentation maps obtained using our self-attention clustering technique. Each segment's label is determined by the cross-attention maps of the prompt nouns.
    }
    \label{fig:segmentation}
    \vspace{-8pt}
\end{figure}

At $t=T_1$, we average the $32^2 \times 32^2$ self-attention maps from the entire denoising process. We obtain an attention map of size $32^2 \times 32^2$, reshape it to $32 \times 32 \times 1024$, and cluster the deep pixels with the K-Means algorithm, where each pixel is represented by the $1024$ channels of the aggregated self-attention maps. Each resulting cluster corresponds to a semantic segment of the generated image. Several segmentation results are illustrated in Figure~\ref{fig:segmentation}.

\begin{figure*}
    \centering
    \setlength{\tabcolsep}{0pt}
    {\scriptsize
    \begin{tabular}{cc ccc cc cc ccc}
        Original (Synthetic) &&
        \multicolumn{3}{c}{$\longleftarrow$ Object level variations $\longrightarrow$} &&
        Original (Real) &&
        \multicolumn{3}{c}{$\longleftarrow$ Object level variations $\longrightarrow$} \\
        \includegraphics[width=0.12\textwidth]{images/our_results/cactus/cactus.jpg} &
        { } &
        \includegraphics[width=0.12\textwidth]{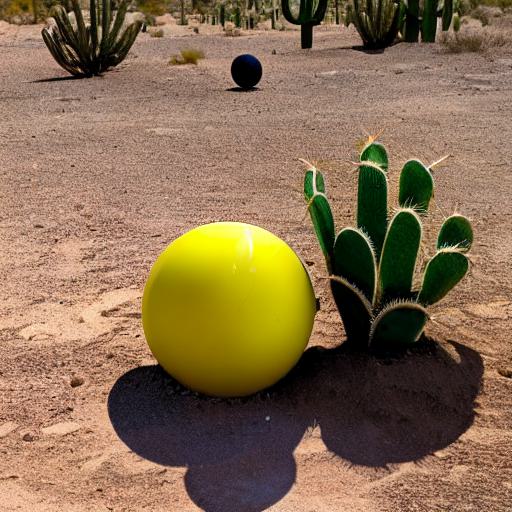} & 
        \includegraphics[width=0.12\textwidth]{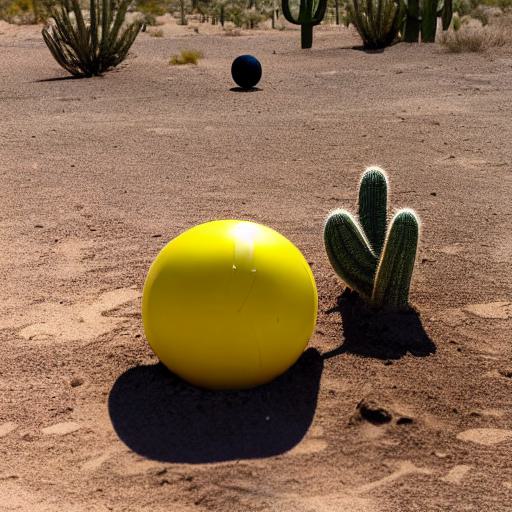} & 
        \includegraphics[width=0.12\textwidth]{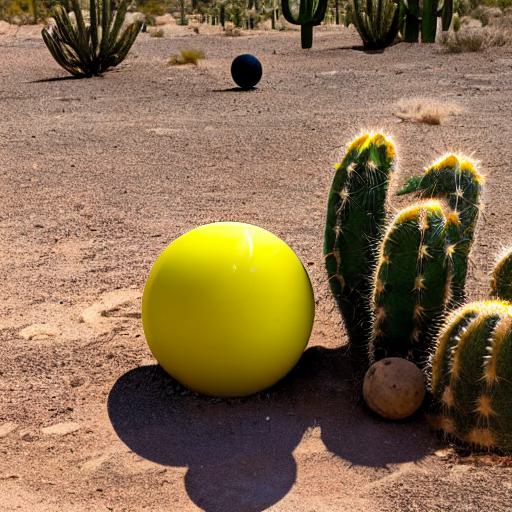} &
        { } &
        \includegraphics[width=0.12\textwidth]{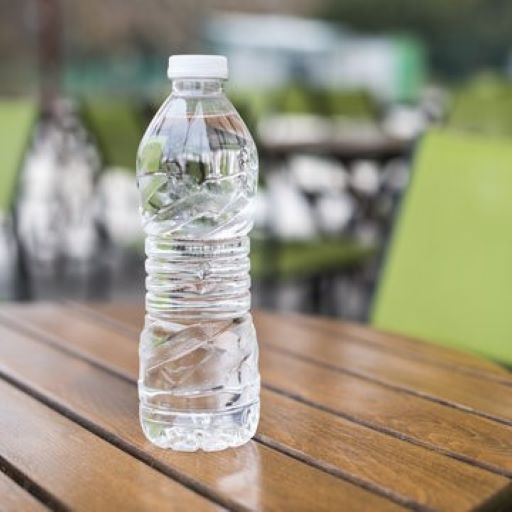} &
        { } &
        \includegraphics[width=0.12\textwidth]{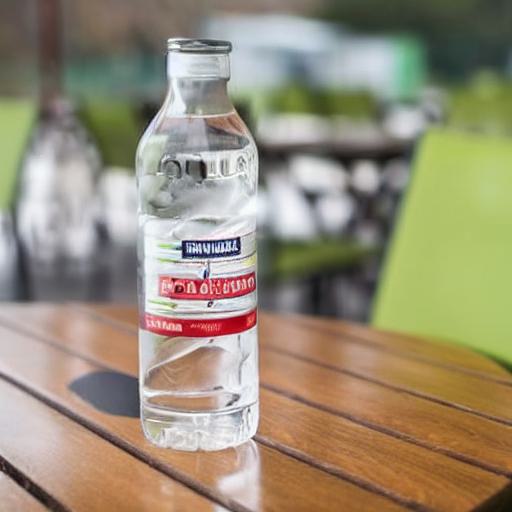} & 
        \includegraphics[width=0.12\textwidth]{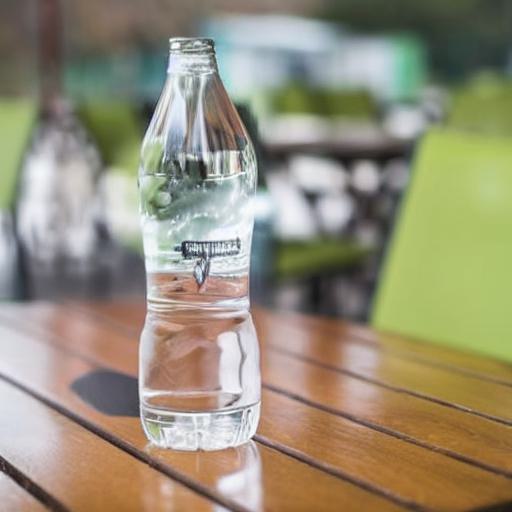} & 
        \includegraphics[width=0.12\textwidth]{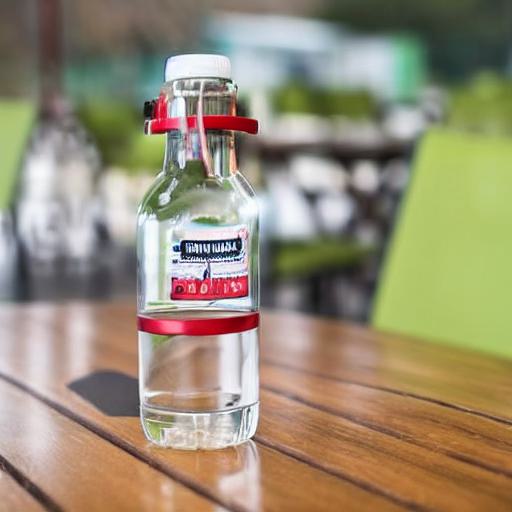} 
        \\
        && \multicolumn{3}{c}{``A \emph{cactus} and a ball in the desert''} & &
        && \multicolumn{3}{c}{``A \emph{bottle} on the table''} \\
        \includegraphics[width=0.12\textwidth]{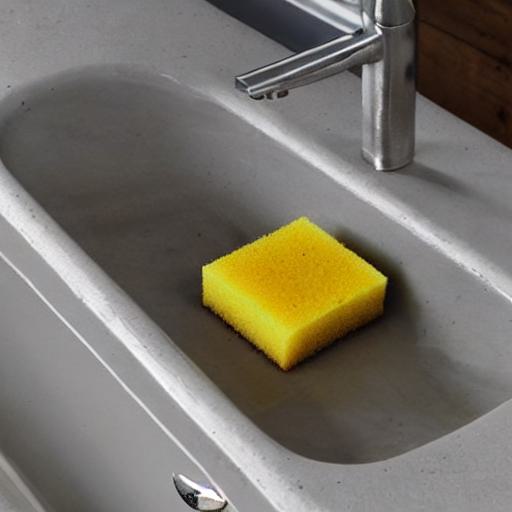} &
        { } &
        \includegraphics[width=0.12\textwidth]{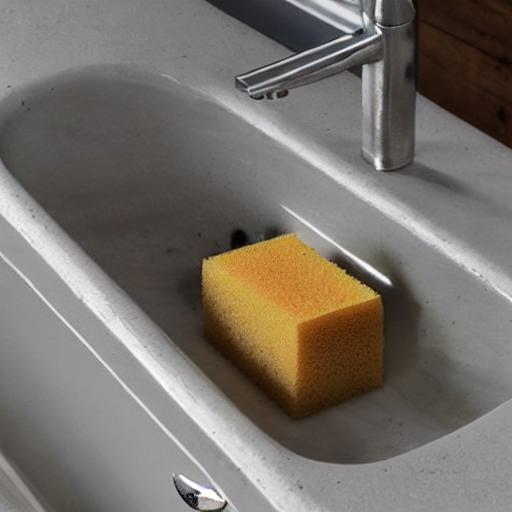} & 
        \includegraphics[width=0.12\textwidth]{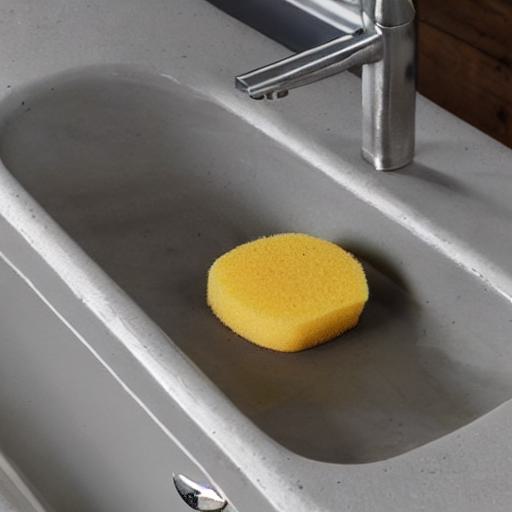} & 
        \includegraphics[width=0.12\textwidth]{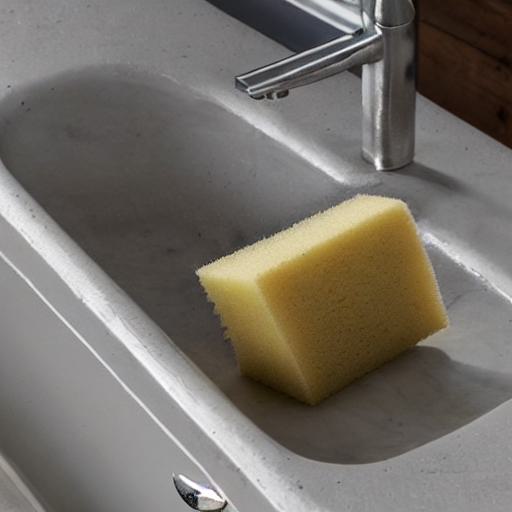} &
        { } &
        \includegraphics[width=0.12\textwidth]{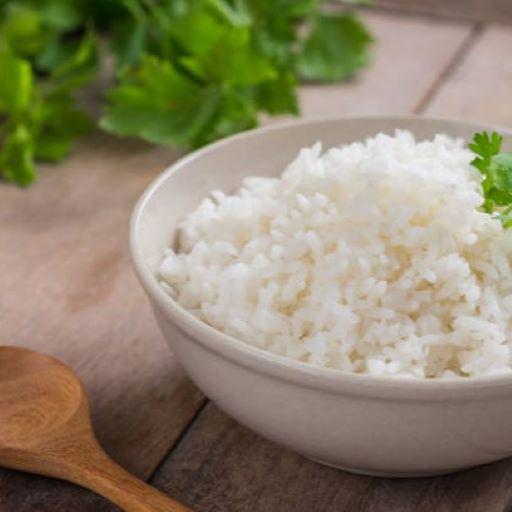} &
        { } &
        \includegraphics[width=0.12\textwidth]{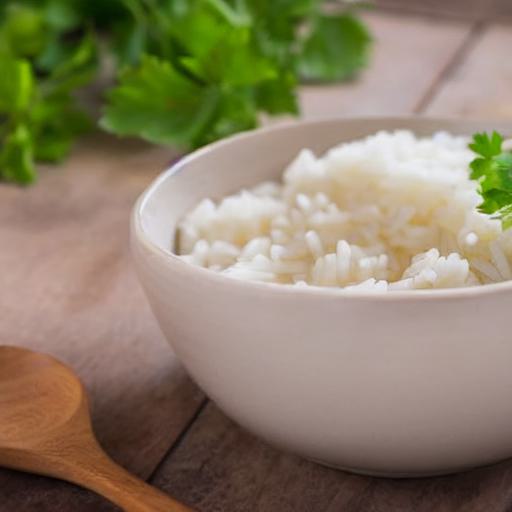} & 
        \includegraphics[width=0.12\textwidth]{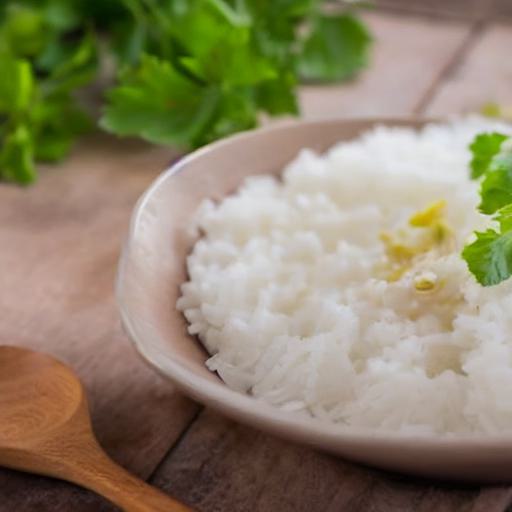} & 
        \includegraphics[width=0.12\textwidth]{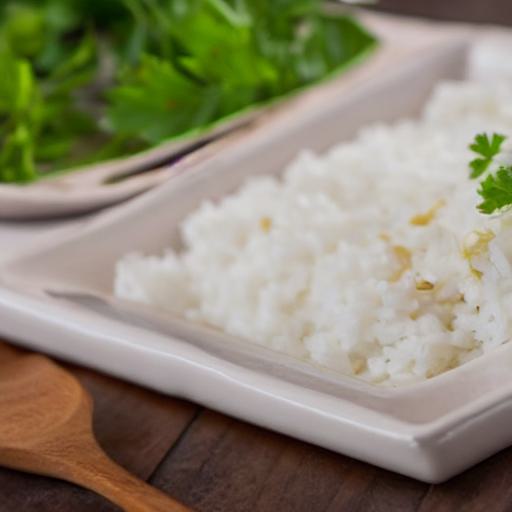} 
        \\
        && \multicolumn{3}{c}{``A \emph{sponge} and sink''} & &
        && \multicolumn{3}{c}{``A \emph{bowl} of rice on the table''} \\
        \includegraphics[width=0.12\textwidth]{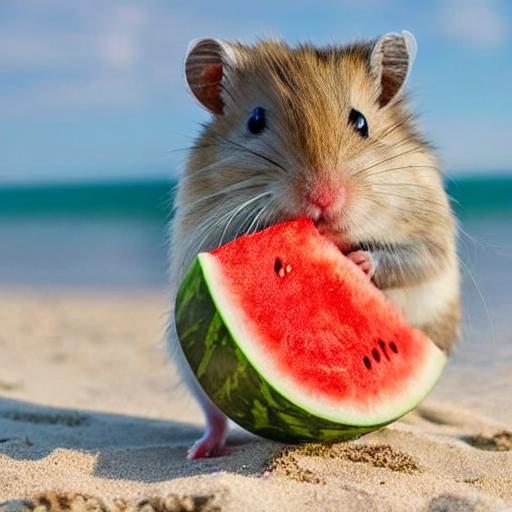} &
        { } &
        \includegraphics[width=0.12\textwidth]{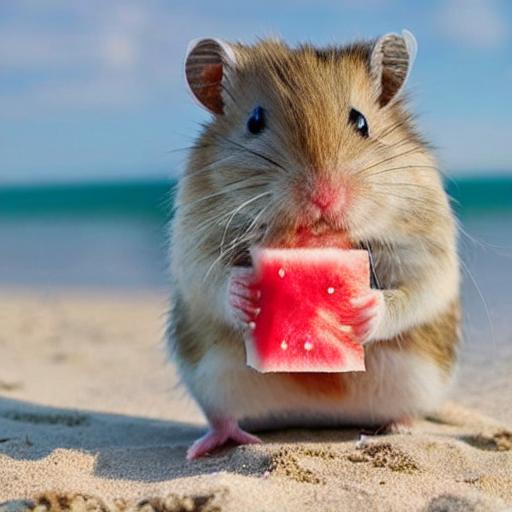} & 
        \includegraphics[width=0.12\textwidth]{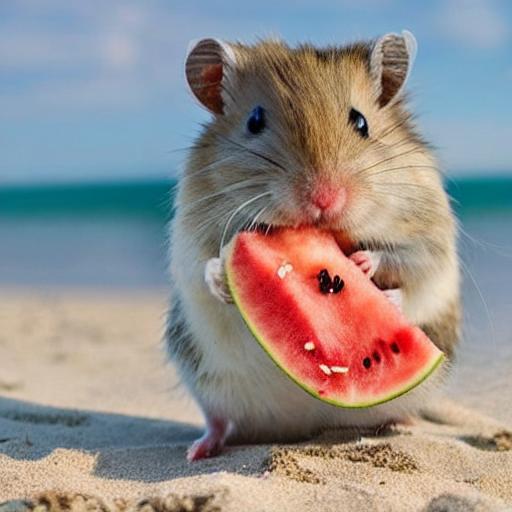} & 
        \includegraphics[width=0.12\textwidth]{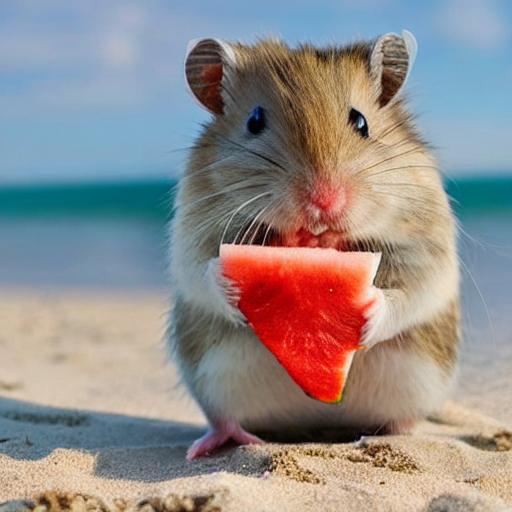} &
        { } &
        \includegraphics[width=0.12\textwidth]{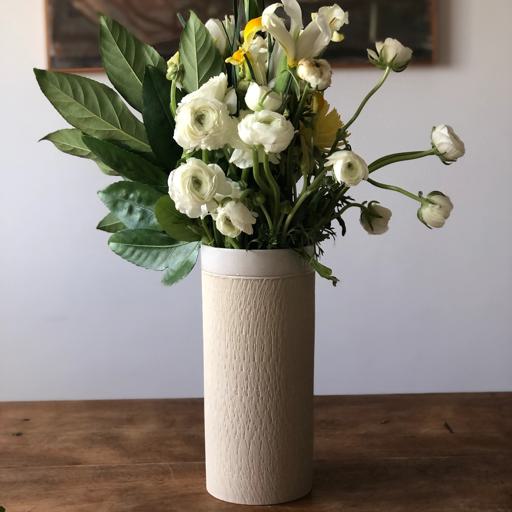} &
        { } &
        \includegraphics[width=0.12\textwidth]{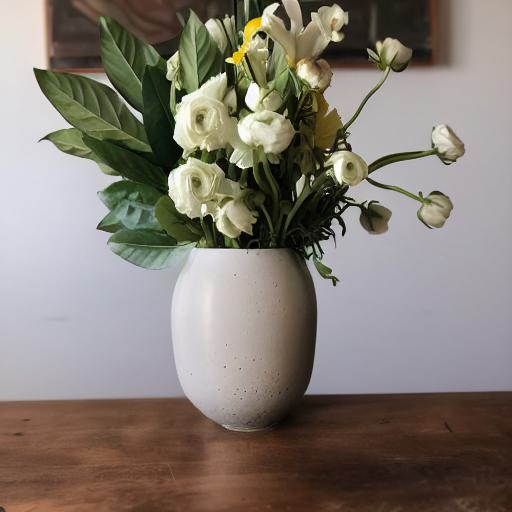} & 
        \includegraphics[width=0.12\textwidth]{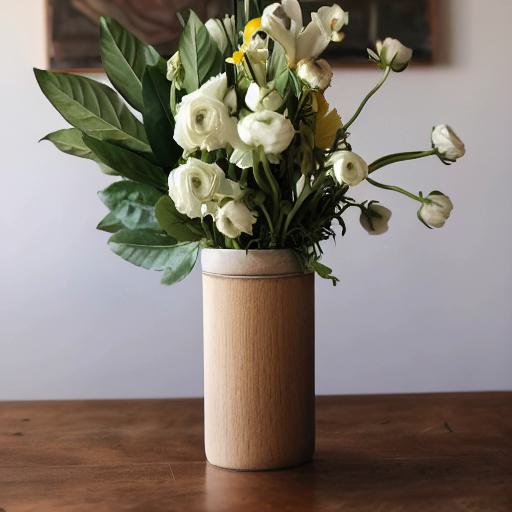} & 
        \includegraphics[width=0.12\textwidth]{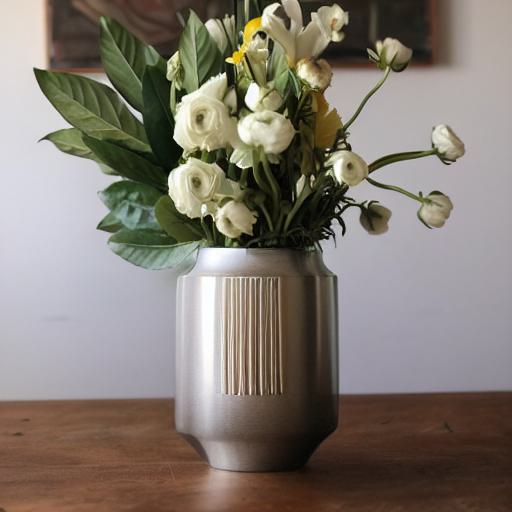} 
        \\
        && \multicolumn{3}{c}{``A hamster eating \emph{watermelon} on the beach'' } & &
        && \multicolumn{3}{c}{``A \emph{vase} with flowers on the table''} \\
    \end{tabular}
    }
    \vspace{5pt}
    \caption{Object-level variations for various scenes, synthetic and real (inverted~\cite{mokady2022null}). For each scene, the leftmost image is the original one. The \emph{emphasized} word corresponds to the modified object. As observed, our method generates various shape variations for each object.}
    \vspace{-10pt}
    \label{fig:our-results}
\end{figure*}

Having extracted the semantic segments, we match each segment with a noun in the input prompt. For each segment, we consider the normalized aggregated cross-attention maps of the prompt's nouns, and match each segment to a noun as follows. 
For segment $i$ that corresponds to a binary mask $M_i$, and for a noun $n$ in the prompt that corresponds to a normalized aggregated cross-attention mask $A_n$, we calculate a score $s(i, n) = \sum(M_i \cdot A_n) / \sum(M_i) $. We label segment $i$ with $\argmax_n {s(i, n)}$ if $\max_n {s(i, n) > \sigma}$ and label it as background otherwise. The threshold value $\sigma$ is fixed across all our experiments.

\vspace{-12pt}
\paragraph{Blending the original and generated images}
We use the segmentation map and the corresponding labels to overwrite relevant regions in the newly generated image. We retain pixels from the original image only if they are labeled as background or as a user-defined object in both the original and new images. 
This approach helps to overcome shape modifications in the object of interest, as illustrated by the example of the basket in Figure~\ref{fig:overview}, where the handle region is taken from the newly generated image. After blending the latent images, we proceed with the denoising process.

\vspace{-5pt}

\section{Experiments}

\subsection{Object Shape Variations} \label{sec:exp-obj-varietions}
We perform experiments to assess the effectiveness of our method in generating object-level shape variations, and compare it to other existing methods. Specifically, we compare our method with methods that directly provide variations for an image by changing seed (inpainting~\cite{Rombach2021HighResolutionIS}, SDEdit~\cite{meng2022sdedit}), as well as recent text-guided image editing baselines (P2P~\cite{hertz2022prompt}, I-Pix2Pix~\cite{brooks2022instructpix2pix}, PnP~\cite{pnpDiffusion2022}, Zero-shot Image2Image Translation~\cite{Parmar2023ZeroshotIT}, Imagic~\cite{kawar2022imagic}). 

It should be noted that our method differs from image editing approaches in two key aspects. First, most editing methods change the object to an object of a different class, whereas our method keeps class and offers alternative options for the same object. Second, editing methods require the user to provide specific instructions on the exact object they would like to obtain, whereas our method allows for an open-ended exploration without relying on such instructions. Therefore, we adopted two different approaches when comparing our method with image editing methods: (i) we refined the input prompt, and (ii) we replaced the word representing the object of interest with proxy words. The second approach, which involves using proxy words, can also be viewed as an ablation study for Mix-and-Match.

We separately evaluate three important aspects of our method. First, we aim at achieving high diversity in the shape of the object of interest. Second, the object class should remain the same. We term this evaluation objective as object faithfulness. Third, regions of the image other than the object of interest should be preserved.

\vspace{-8pt}
\paragraph{Qualitative Experiments}

In Figure~\ref{fig:our-results}, we show a gallery of images. Additional results are provided in the supplementary materials.
Observe the shape diversity in our method's results and the preservation of the original image. 

\begin{figure}
       \centering
        \setlength{\tabcolsep}{0.0pt}
        {\scriptsize
        \begin{tabular}{ccc cccc}
            \multicolumn{7}{c}{``A \emph{sofa} with a cat on it''} \\
            && Original &&
            \multicolumn{3}{c}{$\longleftarrow$ Object level variations $\longrightarrow$} \\
            \raisebox{16pt}{\rotatebox{90}{ ours}} &
            { } &
            \includegraphics[width=0.24\linewidth]{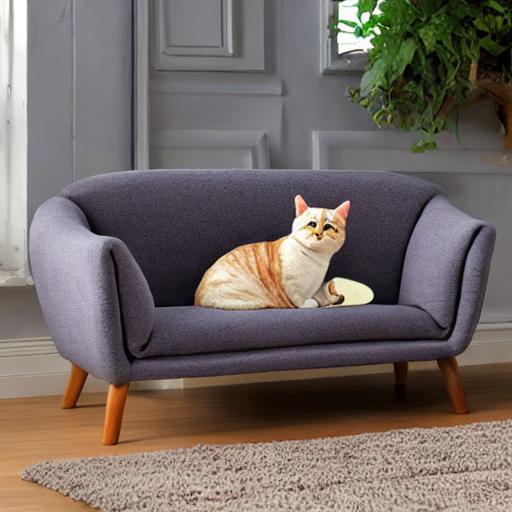} &
            {  } &
            \includegraphics[width=0.24\linewidth]{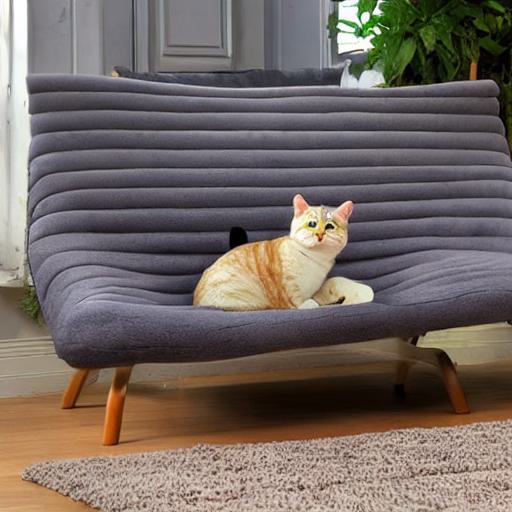} &
            \includegraphics[width=0.24\linewidth]{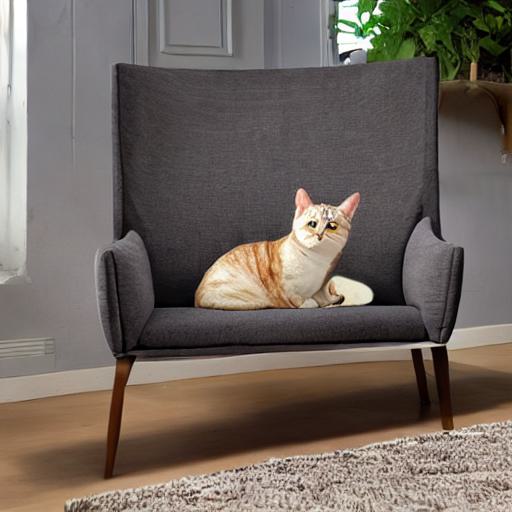} &
            \includegraphics[width=0.24\linewidth]{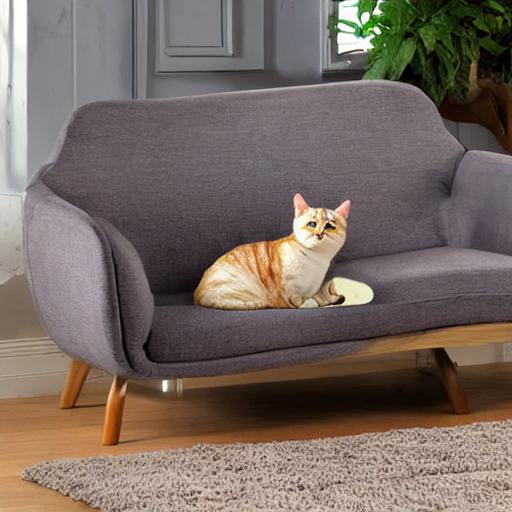} \\ 
            \raisebox{10pt}{\rotatebox{90}{ Inpainting}} &
            { } &
            \includegraphics[width=0.24\linewidth]{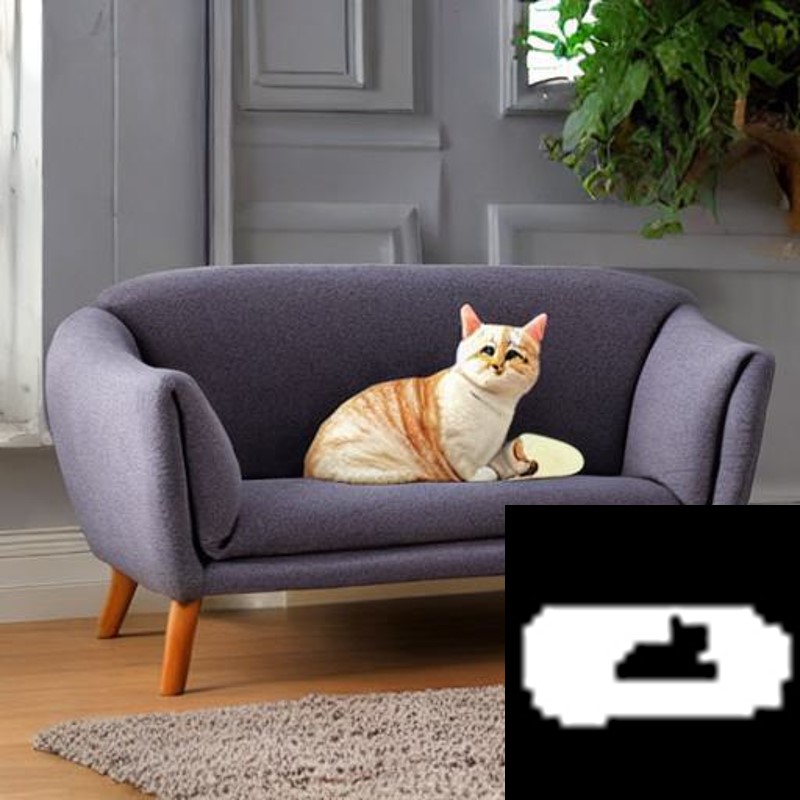} &
            {  } &
            \includegraphics[width=0.24\linewidth]{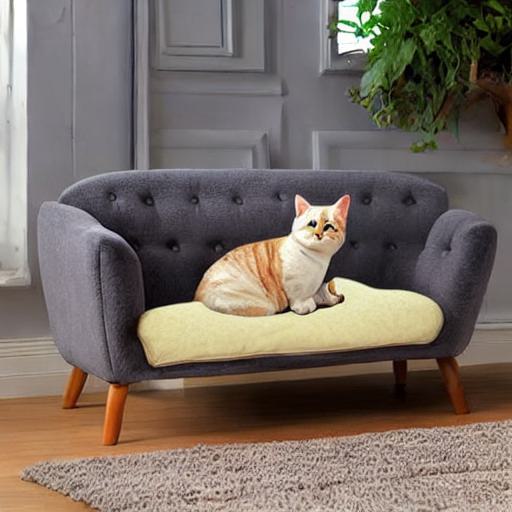} &
            \includegraphics[width=0.24\linewidth]{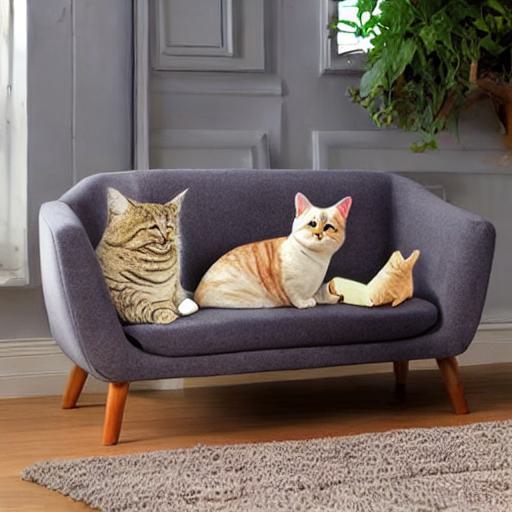} &
            \includegraphics[width=0.24\linewidth]{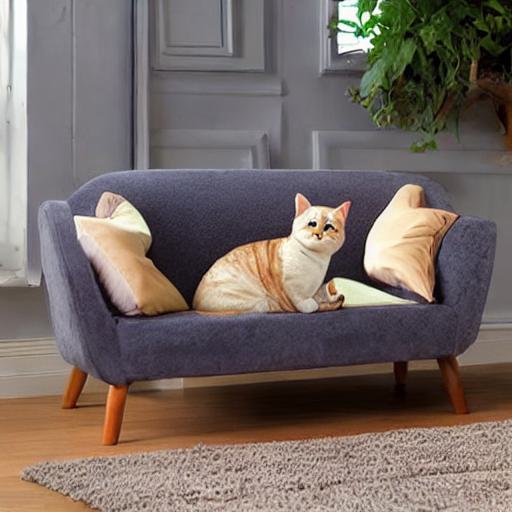}  \\ 
            \raisebox{14pt}{\rotatebox{90}{ SDEdit}} &
            { } &
            \includegraphics[width=0.24\linewidth]{images/compare_inrepainting/ours/sofa.jpg} &
            {  } &
            \includegraphics[width=0.24\linewidth]{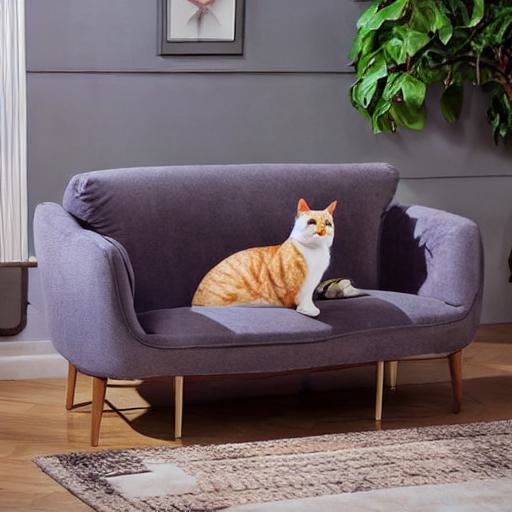} &
            \includegraphics[width=0.24\linewidth]{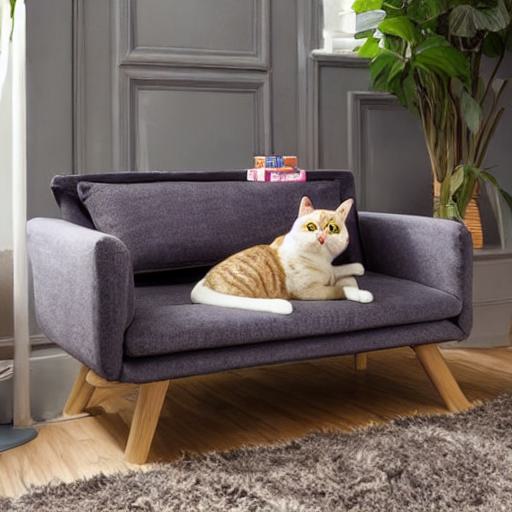} &
            \includegraphics[width=0.24\linewidth]{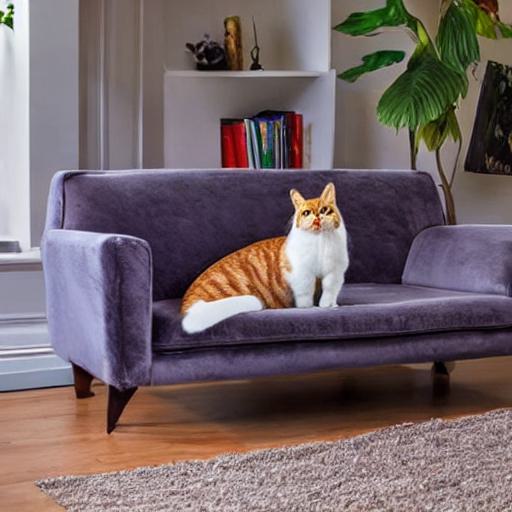} \\
        \end{tabular}
        }
    \vspace{-4pt}
    \captionof{figure}{Comparing to methods that generate variations. As can be seen, our method generates varied shape variations, while other methods change mostly the texture. }
    \vspace{-13pt}
    
    \label{fig:comp_inpainting}
\end{figure}

In Figure~\ref{fig:comp_inpainting} we compare our method with inpainting~\cite{Rombach2021HighResolutionIS} and SDEdit~\cite{meng2022sdedit} by sampling different seeds to get variations. For inpainting, we use a mask obtained from our segmentation technique.
Similarly to Figure~\ref{fig:inpainting}, we observe that applying inpainting results mainly in texture changes, while SDEdit performs small shape changes but does not preserve the background and other objects (\eg, cat). 

\begin{figure}
       \centering
        \setlength{\tabcolsep}{0.0pt}
        {\scriptsize
        \begin{tabular}{c c c c c c c c c}
            \multicolumn{9}{c}{"A \emph{chair} with a dog on it"} \\
            & {} &&
            { Original } &
            \multicolumn{5}{c}{$\longleftarrow$ Object level variations $\longrightarrow$}
            \\
            \raisebox{12pt}{\rotatebox{90}{ Ours }} & {} &&
            \includegraphics[width=0.19\linewidth]{images/comparison/ours/chair.jpg} &
            {  } &
            \includegraphics[width=0.19\linewidth]{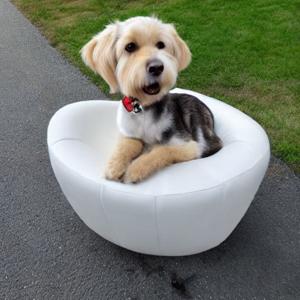} &
            \includegraphics[width=0.19\linewidth]{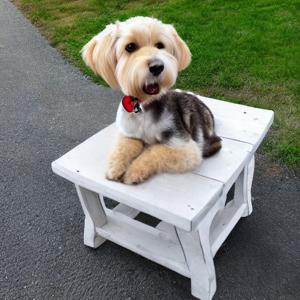} &
            \includegraphics[width=0.19\linewidth]{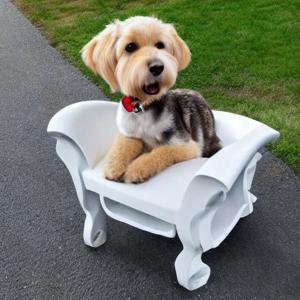} &
            \includegraphics[width=0.19\linewidth]{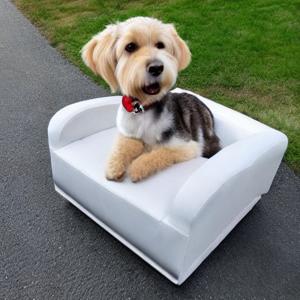} 
             \\
            \raisebox{7pt}{\rotatebox{90}{ P2P (40\%)  }} & {} &&
            \includegraphics[width=0.19\linewidth]{images/comparison/ours/chair.jpg} &
            {  } &
            \includegraphics[width=0.19\linewidth]{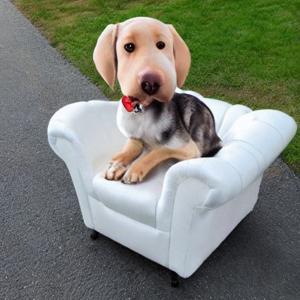} &
            \includegraphics[width=0.19\linewidth]{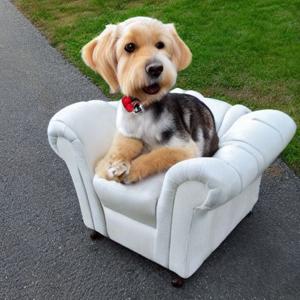} &
            \includegraphics[width=0.19\linewidth]{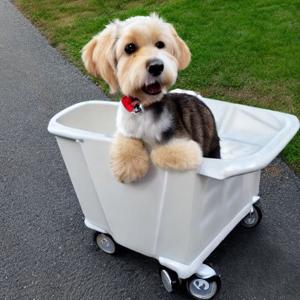} &
            \includegraphics[width=0.19\linewidth]{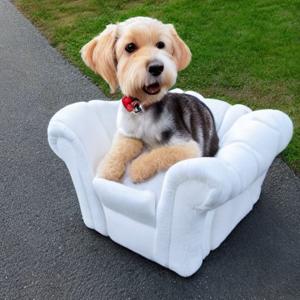} 
             \\
            \raisebox{7pt}{\rotatebox{90}{ P2P (10\%) }} & {} &&
            \includegraphics[width=0.19\linewidth]{images/comparison/ours/chair.jpg} &
            {  } &
            \includegraphics[width=0.19\linewidth]{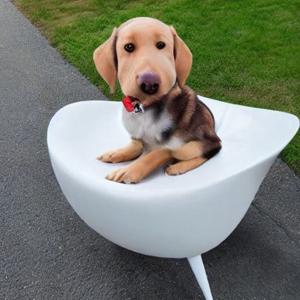} &
            \includegraphics[width=0.19\linewidth]{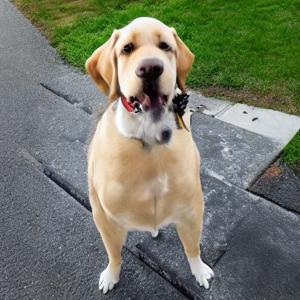} &
            \includegraphics[width=0.19\linewidth]{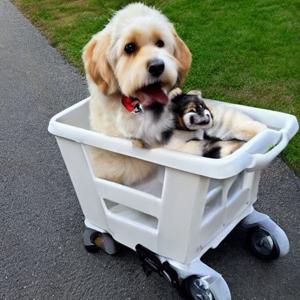} &
            \includegraphics[width=0.19\linewidth]{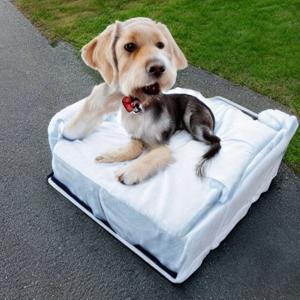} 
             \\
            \raisebox{7pt}{\rotatebox{90}{ I-Pix2Pix }} & {} &&
            \includegraphics[width=0.19\linewidth]{images/comparison/ours/chair.jpg} &
            {  } &
            \includegraphics[width=0.19\linewidth]{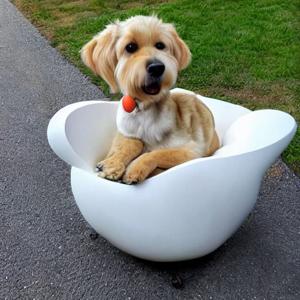} &
            \includegraphics[width=0.19\linewidth]{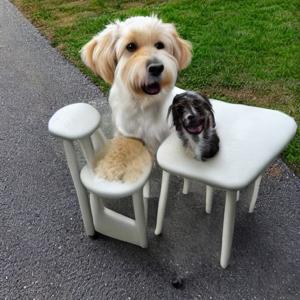} &
            \includegraphics[width=0.19\linewidth]{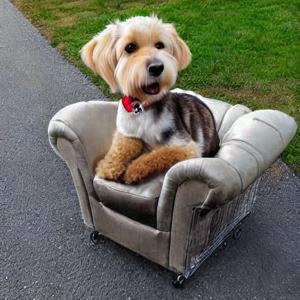} &
            \includegraphics[width=0.19\linewidth]{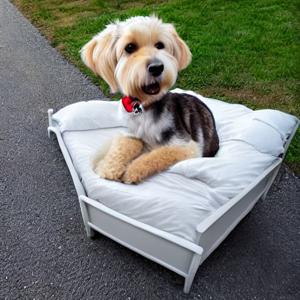}   \\
            \raisebox{15pt}{\rotatebox{90}{ PnP }} & {} &&
            \includegraphics[width=0.19\linewidth]{images/comparison/ours/chair.jpg} &
            {  } &
            \includegraphics[width=0.19\linewidth]{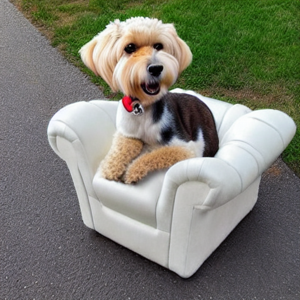} &
            \includegraphics[width=0.19\linewidth]{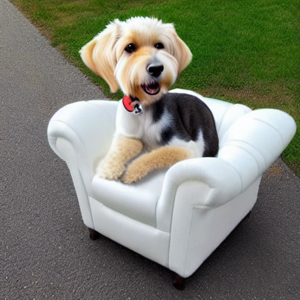} &
            \includegraphics[width=0.19\linewidth]{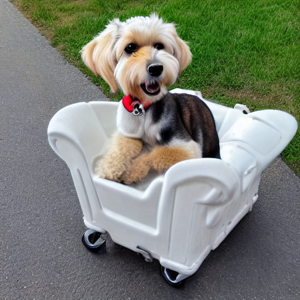} &
            \includegraphics[width=0.19\linewidth]{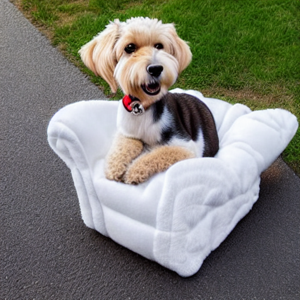} \\
            \raisebox{3pt}{\rotatebox{90}{ ZeroShotI2I }} & {} &&
            \includegraphics[width=0.19\linewidth]{images/comparison/ours/chair.jpg} &
            {  } &
            \includegraphics[width=0.19\linewidth]{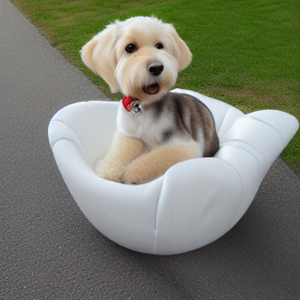} &
            \includegraphics[width=0.19\linewidth]{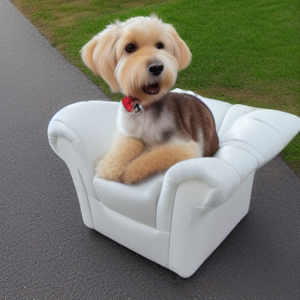} &
            \includegraphics[width=0.19\linewidth]{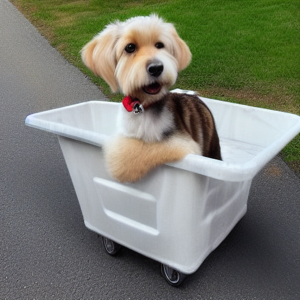} &
            \includegraphics[width=0.19\linewidth]{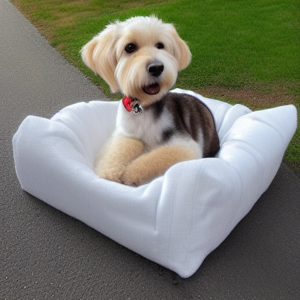} \\

            \raisebox{11pt}{\rotatebox{90}{ Imagic }} & {} &&
            \includegraphics[width=0.19\linewidth]{images/comparison/ours/chair.jpg} &
            {  } &
            \includegraphics[width=0.19\linewidth]{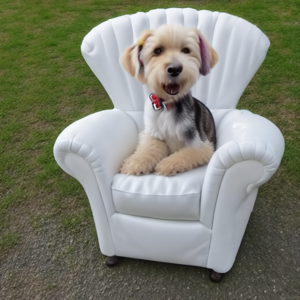} &
            \includegraphics[width=0.19\linewidth]{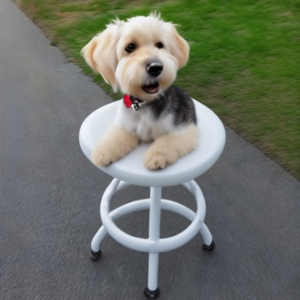} &
            \includegraphics[width=0.19\linewidth]{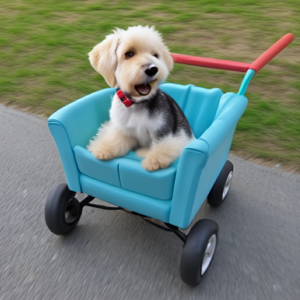} &
            \includegraphics[width=0.19\linewidth]{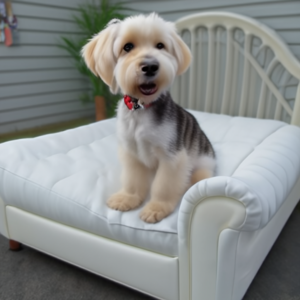} \\
            & {} &&
            { ``Chair'' } &
            {  } &
            { ``Eggchair'' } &
            { ``Stool'' } &
            { ``Cart'' } &
            { ``Bed'' } 
            \\

        \end{tabular}
        }
    \vspace{1mm}
    \captionof{figure}{
    Comparisons to text-guided editing methods. In each column, we show the results of a different word that replaces the original word ``chair'' in the prompt. We apply two different percentages of self-attention map injection steps in P2P.}
    \vspace{-12pt}
    \label{fig:comparision}
\end{figure}

Comparison to text-guided image editing methods is presented in Figure~\ref{fig:comparision}.
For each method, we guided the editing with refined prompts (``eggchair'', ``stool'') and with our automatic proxy words (``cart'', ''bed'', ``stool''). Note that the refined prompts were manually chosen to be types of chairs. 
It should be noted that refining prompts for objects that do not have subtypes (\eg, basket) is more challenging. 
In P2P~\cite{hertz2022prompt}, we show two versions of results, which differ in the number of self-attention injection steps (10\%, 40\%).
Additional information about the configuration of each method and comparison to additional methods are provided in the supplementary materials.

Our diverse results in Figure~\ref{fig:comparision} remain faithful to the class of chairs while preserving the rest of the image.
Not surprisingly, editing methods struggle at keeping the chair when a different object is specified in the prompt. For example, when replacing the chair with a cart, wheels are added. In our method, thanks to Mix-and-Match, we take the shape of the wheels but the fine visual details of the chair.
 
As can be seen in Figure~\ref{fig:comparision}, injecting self-attention maps for 40\% of the denoising steps in P2P~\cite{hertz2022prompt} prevents change in the object of interest (here, a chair). Conversely, when injecting self-attention maps for 10\% of the denoising steps, P2P struggles to preserve the dog and the background.  
Instruct-Pix2Pix~\cite{brooks2022instructpix2pix} results are diverse but inferior to our method in image preservation (eggchair) and faithfulness (bed, stool).
Plug-and-Play~\cite{pnpDiffusion2022} struggles at performing shape changes as it injects the entire self-attention maps along the denoising process.
Compared to Zero-shot Image2Image Translation~\cite{Parmar2023ZeroshotIT}, which requires 1000 prompts with the proxy word, our method preserves the dog colors better, and allows for more diverse shapes (see the stool). 
We also compared our method to Imagic \cite{kawar2022imagic} using Imagen \cite{Saharia2022PhotorealisticTD}. While Imagic produces high-quality results, our method has advantages in preserving the background and being more faithful to the class of a chair. Additionally, our method is more time-efficient than Imagic's optimization-based approach.

\paragraph{Quantitative Experiments}

Given a collection of object-level variations of an image, we measure each objective as follows. For shape diversity, we extract a mask of the object of interest by using CLIPSeg~\cite{Lddecke2021ImageSU}, and average the IoU of each pair of masks in the collection. We define the diversity as $1 - \text{IoU}$. We measure faithfulness by employing CLIP~\cite{Radford2021LearningTV} and computing cosine similarity between an averaged embedding of images containing the object of interest and each of the images in the collection. To quantify image preservation we utilize LPIPS~\cite{zhang2018perceptual}.

We create a dataset of 150 images with various prompts, and generate 20 variations of the object of interest with each method. More details about the construction of the dataset are provided in the supplementary materials.
We test other methods with proxy words (all methods), random seeds (inpainting, SDEdit), and prompt refinement (I-Pix2Pix, P2P).

\begin{figure}
    \centering
    \setlength{\tabcolsep}{0.0pt}
    \begin{tabular}{c c}
        \includegraphics[width=0.5\linewidth]{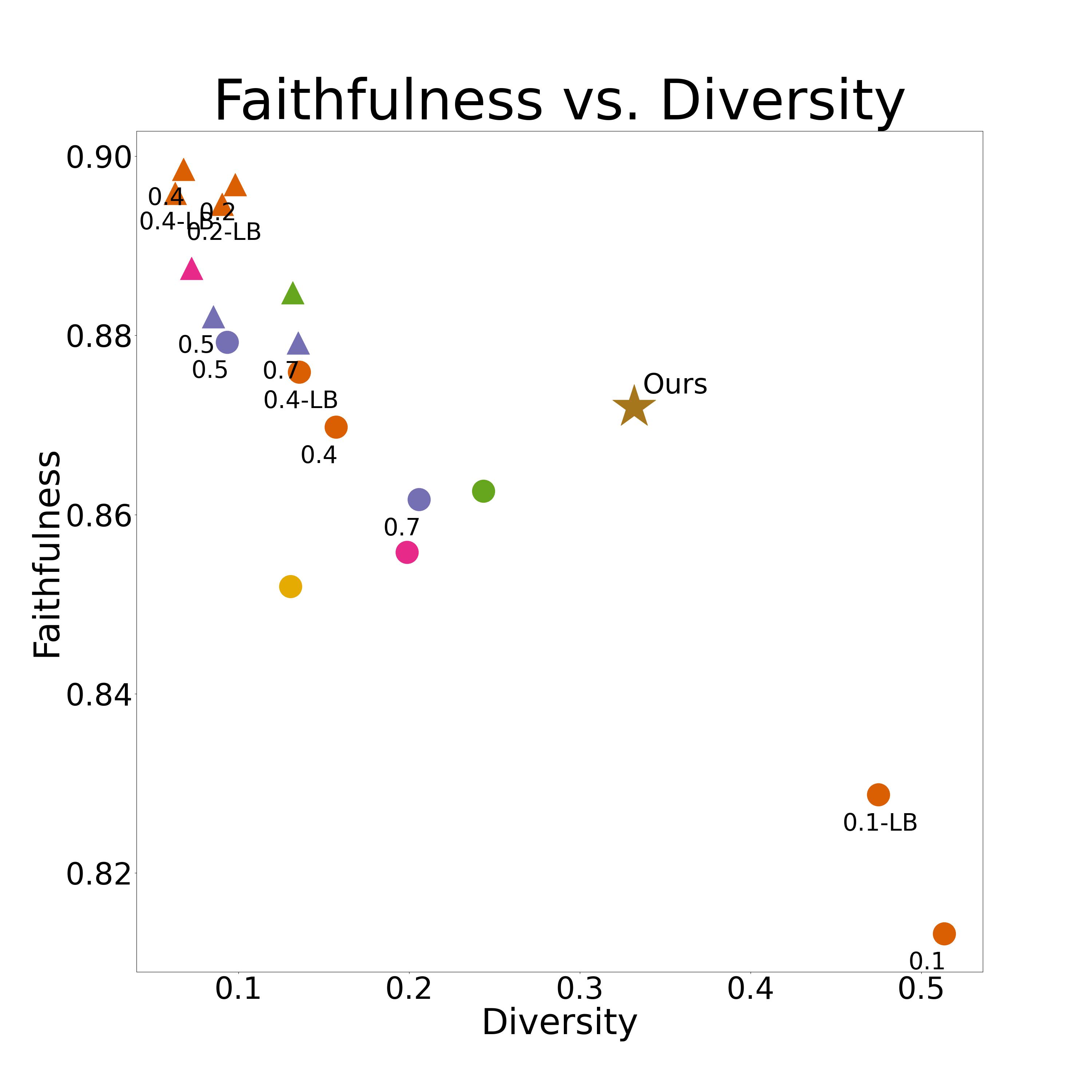} &
        \includegraphics[width=0.5\linewidth]{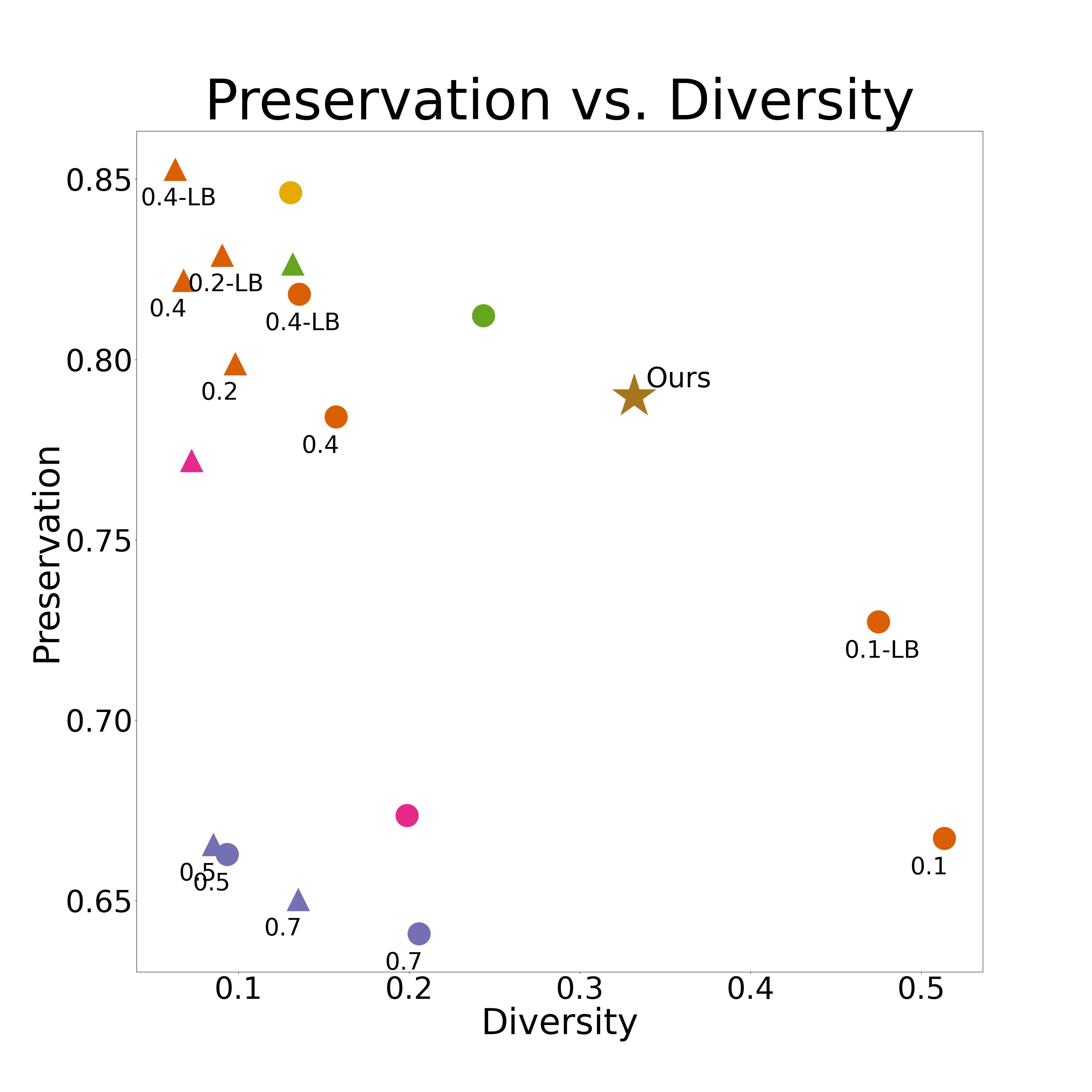} \\
        \multicolumn{2}{c}{\includegraphics[width=\linewidth]{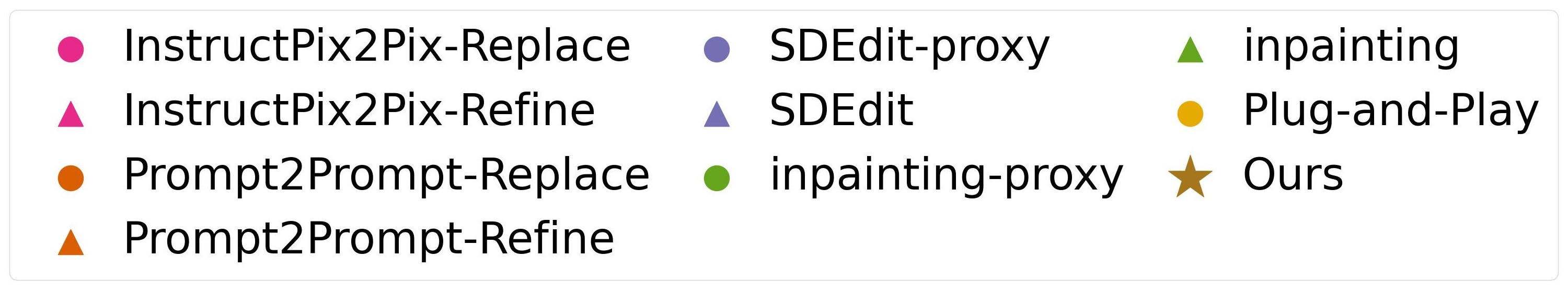}} \\
    \end{tabular}
    \caption{Quantitative comparison with other methods. The graphs above illustrate the trade-off between diversity, faithfulness, and preservation. Refer to Section~\ref{sec:exp-obj-varietions} for more details. }
    \label{fig:quantitative_comparison}
\end{figure}

In Figure~\ref{fig:quantitative_comparison} we present quantitative results. Our method achieves a good trade-off between diversity, faithfulness, and image preservation. 
Methods with higher diversity scores than ours do not preserve the original image and are not faithful to the object of interest, as was also demonstrated in the qualitative comparison. As shown in the graph, methods with better preservation or faithfulness scores than ours, hardly change the shape of the object.

\begin{figure}
       \centering
        \setlength{\tabcolsep}{1pt}
        {\scriptsize
        \begin{tabular}{c c c c c c c }
            { Original } &
            \multicolumn{2}{c}{  } &
            \multicolumn{4}{c}{$\longleftarrow$ Object level variations $\longrightarrow$} \\
            \includegraphics[width=0.185\linewidth]{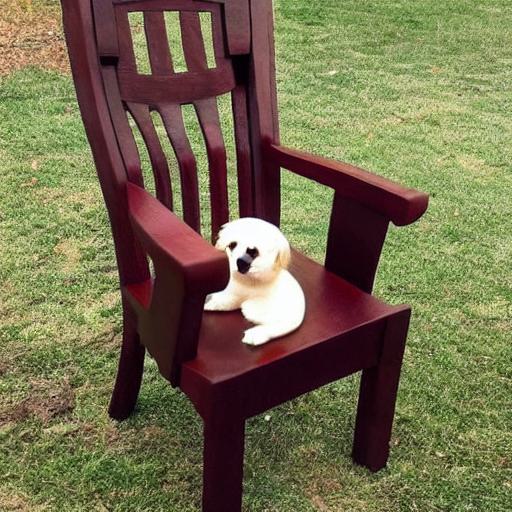} &
            \multicolumn{2}{c}{  } &
            \includegraphics[width=0.185\linewidth]{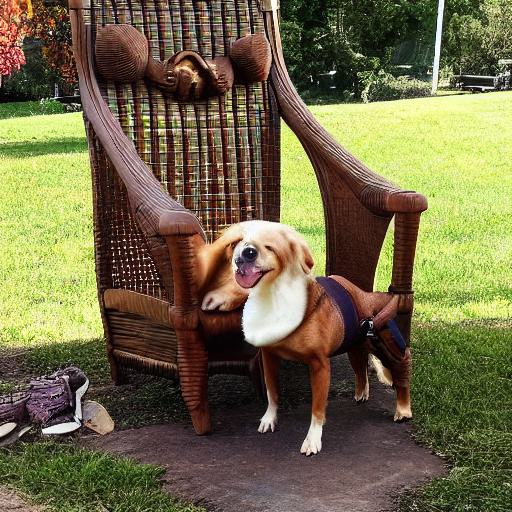} &
            \includegraphics[width=0.185\linewidth]{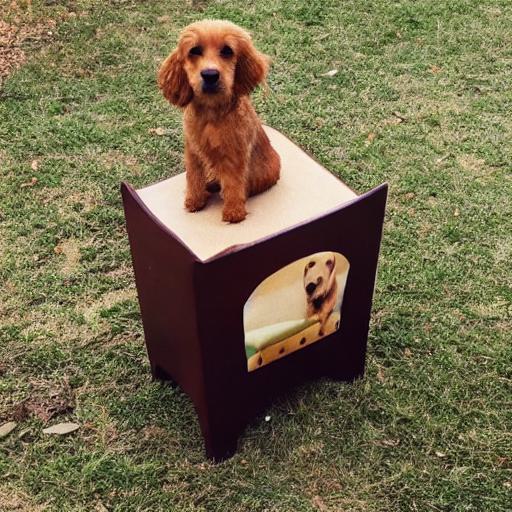} &
            \includegraphics[width=0.185\linewidth]{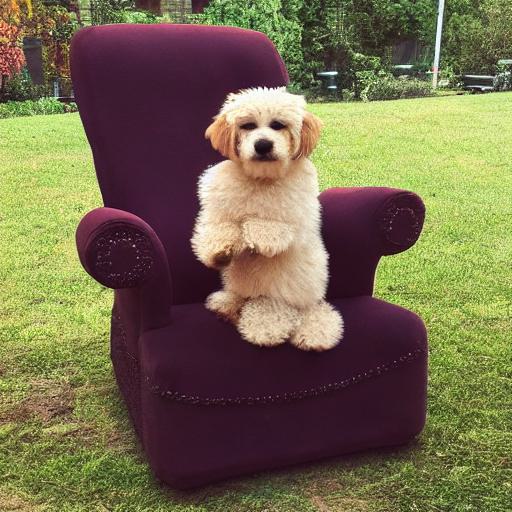} &
            \includegraphics[width=0.185\linewidth]{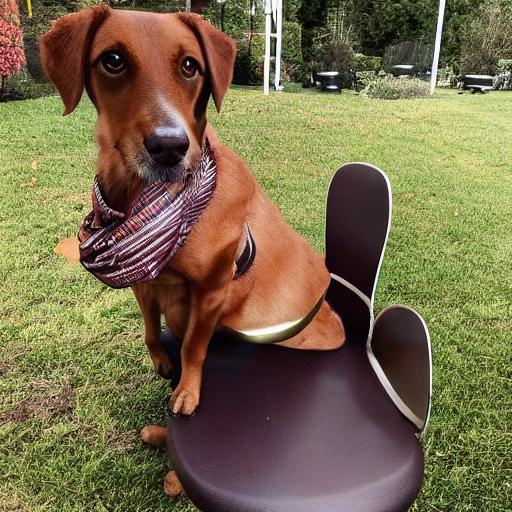} \\
            \multicolumn{3}{c}{  } &
            \multicolumn{4}{c}{ Only Prompt Mixing } \\
            \multicolumn{3}{c}{ } &
            \includegraphics[width=0.185\linewidth]{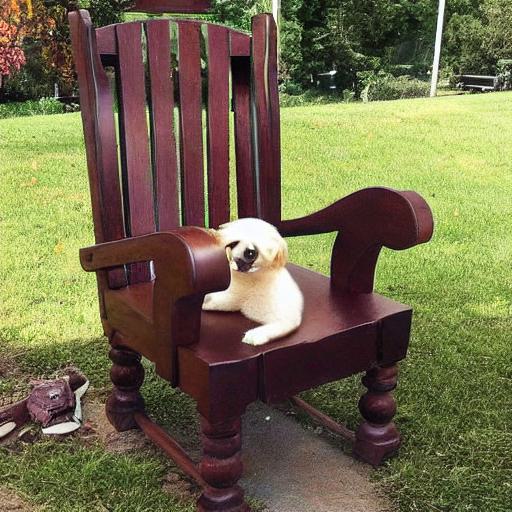} &
            \includegraphics[width=0.185\linewidth]{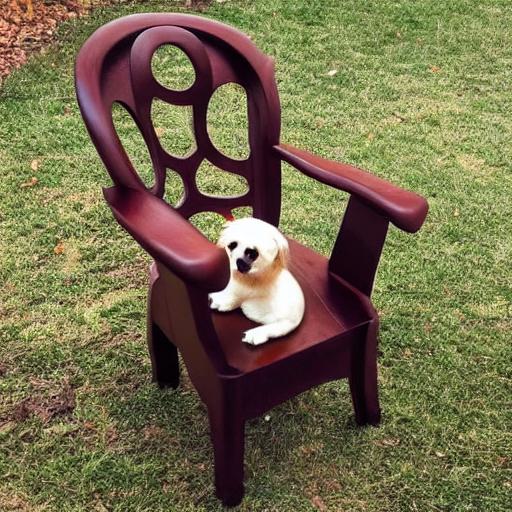} &
            \includegraphics[width=0.185\linewidth]{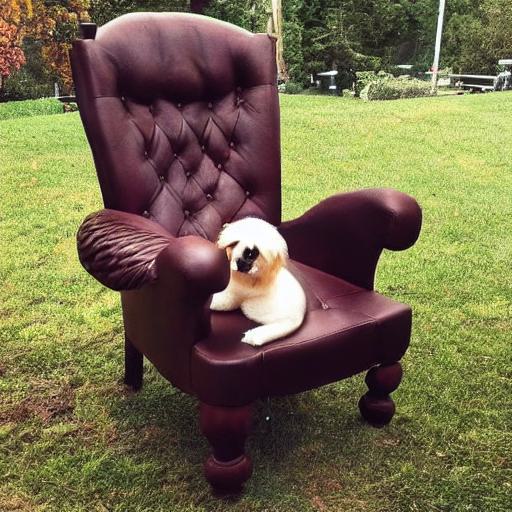} &
            \includegraphics[width=0.185\linewidth]{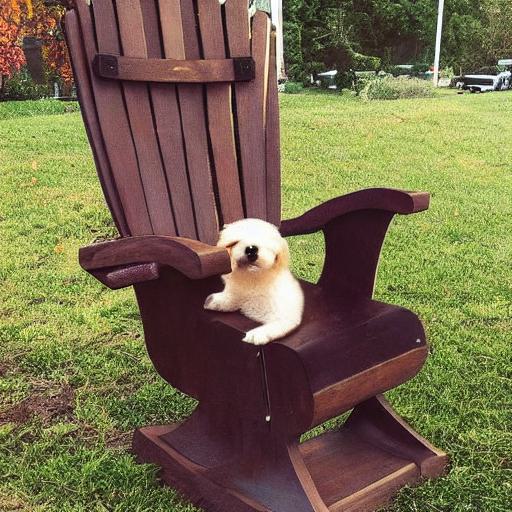} \\
            \multicolumn{3}{c}{  } &
            \multicolumn{4}{c}{ + Attention-Based Shape Localization } \\
            \multicolumn{3}{c}{ } &
            \includegraphics[width=0.185\linewidth]{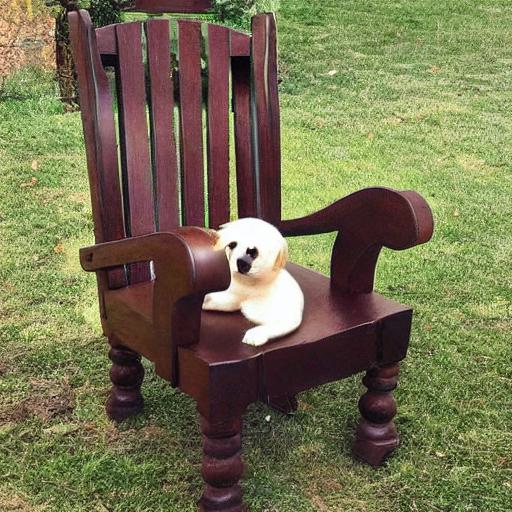} &
            \includegraphics[width=0.185\linewidth]{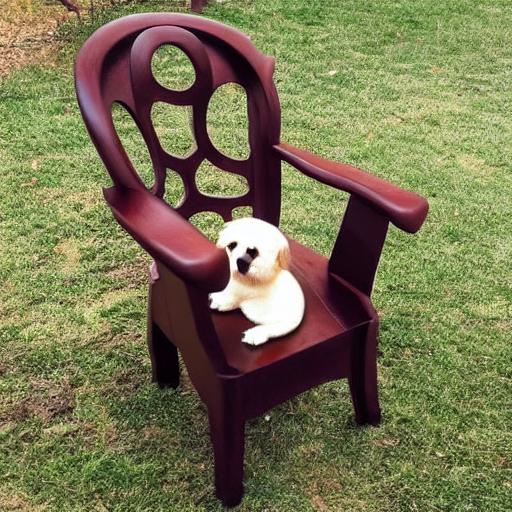} &
            \includegraphics[width=0.185\linewidth]{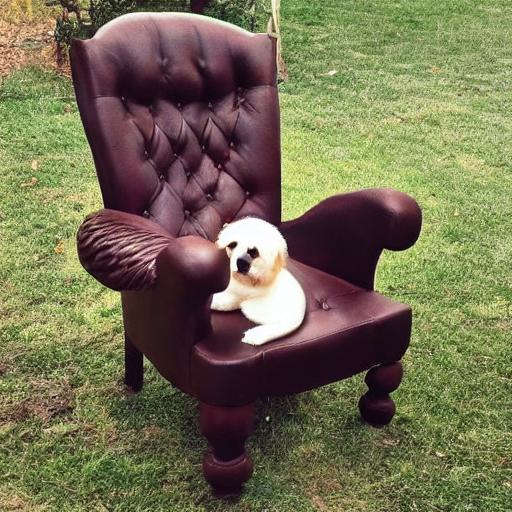} &
            \includegraphics[width=0.185\linewidth]{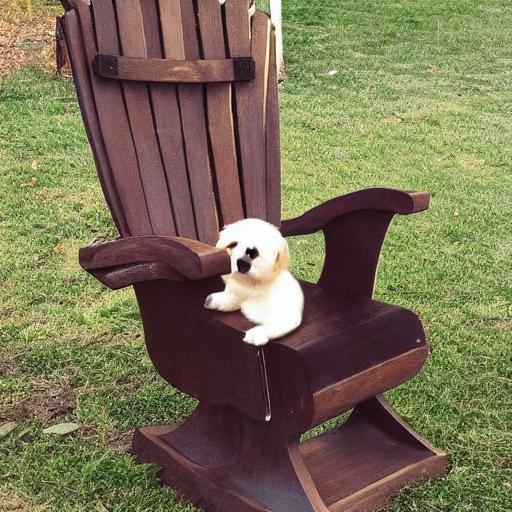} \\
            \multicolumn{3}{c}{  } &
            \multicolumn{4}{c}{ + Controllable Background Preservation } \\
        \end{tabular}
        }
    \vspace{1mm}
    \captionof{figure}{
    Ablating our full object variations pipeline. Original image was crated using the prompt ``A \emph{chair} with a dog on it''. 
    }
    \vspace{-10pt}
    \label{fig:ablation}
\end{figure}

\vspace{-8pt}
\paragraph{Ablation Studies}
We ablate our full pipeline of generating object-level shape variations of a given image, showing the necessity of each part. We present the results in Figure~\ref{fig:ablation}. 
In the first row, we show the results of Mix-and-Match without the localization techniques. As can be seen, Mix-and-Match alone fails to preserve the dog and the background.
Adding the attention-based shape localization technique, where we create a mask for the self-attention map based on the word ``dog'', allows for the preservation of the dog. 
Finally, adding the controllable background preservation technique keeps the background of the original image.

\subsection{Edit Localization}
We integrate our localization techniques with existing text-to-image methods to show improved results when using these methods in conjunction with our techniques.

\begin{figure}
        \setlength{\tabcolsep}{0.5pt}
        {\scriptsize
        \begin{tabular}{c c c c }
            \multicolumn{4}{c}{``a dog with a hat in the park'' $\rightarrow$ ``a dog with a crown in the park''} \\
            \includegraphics[width=0.24\linewidth]{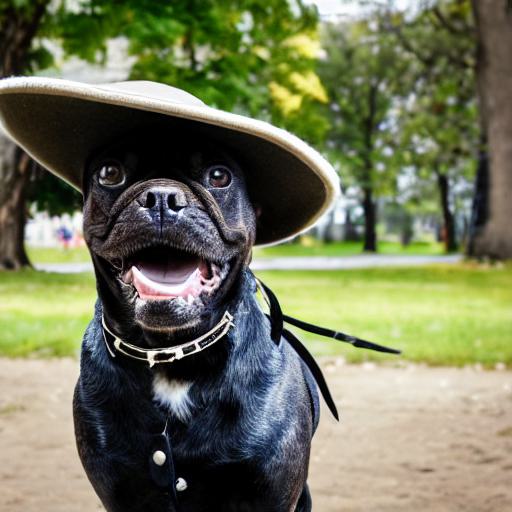} &
            \includegraphics[width=0.24\linewidth]{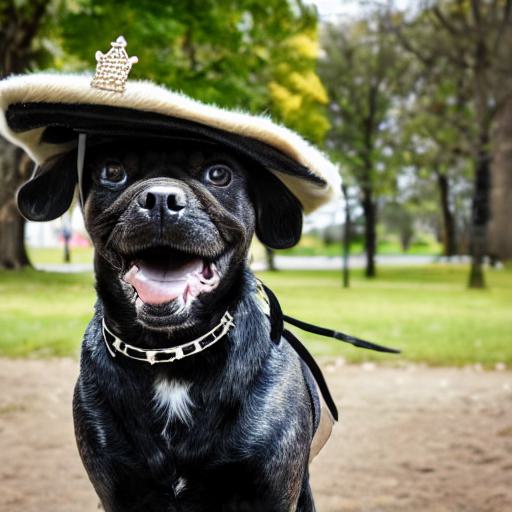} &
            \includegraphics[width=0.24\linewidth]{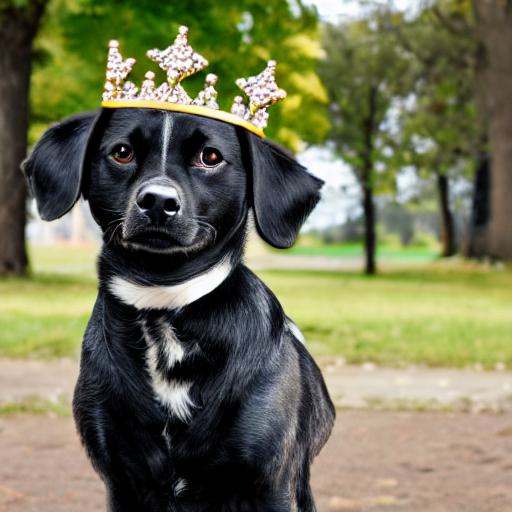} &
            \includegraphics[width=0.24\linewidth]{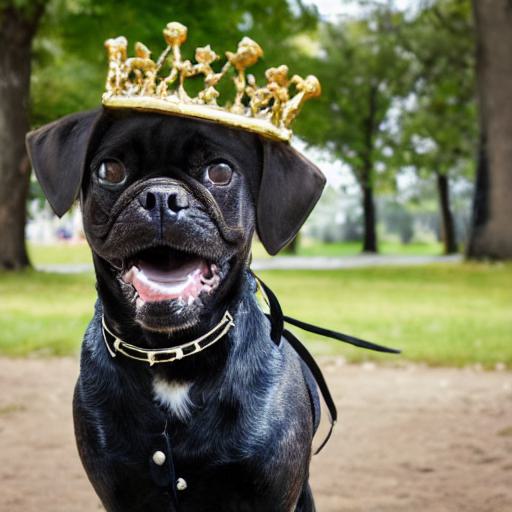} \\
            { \footnotesize  Original } &
            { \footnotesize  P2P w/ } &
            { \footnotesize  P2P w/} &
            { \footnotesize  P2P w/ our } \\
            { \footnotesize  } &
            { \footnotesize  40\% injection} &
            { \footnotesize  20\% injection} &
            { \footnotesize localization } \\           
        \end{tabular}
        \vspace{2mm}}
    \caption{
    Comparison between P2P, which injects the entire self-attention maps, to P2P integrated with our attention-based shape localization technique.
    As demonstrated above, with our localization method P2P replaces the hat while preserving the dog.
    } 
    \label{fig:local_ablation}
\end{figure}

\vspace{-14pt}
\paragraph{Attention-Based Shape Localization}
Previous methods~\cite{hertz2022prompt, pnpDiffusion2022} have injected the entire self-attention map to preserve the shapes of the original image. To demonstrate the effectiveness of our attention-based shape localization technique, we integrate it with P2P~\cite{hertz2022prompt} and show the results in Figure~\ref{fig:local_ablation} where we aim to change the hat of the dog into a crown. The figure shows the original generated image on the left, followed by P2P where we injected the entire self-attention map during 40\% and 20\% of the denoising steps, and P2P with our localization technique.
As can be seen, using P2P involves a tradeoff between accurately changing the hat into a crown and preserving the original dog. By integrating our method, we were able to selectively inject only the rows and columns that corresponded to the dog, achieving the desired transformation of the hat into a crown while preserving the original shape of the dog.

\begin{figure}
    \centering
    \setlength{\tabcolsep}{1pt}
    {\scriptsize
    \begin{tabular}{c c c c c c c c}
        \raisebox{22pt}{\rotatebox{90}{ PnP}} &
        \raisebox{5pt}{\rotatebox{90}{ ``golden robot'' }} & &
        \includegraphics[width=0.22\linewidth]{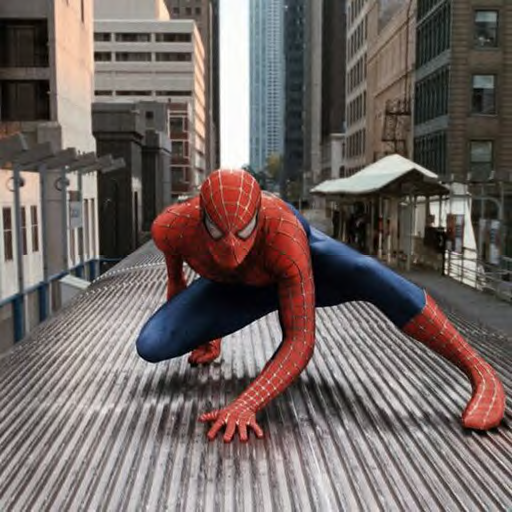} &
        \includegraphics[width=0.22\linewidth]{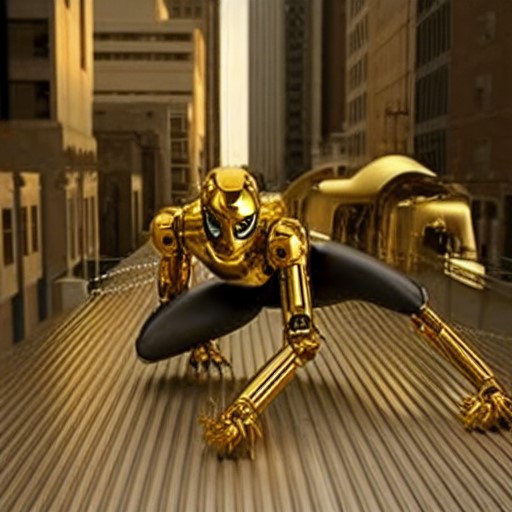} &
        \includegraphics[width=0.22\linewidth]{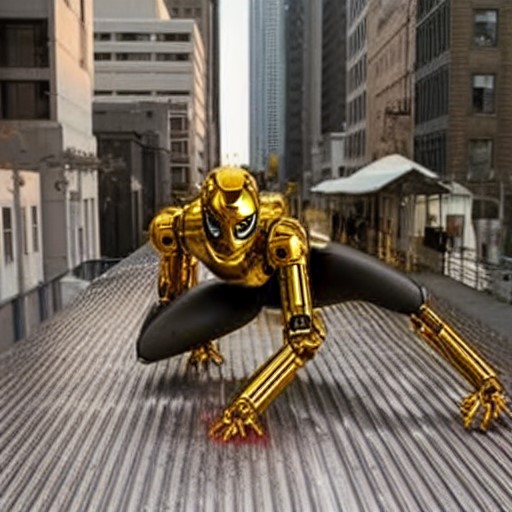} &
        \includegraphics[width=0.22\linewidth]{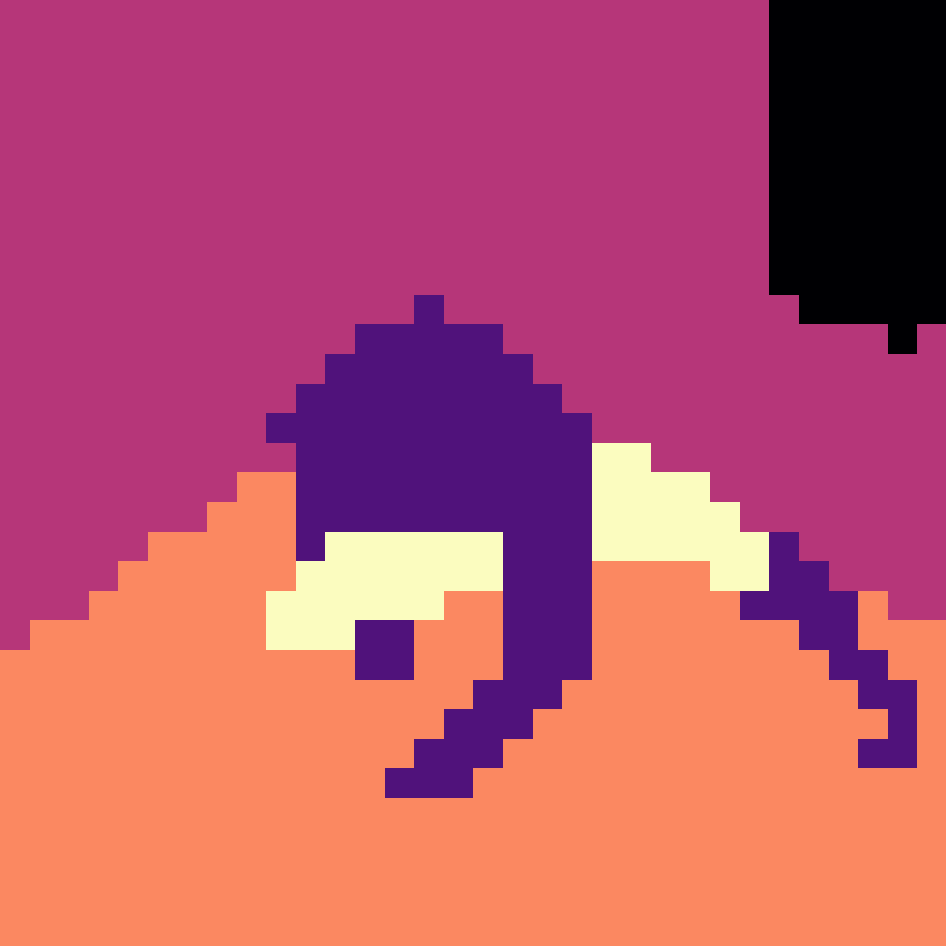} \\
        \raisebox{8pt}{\rotatebox{90}{ P2P (w/ LB)}} &
        \raisebox{3pt}{\rotatebox{90}{ ``cat'' $\rightarrow$ ``bird'' }} & &
        \includegraphics[width=0.22\linewidth]{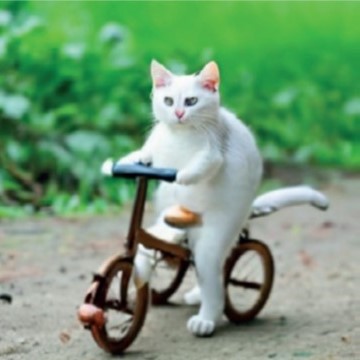} &
        \includegraphics[width=0.22\linewidth]{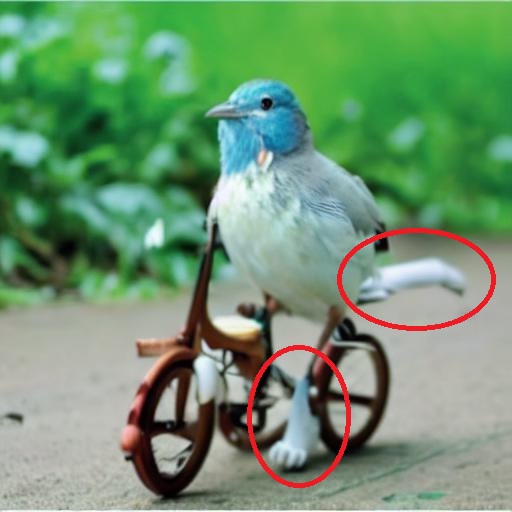} &
        \includegraphics[width=0.22\linewidth]{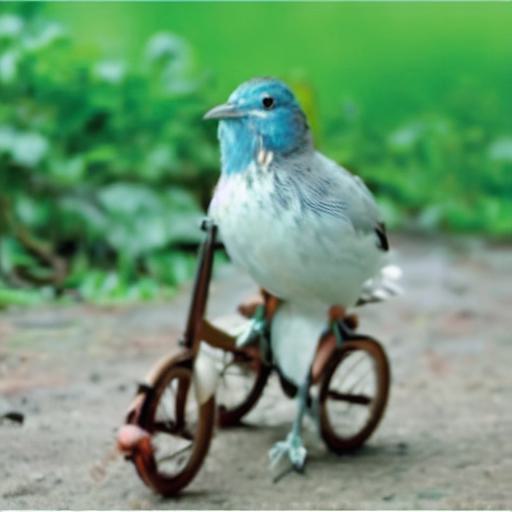} &
        \includegraphics[width=0.22\linewidth]{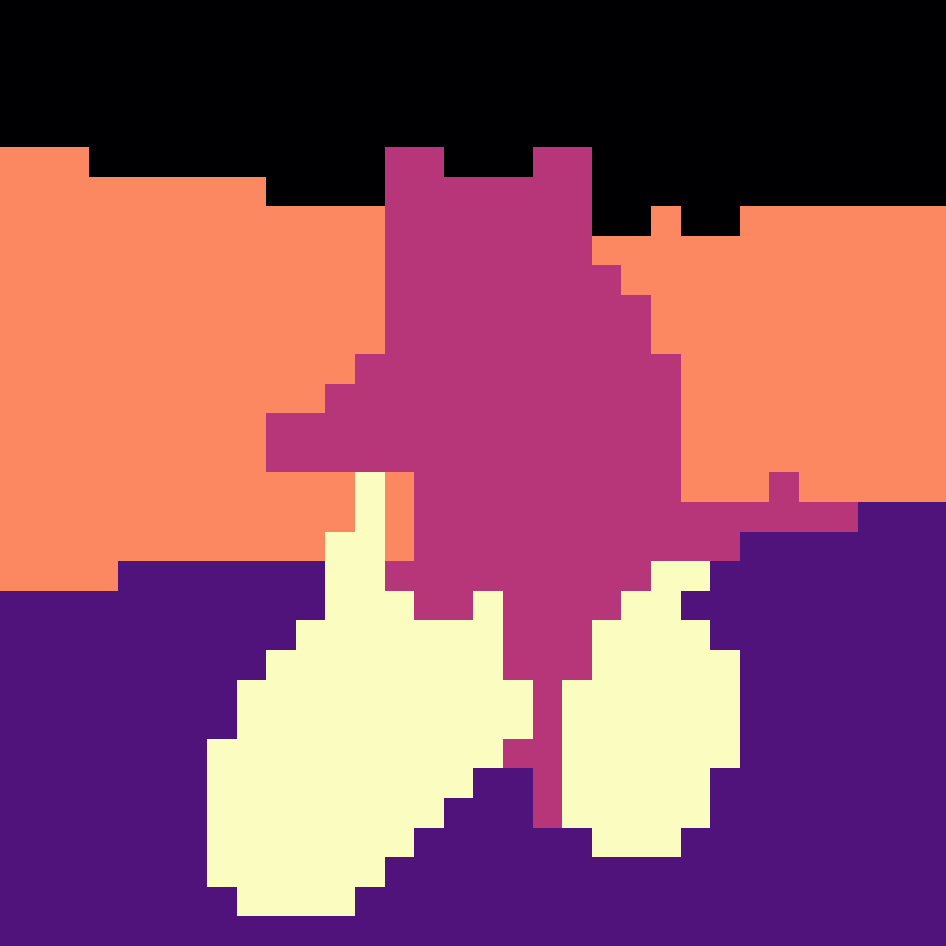} \\
        \raisebox{8pt}{\rotatebox{90}{ SDEdit}} &
        \raisebox{3pt}{\rotatebox{90}{ ``flamingio'' $\rightarrow$ }} &
        \raisebox{3pt}{\rotatebox{90}{ ``cassowary'' }} &
        \includegraphics[width=0.22\linewidth]{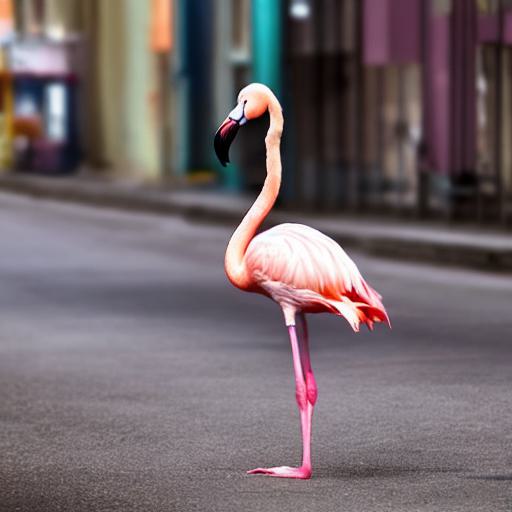} &
        \includegraphics[width=0.22\linewidth]{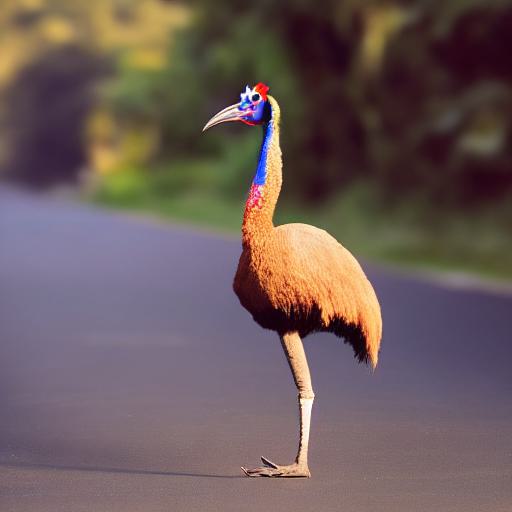} &
        \includegraphics[width=0.22\linewidth]{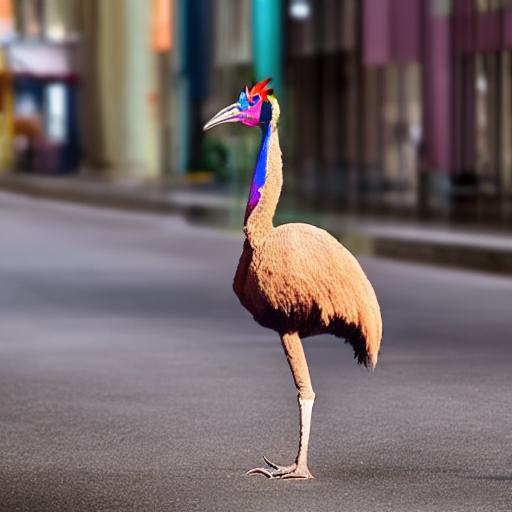} &
        \includegraphics[width=0.22\linewidth]{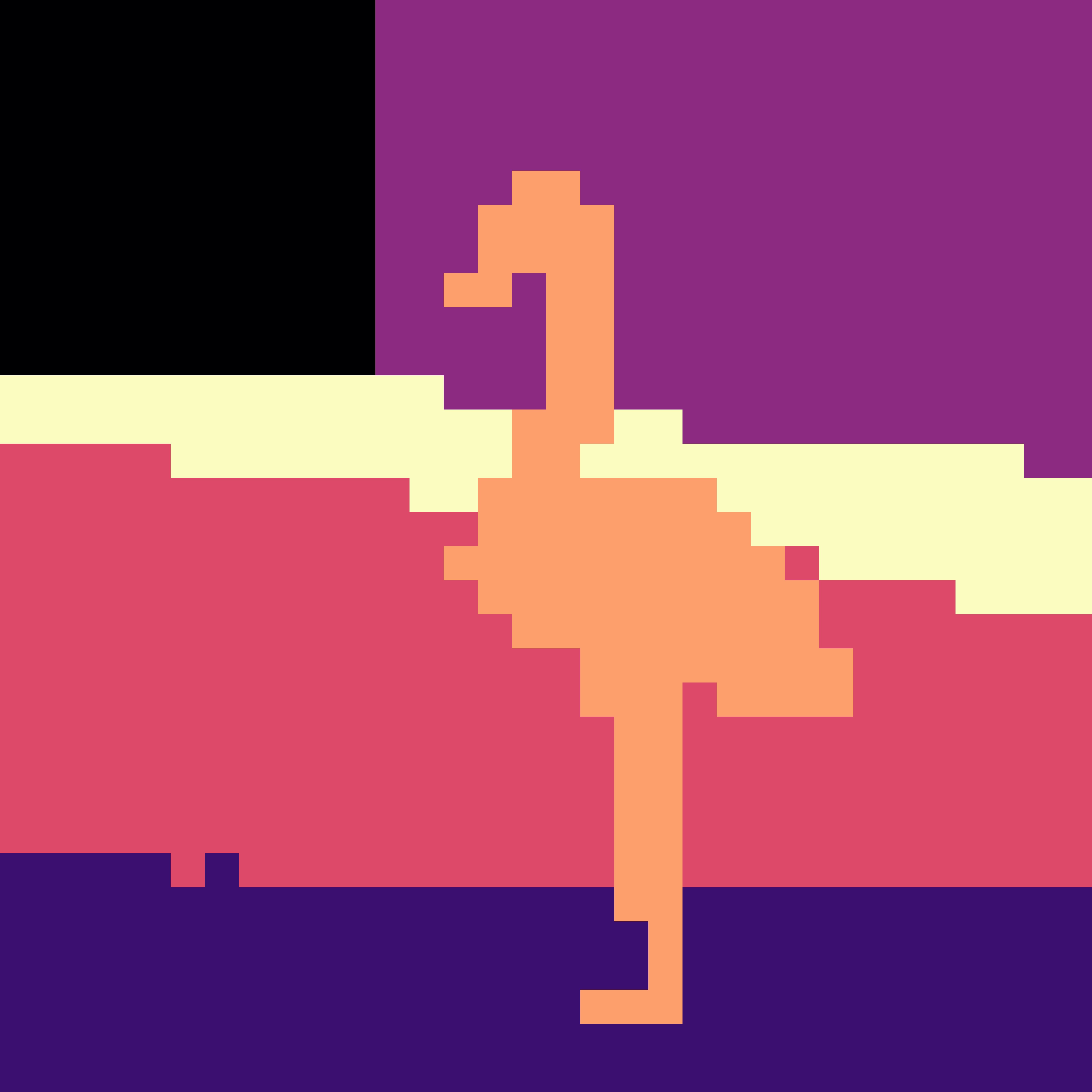} \\
        & & & Input & Edit & $+$ CBP & Segmentation
    
    \end{tabular}
    \vspace{1mm}}
    
    \caption{Our controllable background preservation (CBP) technique integrated with editing methods to localize their edits better. In P2P~\cite{hertz2022prompt} our mask is more accurate as observed by the cat's tail and feet that are added to the bird without CBP.}
    \vspace{-10pt}
    \label{fig:editing_bg}
\end{figure}

\vspace{-10pt}
\paragraph{Controllable Background Preservation}
We test our background preservation technique with PnP~\cite{pnpDiffusion2022}, P2P~\cite{hertz2022prompt}, and SDEdit~\cite{meng2022sdedit} and show the results in Figure~\ref{fig:editing_bg}. 
For P2P we use their local blending.
To integrate each method with our technique, we first invert each image~\cite{mokady2022null}. Then, we segment and label each segment of the image with our technique, and create a mask of the main object in the image. At step $t=35$ of the denoising process we perform blending between the edited image and the original one, taking the background from the original image, and the object from the edited image. 

As can be seen, our method achieves plausible segmentation maps even for inverted images. For PnP and SDEdit, our method allows editing the object while keeping the background as in the original image. In P2P we see that our technique localizes the edit better, removing the cat's tail and feet from the image.

\section{Discussion and Conclusion}

We have presented a method for exploring object-level shape variations in an image, which addresses two main technical challenges: changing the shape of an object and localizing the change. To achieve this, we introduced a Mix-and-Match technique to generate shape variations, and built upon self-attention maps injection to preserve the original image structure, while enabling changes to the object of interest. Furthermore, we demonstrated how the geometric information encoded in self-attention maps can be used for image segmentation, which allows for guiding the preservation of the background. While our method produces plausible variations of an image, we acknowledge that automatic proxy-words may fail at times. In the future, we would like to develop means to explore a continuous words space rather than the current discrete one.

\section*{Acknowledgement}
We thank Ron Mokady and Matan Cohen for their early feedback and helpful suggestions. 
We thank Shiran Zada for assistance with Imagen comparisons.
We also want to give a special thanks Yuval Alaluf and Rinon Gal for their comments and discussions throughout the writing of this paper.
This work was supported by a research gift from Meta, by the Deutsch Foundation and the Yandex Initiative in AI, and by the Israel Science Foundation under Grant No. 2492/20.

{\small
\bibliographystyle{ieee_fullname}
\bibliography{egbib}
}

\begin{appendices}
        \section{Additional Details }

\subsection{Prompt-Mixing}

\begin{figure}[h!]
    \centering
    \setlength{\tabcolsep}{0pt}
    {\scriptsize
    \begin{tabular}{ccc c c c c}
        \multicolumn{3}{c}{Single Prompt} &
        & 
        \multicolumn{3}{c}{Prompt-Mixing} \\
    
        \includegraphics[width=0.19\linewidth]{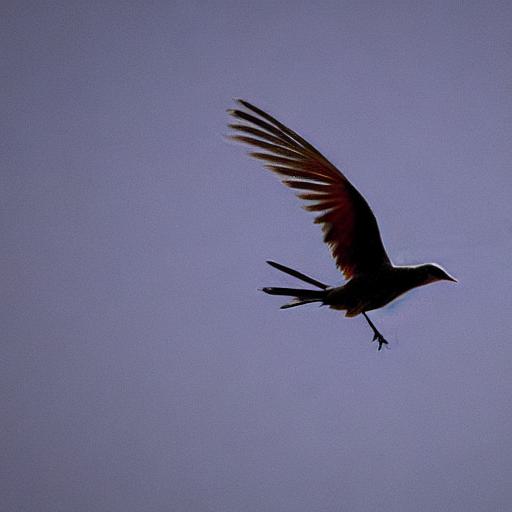} &
        \includegraphics[width=0.19\linewidth]{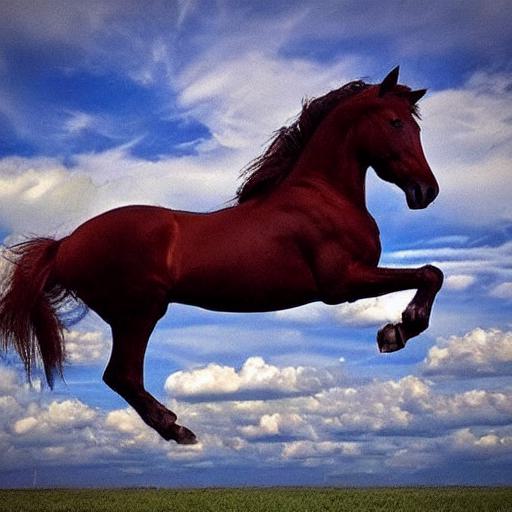} &
        \includegraphics[width=0.19\linewidth]{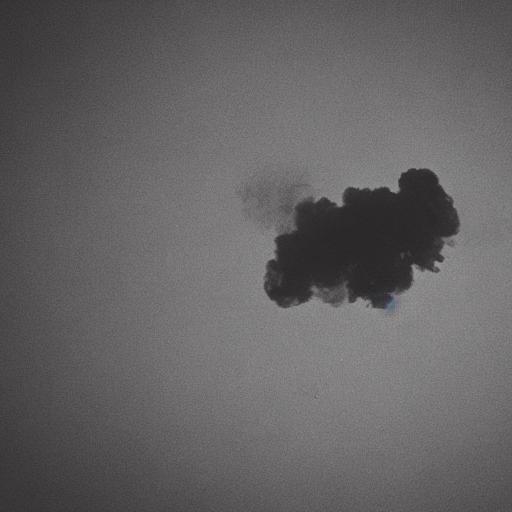} &
        { } &
        \includegraphics[width=0.19\linewidth]{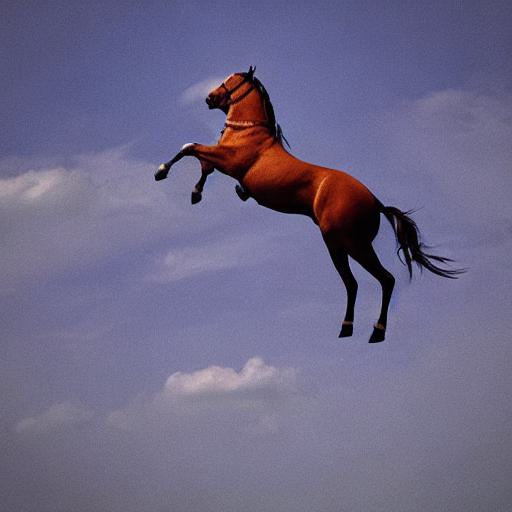} &
        { } &
        \includegraphics[width=0.19\linewidth]{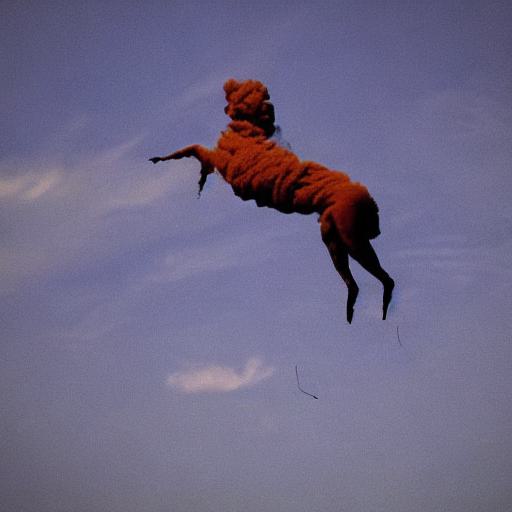} \\
        \includegraphics[width=0.19\linewidth]{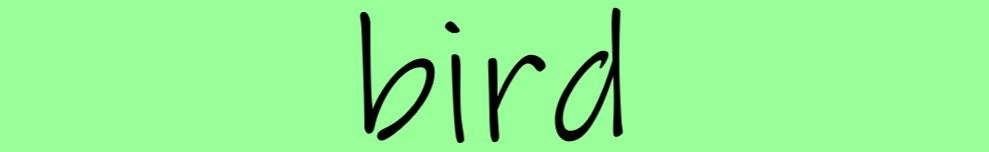} &
        \includegraphics[width=0.19\linewidth]{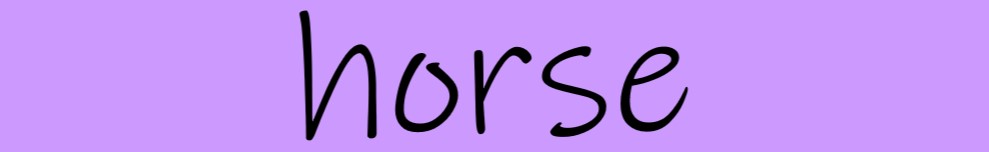} &
        \includegraphics[width=0.19\linewidth]{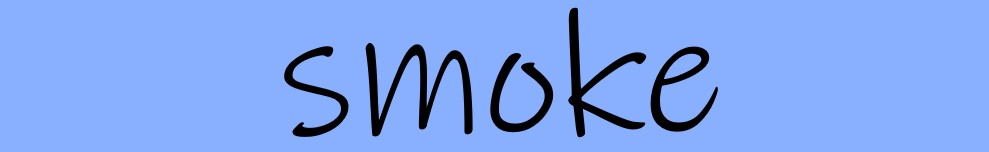} &
        { } &
        \includegraphics[width=0.19\linewidth]{images/analysis/balls_pyramids_bar.jpg} &
        {}&
        \includegraphics[width=0.19\linewidth]{images/analysis/balls_pyramids_fluffies_bar.jpg} \\
        \includegraphics[width=0.19\linewidth]{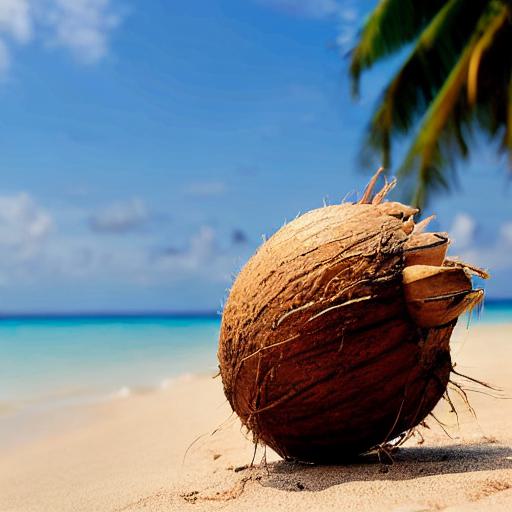} &
        \includegraphics[width=0.19\linewidth]{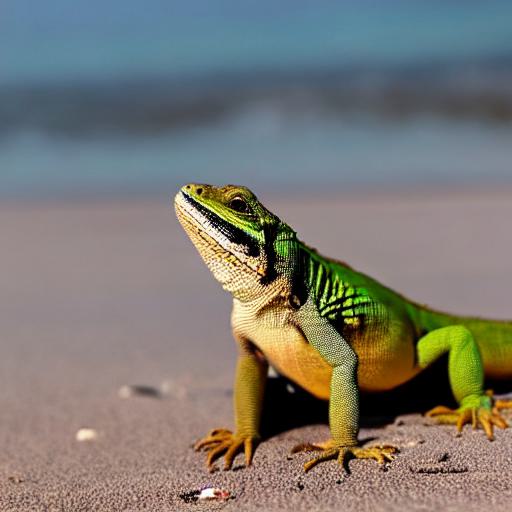} &
        \includegraphics[width=0.19\linewidth]{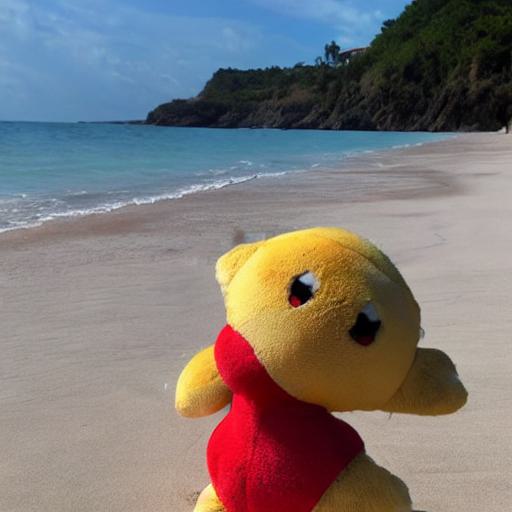} &
        { } &
        \includegraphics[width=0.19\linewidth]{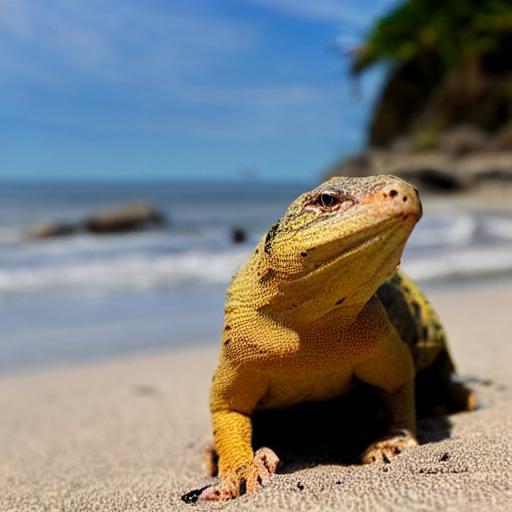} &
        {} &
        \includegraphics[width=0.19\linewidth]{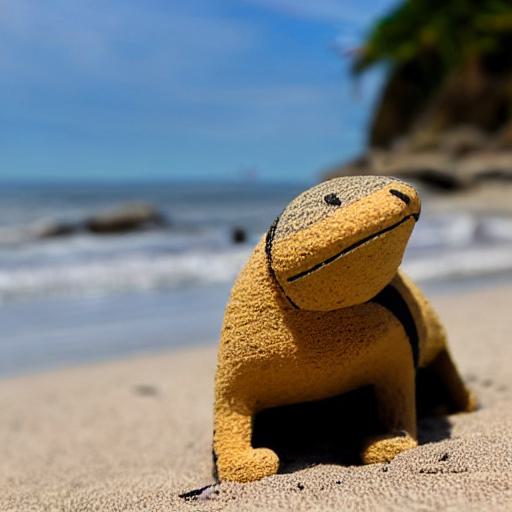} \\
        \includegraphics[width=0.19\linewidth]{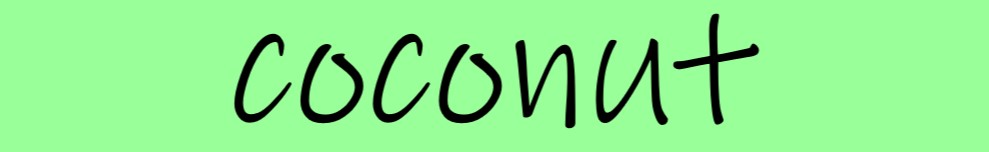} &
        \includegraphics[width=0.19\linewidth]{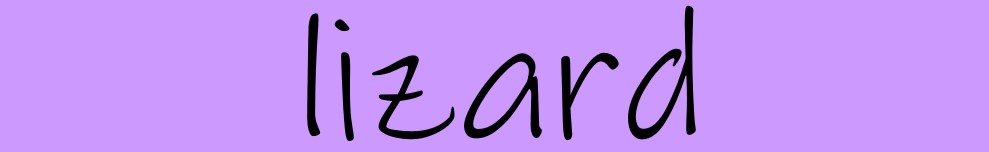} &
        \includegraphics[width=0.19\linewidth]{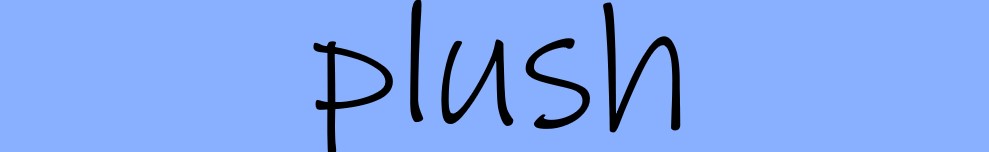} &
        { } &
        \includegraphics[width=0.19\linewidth]{images/analysis/balls_pyramids_bar.jpg} &
        {} &
        \includegraphics[width=0.19\linewidth]{images/analysis/balls_pyramids_fluffies_bar.jpg} 
    \end{tabular}
    }
    
    \caption{
    Prompt-mixing. The colored bars under the images on the right represent the corresponding
word used along the denoising process. This figure complements Figure 4 in the main paper.
    }
    \vspace{-8pt}
    \label{fig:analysis-stages-supp}
\end{figure}

\paragraph{Denoising Diffusion Process Stages}
Additional examples of the denoising stages are shown in Figure~\ref{fig:analysis-stages-supp}. The used prompts are ``A $\left<w\right>$ is flying in the sky'' (first row), and ``A $\left<w\right>$ on the beach'' (second row).

\vspace{-8pt}

\paragraph{Cross-Attention Injection} \label{sec:cross-attention-injection}

\begin{figure}
    \centering
    \setlength{\tabcolsep}{0pt}
    {\scriptsize

    \begin{tabular}{c c cc cc cc}
        \includegraphics[width=0.19\linewidth]{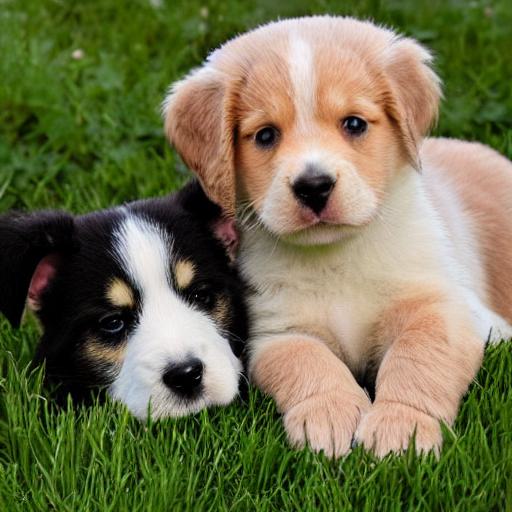} &
        \includegraphics[width=0.19\linewidth]{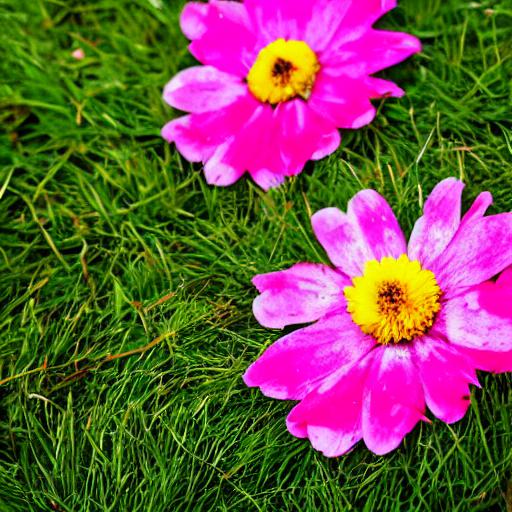} &
        { } &
        \includegraphics[width=0.19\linewidth]{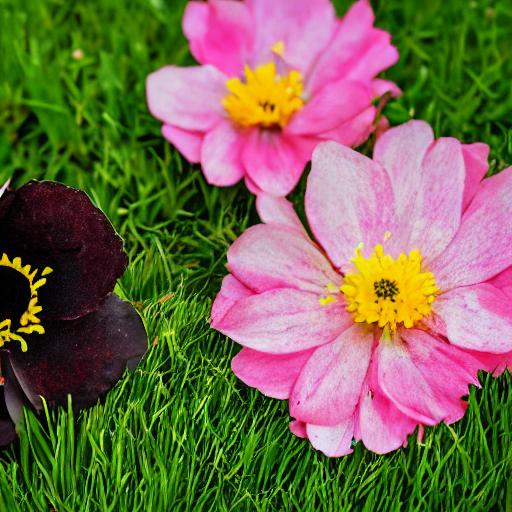} &
        { } &
        \includegraphics[width=0.19\linewidth]{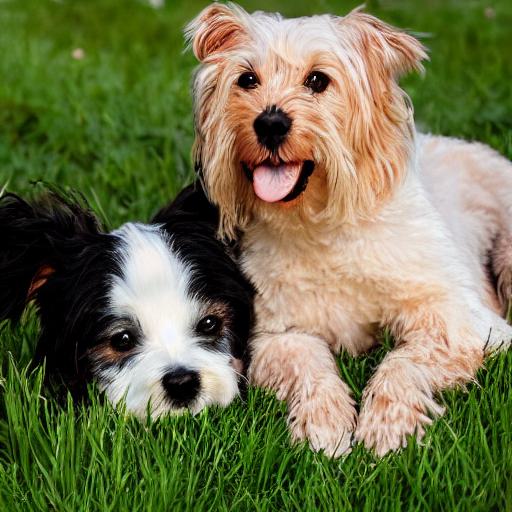} &
        { } &
        \includegraphics[width=0.19\linewidth]{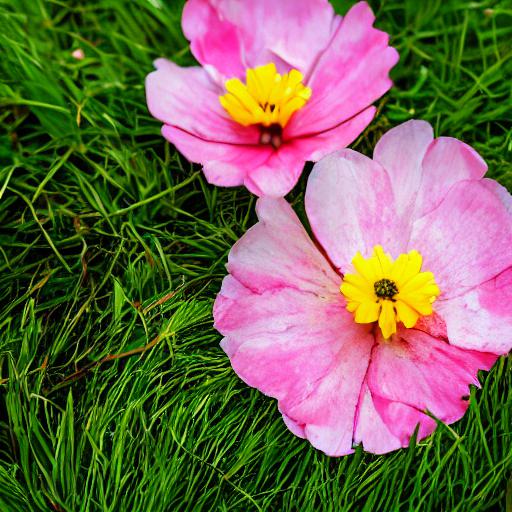} \\
        \includegraphics[width=0.19\linewidth]{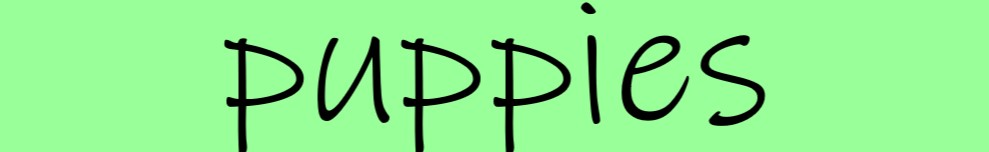} &
        \includegraphics[width=0.19\linewidth]{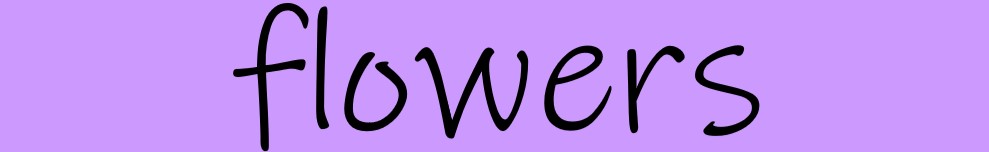} &
        { } &
        \includegraphics[width=0.19\linewidth]{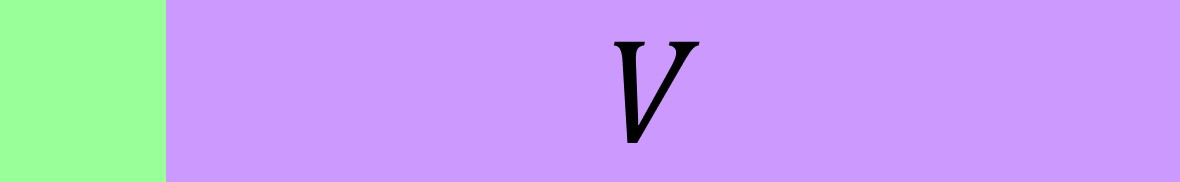} &
        { } &
        \includegraphics[width=0.19\linewidth]{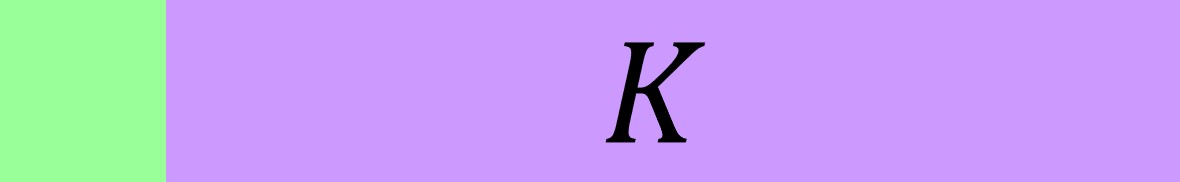} &
        { } &
        \includegraphics[width=0.19\linewidth]{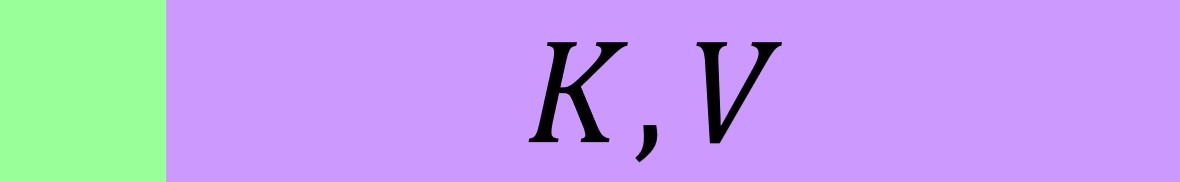} \\
        \includegraphics[width=0.19\linewidth]{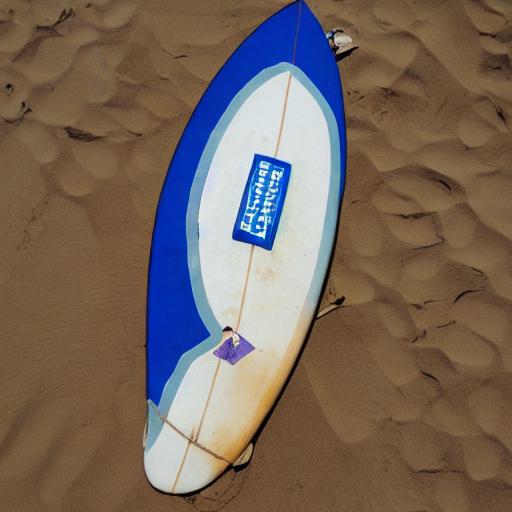} &
        \includegraphics[width=0.19\linewidth]{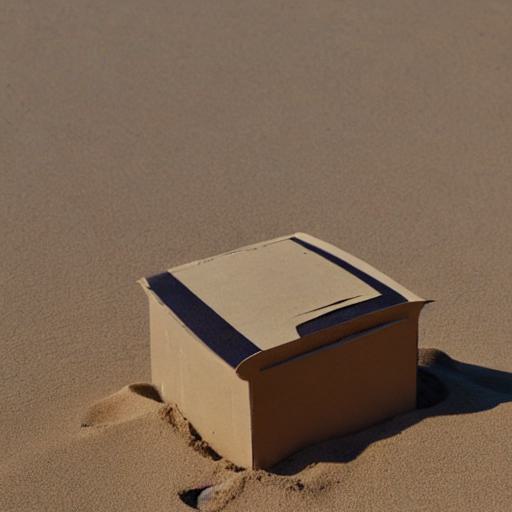} &
        { } &
        \includegraphics[width=0.19\linewidth]{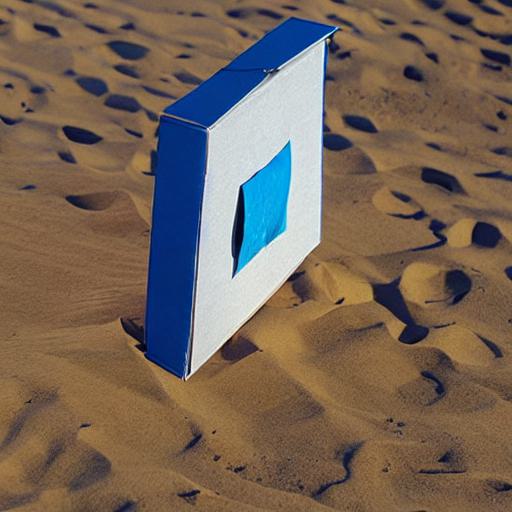} &
        { } &
        \includegraphics[width=0.19\linewidth]{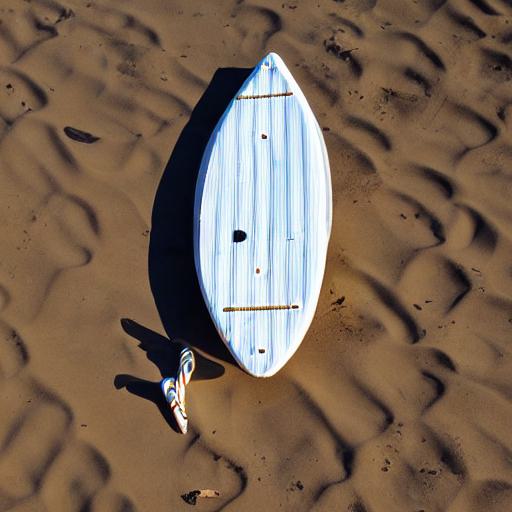} &
        { } &
        \includegraphics[width=0.19\linewidth]{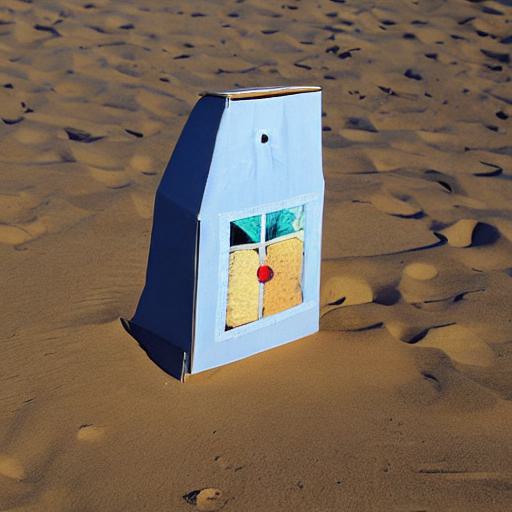} \\
        \includegraphics[width=0.19\linewidth]{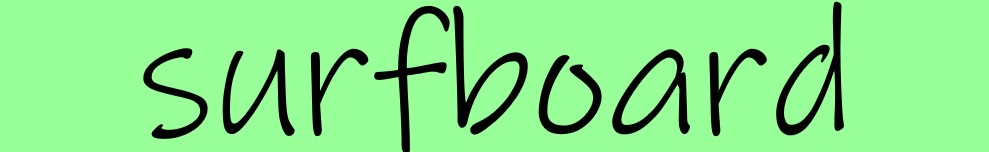} &
        \includegraphics[width=0.19\linewidth]{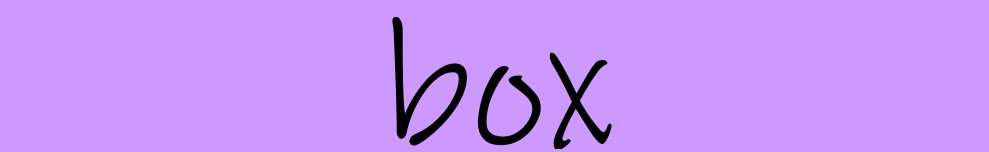} &
        { } &
        \includegraphics[width=0.19\linewidth]{images/analysis/V_bar.jpg} &
        { } &
        \includegraphics[width=0.19\linewidth]{images/analysis/K_bar.jpg} &
        { } &
        \includegraphics[width=0.19\linewidth]{images/analysis/KV_bar.jpg} \\
        \includegraphics[width=0.19\linewidth]{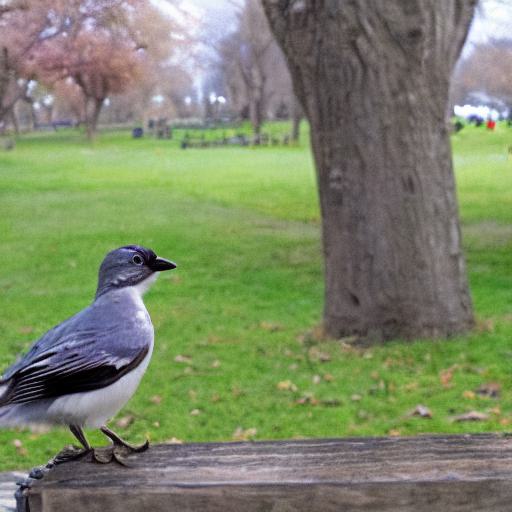} &
        \includegraphics[width=0.19\linewidth]{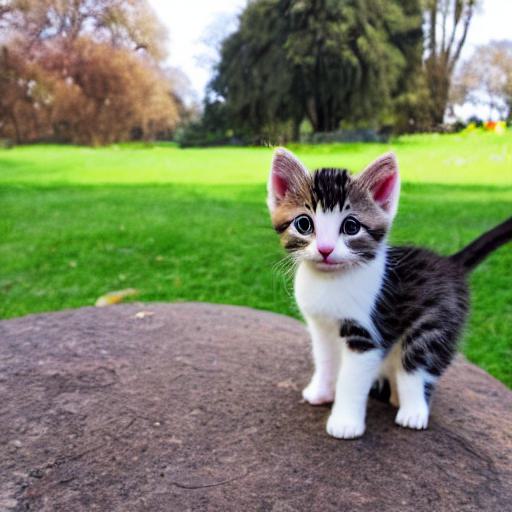} &
        { } &
        \includegraphics[width=0.19\linewidth]{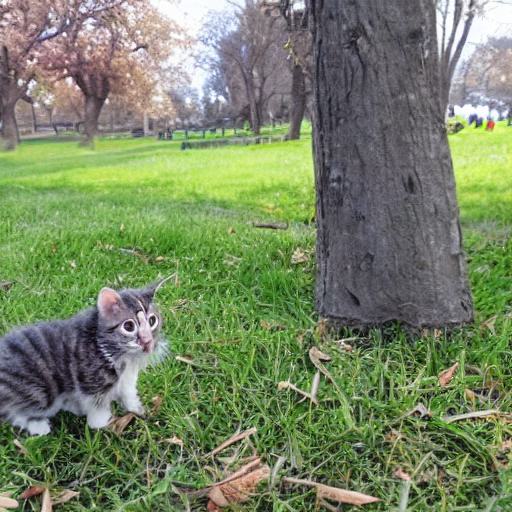} &
        { } &
        \includegraphics[width=0.19\linewidth]{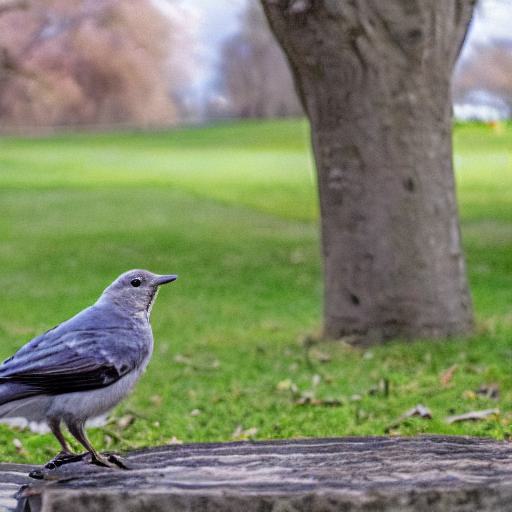} &
        { } &
        \includegraphics[width=0.19\linewidth]{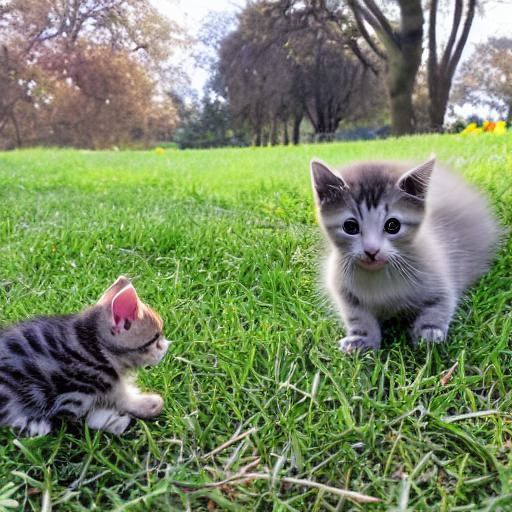} \\
        \includegraphics[width=0.19\linewidth]{images/analysis/supp/bird_.jpg} &
        \includegraphics[width=0.19\linewidth]{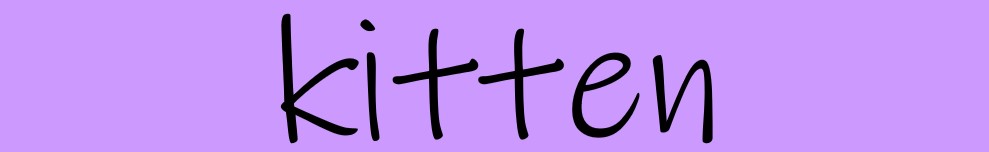} &
        { } &
        \includegraphics[width=0.19\linewidth]{images/analysis/V_bar.jpg} &
        { } &
        \includegraphics[width=0.19\linewidth]{images/analysis/K_bar.jpg} & 
        { } &
        \includegraphics[width=0.19\linewidth]{images/analysis/KV_bar.jpg} \\
        
    \end{tabular}
    }
    \vspace{1.5mm}
    \caption{
    Cross-attention injection. The bars represent the word used along the denoising process. 
    During the purple interval, we use the modified prompt (\ie, with the ``purple'' word) to compute the cross-attention components ($K$, $V$) indicated over the purple bar. 
    We used the prompts: ``Two $\left<w\right>$ on grass'' (first row), ``A $\left<w\right>$ on sand'' (second row), and ``A $\left<w\right>$ in the park'' (third row).
    }
    \label{fig:analysis-kv}
\end{figure}

We validate the intuition about the role of the Keys and Values in the cross-attention layers, as mentioned in the main paper. This experiment explains the design choice of modifying only the Values with the altered prompt when applying prompt-mixing.
We analyze prompt-mixing where the altered prompt is used to compute a different number of components of the cross-attention layer. In Figure~\ref{fig:analysis-kv}, the first two columns correspond to the generation of an image with a single prompt $P(w)$ along the entire denoising process. 
In the other columns, we perform prompt-mixing where we use $P_{[T, T_3]}({\color{ao}{w_1}})$ and $P_{[T_3, 0]}({\color{magenta}{w_2}})$.
In other words, we take the layout from {\color{ao}{$w_1$}} (``puppies'', ``surfborad'', ``bird''), and the shape and fine visual details from {\color{magenta}{$w_2$}} (``flowers'', ``box'', ``kitten''). 

In the third column, we use $P({\color{magenta}{w_2}})$ to compute the Values and $P({\color{ao}{w_1}})$ to compute the Keys in $[T_3, 0]$,
the layouts of the obtained images are similar to these obtained by the images generated by $P({\color{ao}{w_1}})$ in the first column. For example, in the first row, a black flower is added to capture the black puppy.
In the fourth column, where we use $P({\color{magenta}{w_2}})$ to compute the Keys and $P({\color{ao}{w_1}})$ to compute the Values, the appearance of the objects is still of the objects generated by $P({\color{ao}{w_1}})$ in the first column. This strengthens our intuition that the Values of cross-attention layers control object shape and appearance while the Keys hardly affect it. 
In the fifth column, we use $P({\color{magenta}{w_2}})$ to compute both the Keys and the Values in $[T_3, 0]$. As can be seen, the images in the fifth column capture the layout of the images generated by $P({\color{ao}{w_1}})$ worse than the images in the third column. For example, in the first row there are two flowers and the image looks similar to the one generated solely by $P({\color{magenta}{w_2}})$ in the second column. Additionally, in the third row there are two kittens, one instead of the bird and the other at the same place as the kitten in the image generated solely by $P({\color{magenta}{w_2}})$.
Therefore, when applying prompt mixing, the modified prompts are used to compute only the Values of the cross-attention layers in the corresponding timestep interval.

\paragraph{Proxy Words} \label{sec:proxy_words}
We utilize proxy words as a technique for navigating through the prompt embedding space in a meaningful manner. Another potential method is to add noise to the prompt embedding or the cross-attention Values, but we have observed that this approach performs significantly worse and results in lower diversity. Our experimental findings indicate that our proxy-word selection method occasionally produces unexpected words that generate convincing variations in shape. For instance, the word ``something'' frequently appears as a proxy word in our examples. Additionally, we have discovered that certain words with a significant CLIP distance from the object of interest can still generate reasonable variations. However, we opt to use our simple ranking mechanism that avoids more intricate selections for words (considering for instance word concreteness) as we find that it enhances the likelihood of producing plausible variations. 

\begin{figure*}
	\centering
	\setlength{\tabcolsep}{0pt}
		\begin{tabular}{cc c c c ccc c c}
			 & \multicolumn{3}{c}{$\longleftarrow$ close words$\longrightarrow$} & \multicolumn{3}{c}{$\longleftarrow$ out of context words$\longrightarrow$} \\
			Original &
			``bag''&
			``wallet'' &
			``carry'' &
												          
			``gift'' &
			``butterfly'' &
			``lamp'' \\
			\includegraphics[width=0.12\textwidth]{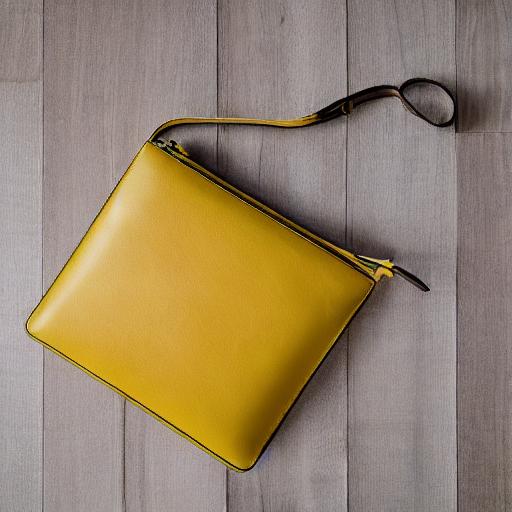} & 
			\includegraphics[width=0.12\textwidth]{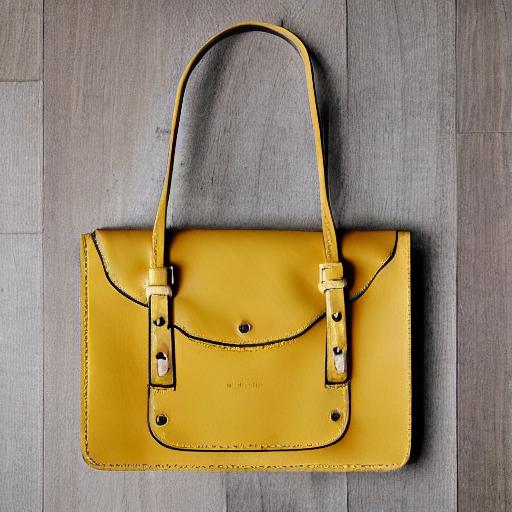} & 
			\includegraphics[width=0.12\textwidth]{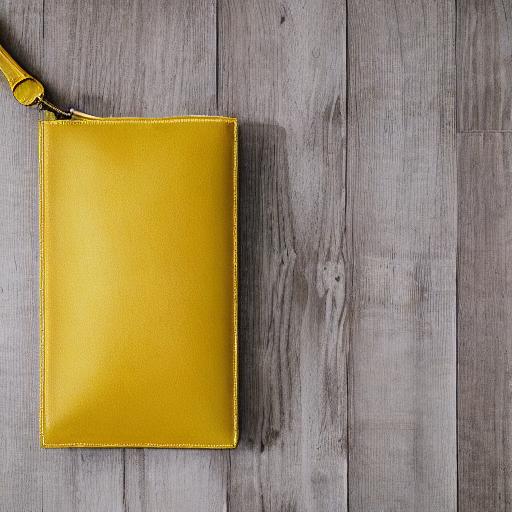} & 
			\includegraphics[width=0.12\textwidth]{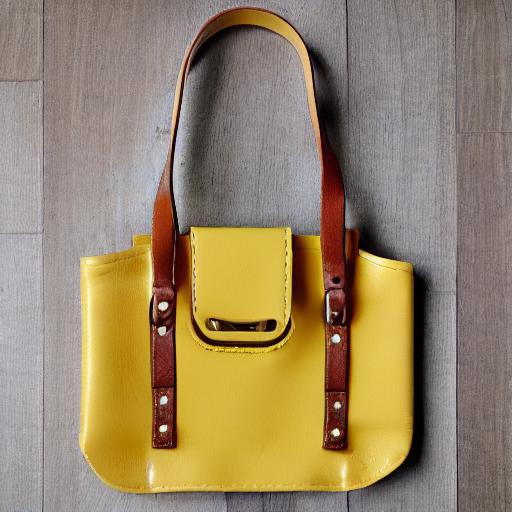} { } { }&
			\includegraphics[width=0.12\textwidth]{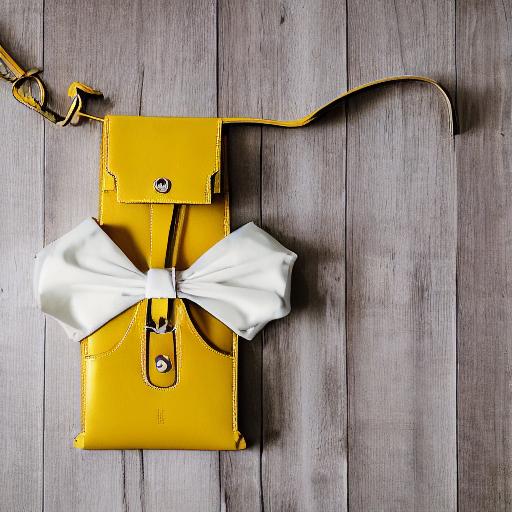} & 
			\includegraphics[width=0.12\textwidth]{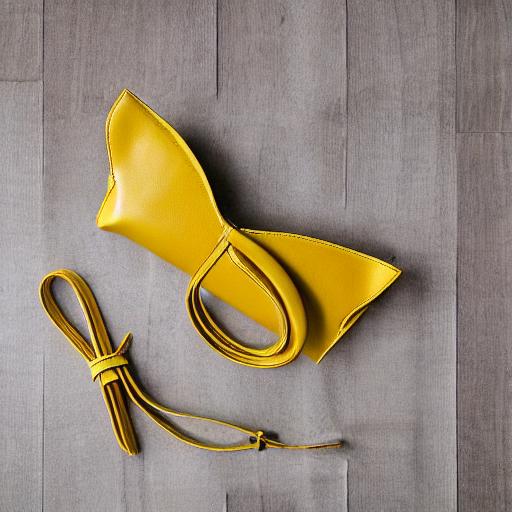} & 
			\includegraphics[width=0.12\textwidth]{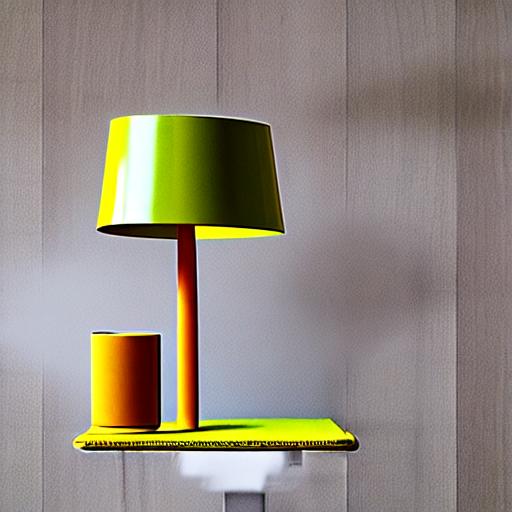} & 
			\\		
			\includegraphics[width=0.12\textwidth]{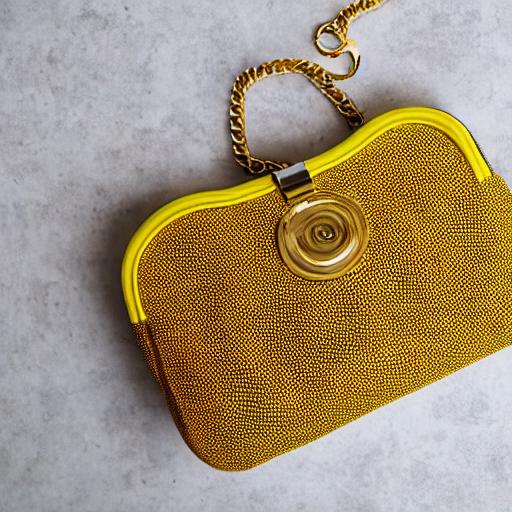} & 
			\includegraphics[width=0.12\textwidth]{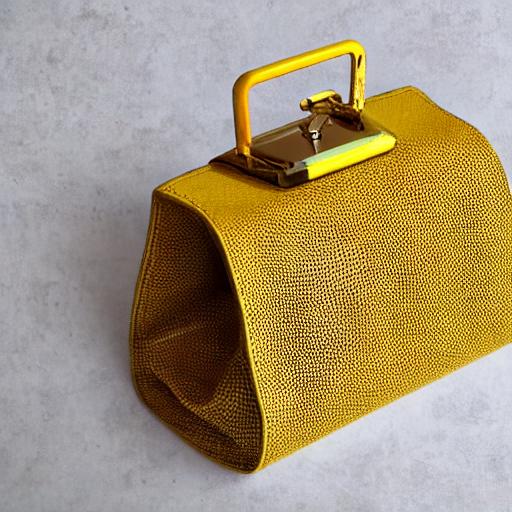} & 
			\includegraphics[width=0.12\textwidth]{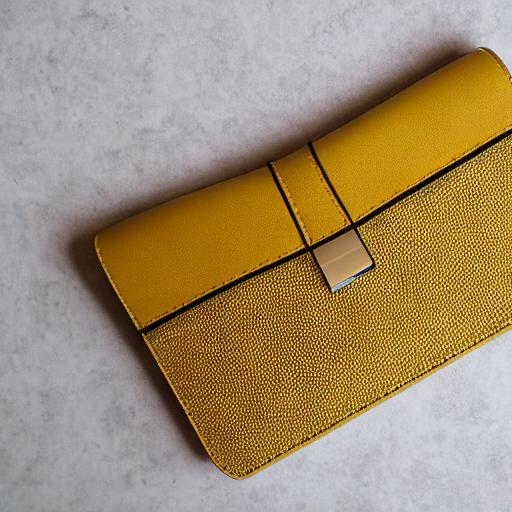} & 
			\includegraphics[width=0.12\textwidth]{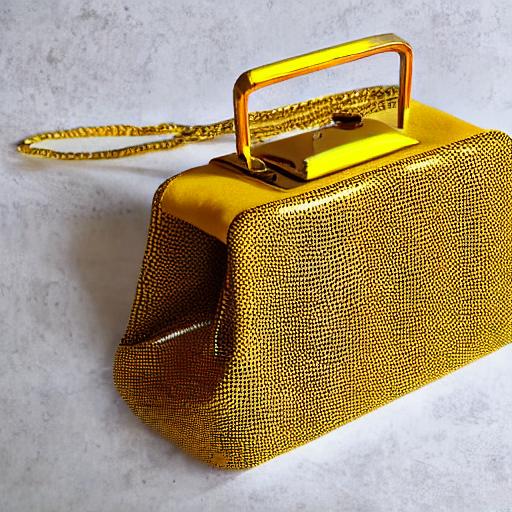} { } { }&
			\includegraphics[width=0.12\textwidth]{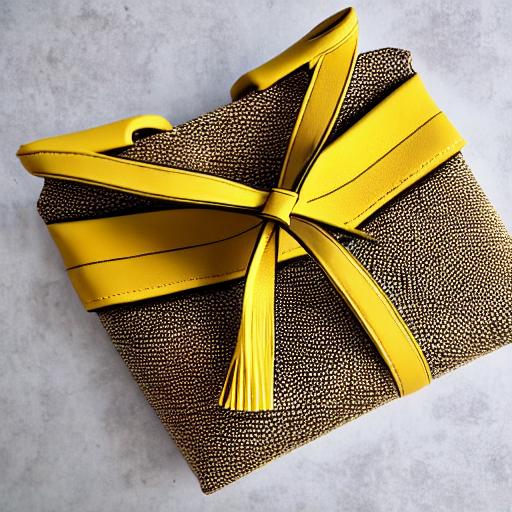} & 
			\includegraphics[width=0.12\textwidth]{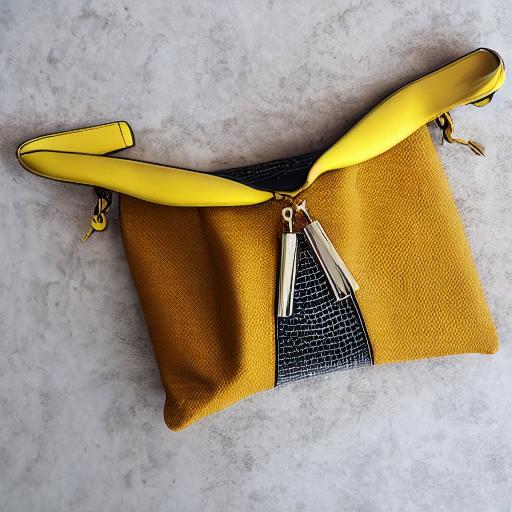} & 
			\includegraphics[width=0.12\textwidth]{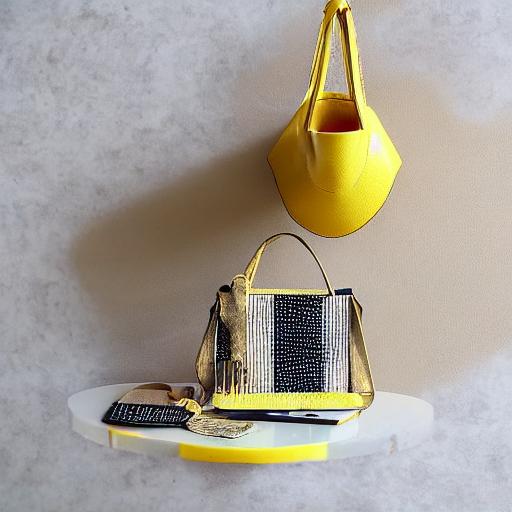} & 
			\\		
		\end{tabular}
	\vspace{5pt}
	\caption{Our method traverses the CLIP space using proxy words, enabling us to generate object-level variations. 
    Here, we generate two different outputs using the prompt ``Luxury yellow \emph{purse} on a table'' shown in the ``original'' column. We then apply Mix-and-Match using proxy words, where the left part shows auto-suggested proxy words using our ranking technique. In the right part, we show words from a large distance in CLIP's space. 
    As illustrated above, distant words tend to yield objects which can no longer be identified as the original object, while our method allows for generating convincing object-level variations.}
    \vspace{-8pt}
	\label{fig:proxy-words}
\end{figure*}

In Figure~\ref{fig:proxy-words}, we show a few proxy-words examples. For the prompt ``Luxury yellow \emph{purse} on a table'' we take three high-ranked proxy-words and three words with a large CLIP distance. As can be seen, even words that are loosely related to the word ``purse'' (\eg, ``butterfly'') sometimes allow for generating plausible shape variations. However, closer words tend to produce more successful variations.

\begin{figure}
        \setlength{\tabcolsep}{0.5pt}
        \centering
        {\scriptsize
        \begin{tabular}{c c c c}
            \multicolumn{4}{c}{``a dog with a hat in the park'' $\rightarrow$ ``a dog with a crown in the park''} \\
            \vspace{3mm}
            \raisebox{23pt}{\multirow{5}{*}{\includegraphics[width=0.33\linewidth]{images/local_ablation/155/originial.jpg}}} &&
            \includegraphics[width=0.33\linewidth]{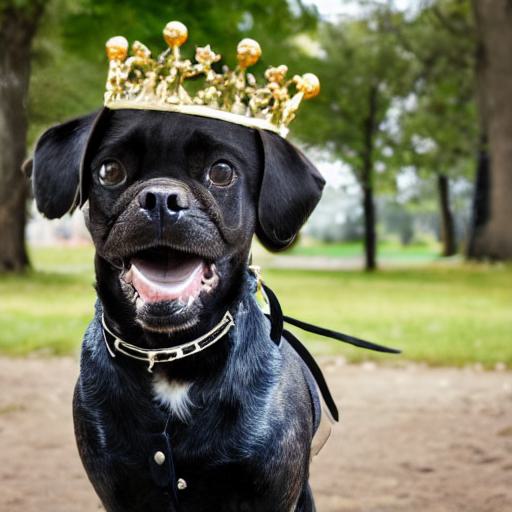} &
            \includegraphics[width=0.33\linewidth]{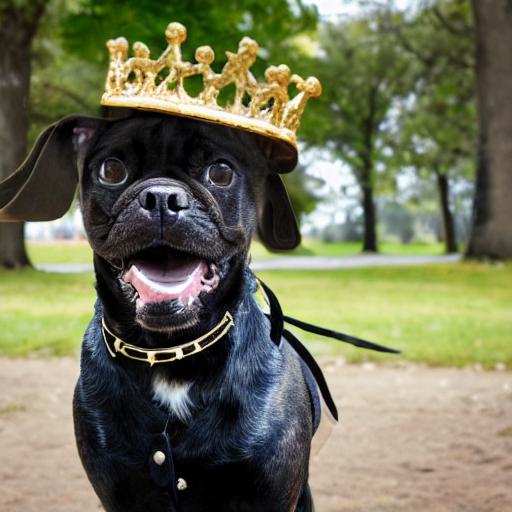} \\
            &&{ \footnotesize  P2P w/ } &
            { \footnotesize  P2P w/ } \\
            &&{ \footnotesize  our rows mask} &
            { \footnotesize our columns mask} \\
            &&\includegraphics[width=0.33\linewidth]{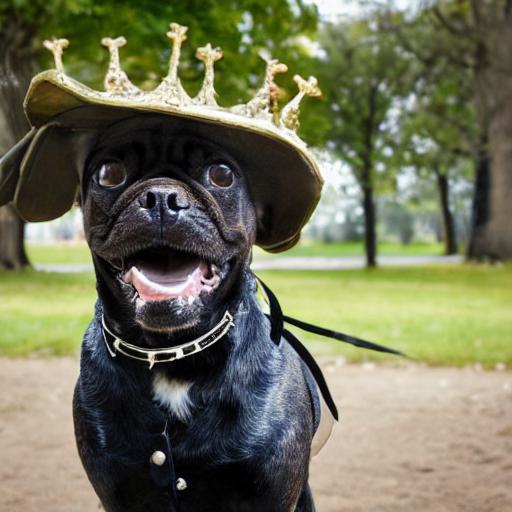} &
            \includegraphics[width=0.33\linewidth]{images/local_ablation/155/ours_02_45_both_no_obj.jpg} \\
            &&{ \footnotesize  P2P w/ } &
            { \footnotesize  P2P w/ both our masks} \\
            { \footnotesize  Original } &&
            { \footnotesize both our masks} &
            { \footnotesize without hat pixels} \\           
        \end{tabular}
        \vspace{1pt}}
    \caption{
    Comparison between different masks used for self-attention map injection, integrated with P2P.
    } 
    \label{fig:local_ablation_supp}
\end{figure}

\subsection{Attention-Based Shape Localization}

\paragraph{Self-Attention Injection Mask Controls}
As we describe in Section~\ref{sec:attn-based-loc}, 
we selectively inject the self-attention map of the original image into the generated image, by constructing a mask that contains rows and columns corresponding to pixels of the object we aim to preserve. Specifically, we define the mask as follows:
\begin{equation} \label{eq:row_cols_masl}
    M_t^{(l)}[i, j] = 
    \begin{cases}
           1 &  i\in O_t^{(l)} \enspace\text{or}\enspace j \in O_t^{(l)}\\
           0 &\text{otherwise}, \\ 
         \end{cases}
\vspace{-2pt}
\end{equation}
where $O_t^{(l)}$ is the set of pixels corresponding to the object we aim to preserve.
Injecting only the rows (\ie, setting 1 in $M_t^{(l)}$ if $i \in O_t^{(l)}$) keeps the effect of each pixel in the image on the pixels in $O_t^{(l)}$, but allows the pixels in $O_t^{(l)}$ to change their effect on other pixels in the image. When using such a ``rows'' mask, pixels in the image which weren't part of the object ($O_t^{(l)}$) in the original image, can behave as part of the object in the new image. This can be seen in Figure~\ref{fig:local_ablation_supp}, where ears were added to the dog in pixels that contained the hat in the original image. Conversely, injecting only the columns (\ie, setting 1 in $M_t^{(l)}$ if $j \in O_t^{(l)}$) allows pixels in the image to affect differently on pixels of the object to preserve ($O_t^{(l)}$) than in the original image, but keeps the effect of the pixels in $O_t^{(l)}$ on the other pixels in the image. In our experiments, we found the mask that contains both the rows of the columns (Equation~\ref{eq:row_cols_masl}) to be the most robust.

In some cases, using both ``rows'' and ``columns'' masks to preserve an object limits the required geometric changes of the object of interest (\ie, the object we aim at obtaining variations of). 
We define $O_t^{'(l)}$ as the set of pixels corresponding to the object of interest. In such cases, the relations between each pixel in $O_t^{(l)}$ to each pixel of $O_t^{'(l)}$ remain as they were in the original image. In those cases we suggest the following mask: 
\begin{equation}
    M_t^{(l)}[i, j] = 
    \begin{cases}
           1 &  i\in O_t^{(l)} \setminus O_t^{'(l)} \enspace\text{or}\enspace j \in O_t^{(l)} \setminus O_t^{'(l)} \\
           0 &\text{otherwise}. \\ 
         \end{cases}
\end{equation}

Removing the pixels of the object of interest from the mask unleashes the object of interest from its connection to the object we are preserving.
The choice of whether to remove $O_t^{'(l)}$ from the mask depends on the user's preferences.

\subsection{Self-Segmentation}
\begin{figure}
       \centering
        \setlength{\tabcolsep}{1pt}
        {\scriptsize
        \begin{tabular}{cccc}
            \includegraphics[width=0.24\linewidth]{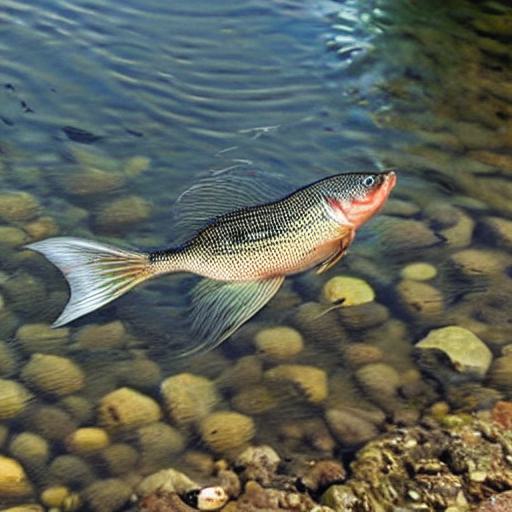} &
            \includegraphics[width=0.24\linewidth]{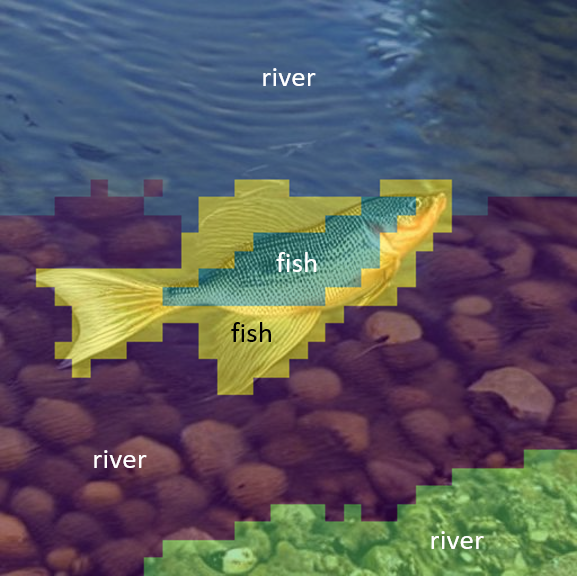} &
            \includegraphics[width=0.24\linewidth]{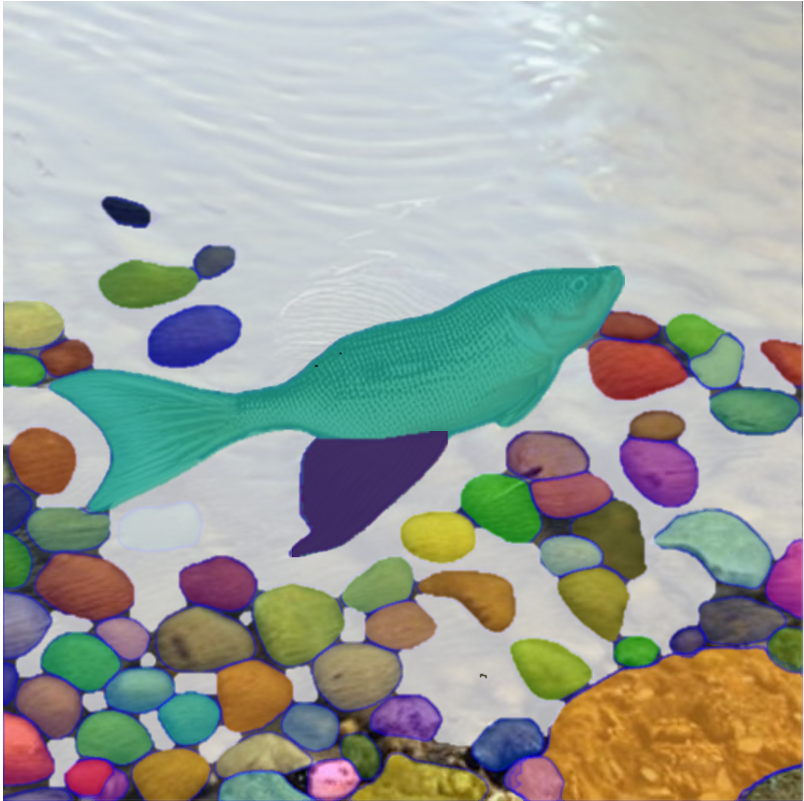} &
            \includegraphics[width=0.24\linewidth]{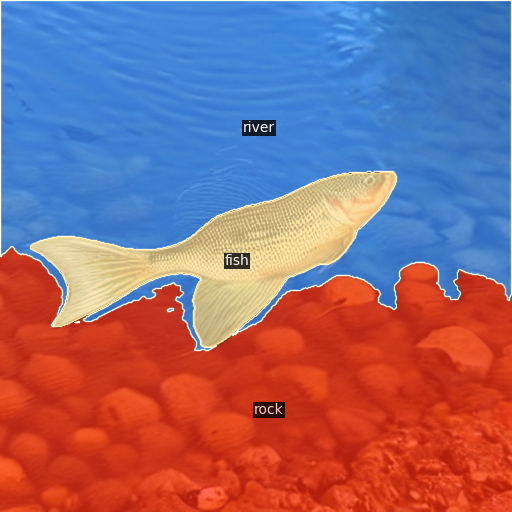} 
            \\
            \includegraphics[width=0.24\linewidth]{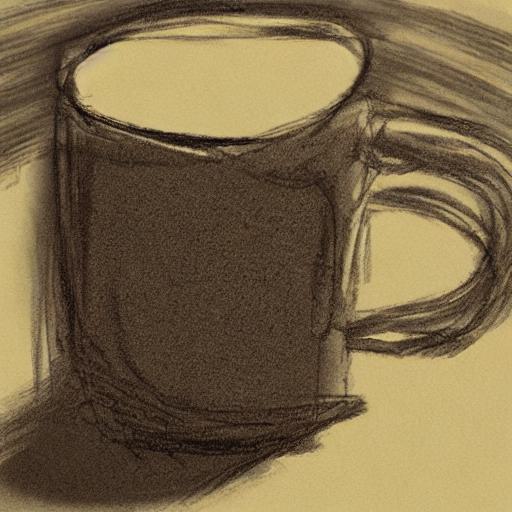} &
            \includegraphics[width=0.24\linewidth]{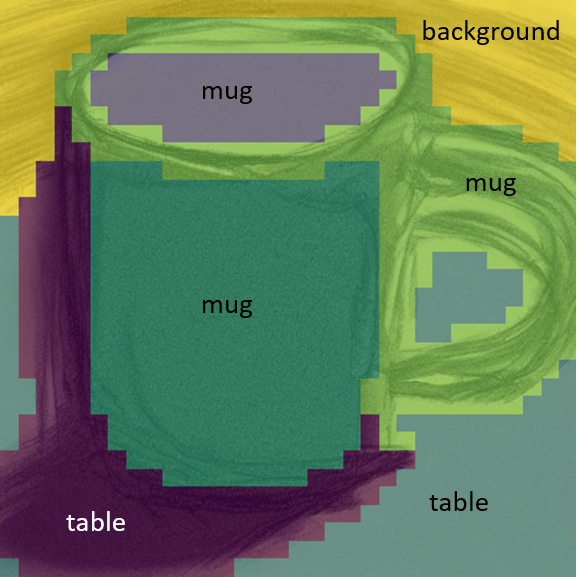} &
            \includegraphics[width=0.24\linewidth]{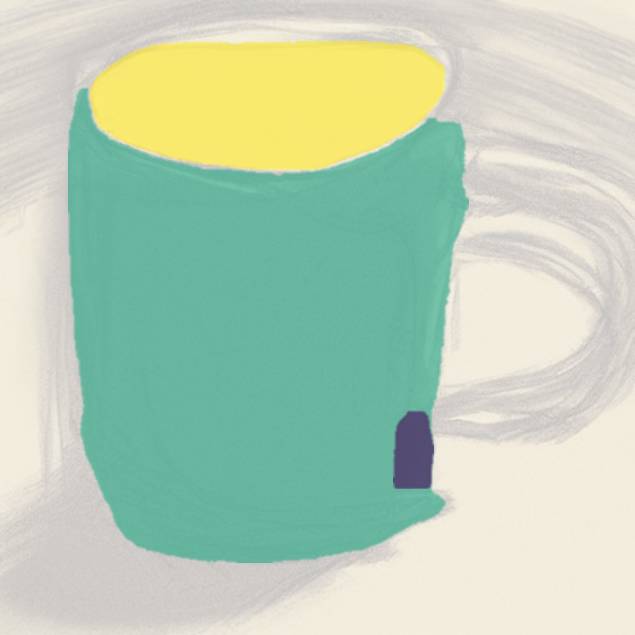} &
            \includegraphics[width=0.24\linewidth]{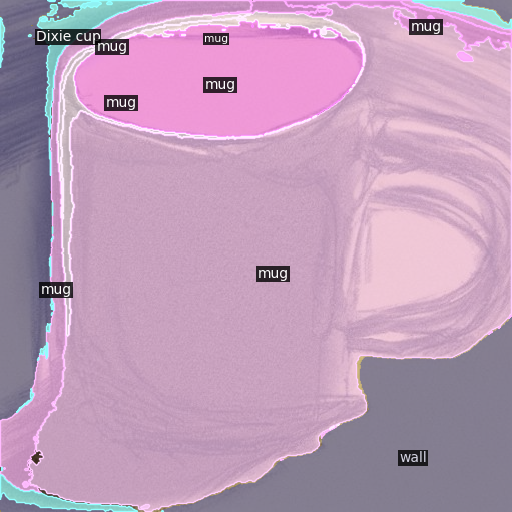} \\
            Original & Ours & SAM & ODISE
        \end{tabular}
        }
        \caption{Comparing our self-segmentation technique with state-of-the-art segmentation methods on challenging cases. Observe the transparent fin of the fish. }
        \vspace{-12pt}
        \label{fig:seg-supp}
\end{figure}

The self-segmentation technique that we introduce leverages the generative prior of large diffusion models, rather than an external segmentation model. 
While large segmentation models (e.g., SAM~\cite{kirillov2023segment}, ODISE~\cite{xu2023open}) trained with a lot of supervision are strong, they also come with heavy computational costs. Our self-segmentation utilizes the rich features that are used to synthesize the given image. This allows dealing with challenging scenarios, like the fish with a transparent fin or a non-realistic image such as the sketch of the cup shown in Figure~\ref{fig:seg-supp}. Zoom in to observe the upper fin of the fish.

        \section{Experiments Settings}

\subsection{Implementation Details}
In our experiments, we used $T = 50, T_3 = 44 \pm 1, T_2 = 34 \pm 1, T_1 = 15\pm 1$. We found that the optimal $T_3, T_2$ which indicates the prompt-mixing range may slightly change between different prompts and seeds. 

For segmenting the images using the self-attention maps, we used $5$ clusters across all our experiments. Note, that since we automatically label each cluster, it is possible to increase the number of clusters without affecting the results. We used $\sigma = 0.3$ for the clusters labeling.

Our method does not require optimization, and can therefore not consume a lot of memory. 

\vspace{-4pt}

\subsection{Evaluation Setup for Alternative Methods}
As we describe in Section~\ref{sec:exp-obj-varietions}, we compare our method to two types of methods: (i) Non-deterministic methods that provide different results for different seeds (SDEdit\cite{meng2022sdedit} and Inpainting\cite{Rombach2021HighResolutionIS}), and (ii) Text-guided image editing methods (Prompt-to-Prompt and Instruct-Pix2Pix).

The na\"{\i}ve way to attain variations with non-deterministic methods is to insert the input image into the method with a different seed each time. We also compare to these methods using our auto-generated proxy words, where we use the original prompt, with the proxy word replacing the word corresponding to the object of interest. In Figure~\ref{fig:quantitative_comparison}, both approaches are presented, the na\"{\i}ve way (indicated by the method's name, \eg, SDEdit) and each method when using proxy words (\eg, SDEdit-proxy).

As mentioned in Section~\ref{sec:exp-obj-varietions}, we used text-guided image editing methods to create variations by refining the prompt, adding adjectives to the explored object, or replacing the explored object with our proxy words.
Prompt-to-Prompt\cite{hertz2022prompt} supports both refining a prompt and replacing a word in a prompt. To apply Plug-n-Play\cite{pnpDiffusion2022}, we used prompts with proxy words.
Instruct-Pix2Pix\cite{brooks2022instructpix2pix} receives as input a prompt that defines the required edit. To refine an object, we used the prompt ``make the \{object name\} more \{adjective\} and keep the \{other object\}'', where the object name is the object of interest, and we use a different adjective for different variations. We also added the suffix ``keep the {other object}'' to preserve other objects in the image. We used to prompt ``change the \{object name\} to a \{proxy word\} and keep the \{other object\}'' to replace the object with a proxy word.
Zero-shot-Image2Image\cite{Parmar2023ZeroshotIT} supports only replacing an object, and therefore we didn't use prompt refinement with this method. In this method, 1000 sentences containing both the original and target words are required to replace an object. We used ChatGPT to create 1000 sentences for each proxy word.

\vspace{-4pt}

\subsection{Quantitative Comparison Dataset}
We first created three templates of prompts: ``A \{mug\} with {$w$}'', ``A \{sofa\} with a {$w$} on it'' and ``A \{basket\} with {$w$}''. For each template, we created five prompts by replacing $w$ with 5 different words (\eg, ``A \{mug\} with tea'' and ``A \{mug\} with hot chocolate'').
For each prompt we generated 10 initial images, using different seeds, and got a dataset that contains 150 images. With each method, we created 20 variations of the object of interest for each initial image in the dataset.

        \section{Additional Results}

\subsection{Comparison to Other Methods}

\begin{figure*}
       \centering
        \setlength{\tabcolsep}{0.0pt}
        \begin{tabular}{c c c c c c c c c}
            &&&& \multicolumn{5}{c}{$\longleftarrow$ Object level variations $\longrightarrow$}
             \\
            & {} &&
            { Original } &
            \multicolumn{5}{c}{``A hamster in a \emph{tent} at the mountains. 8k''}
            \\
            \raisebox{24pt}{\rotatebox{90}{ Ours }} & {} &&
            \includegraphics[width=0.15\linewidth]{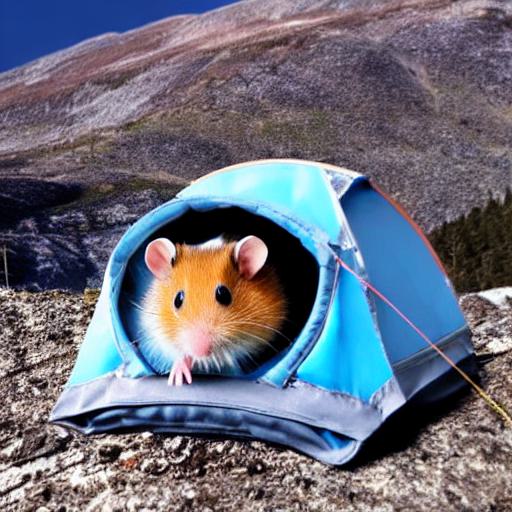} &
            {  } &
            \includegraphics[width=0.15\linewidth]{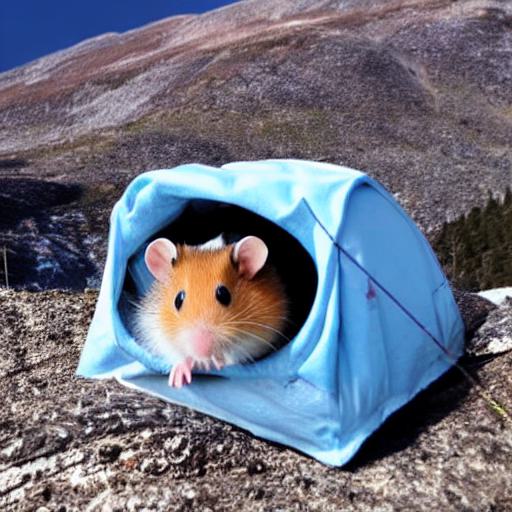} &
            \includegraphics[width=0.15\linewidth]{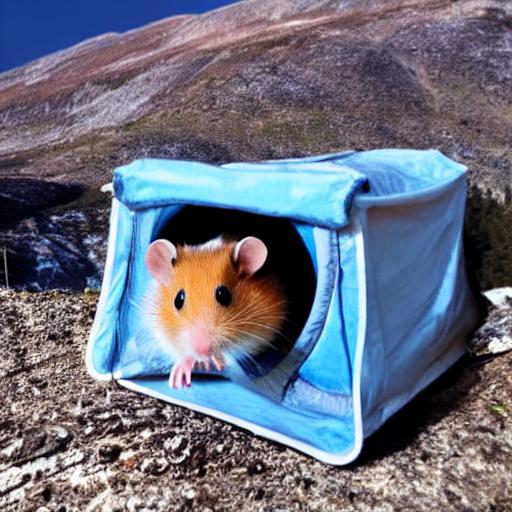} &
            \includegraphics[width=0.15\linewidth]{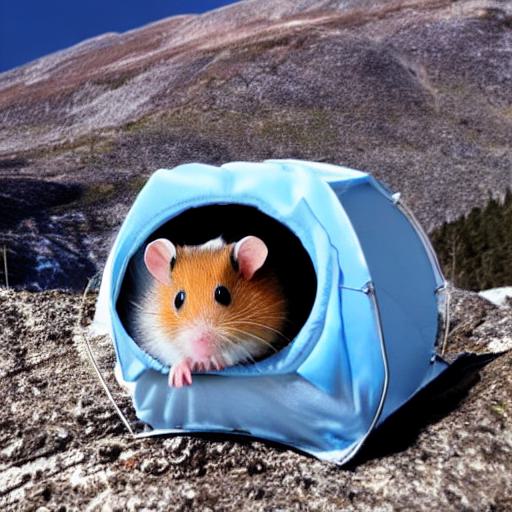} &
            \includegraphics[width=0.15\linewidth]{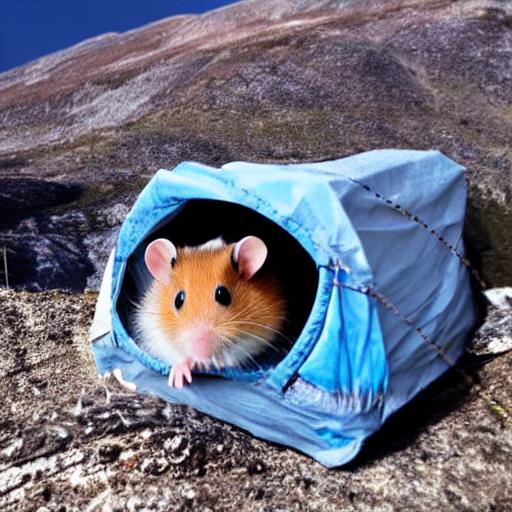}
             \\
            \raisebox{14pt}{\rotatebox{90}{ P2P (40\%)  }} & {} &&
            \includegraphics[width=0.15\linewidth]{images/comparison_tent_mountains/tent.jpg} &
            {  } &
            \includegraphics[width=0.15\linewidth]{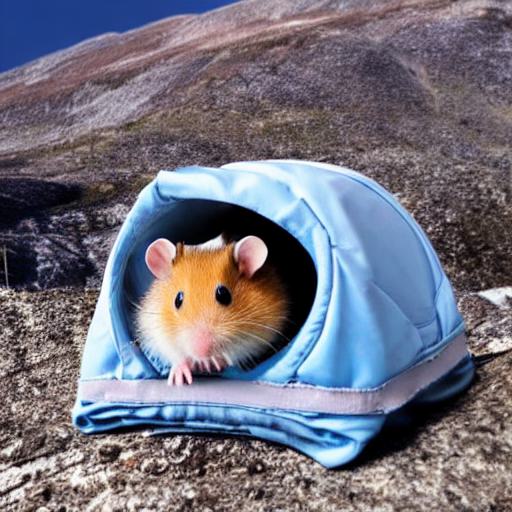} &
            \includegraphics[width=0.15\linewidth]{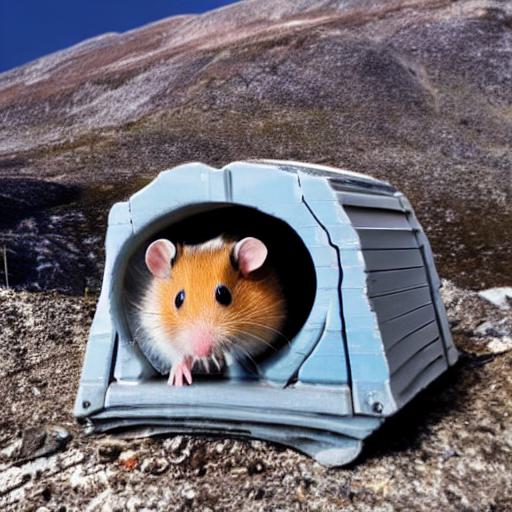} &
            \includegraphics[width=0.15\linewidth]{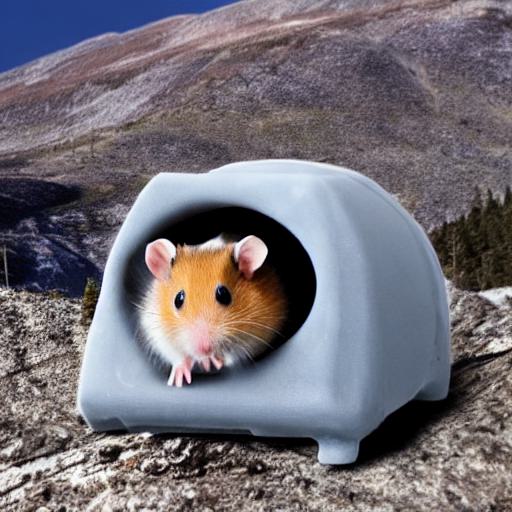} &
            \includegraphics[width=0.15\linewidth]{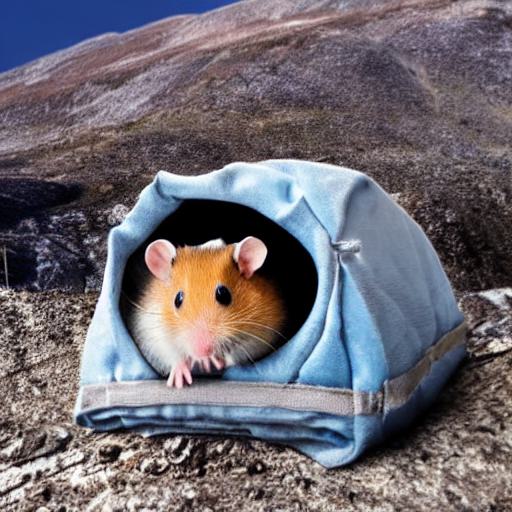}
             \\
            \raisebox{14pt}{\rotatebox{90}{ P2P (10\%) }} & {} &&
            \includegraphics[width=0.15\linewidth]{images/comparison_tent_mountains/tent.jpg} &
            {  } &
            \includegraphics[width=0.15\linewidth]{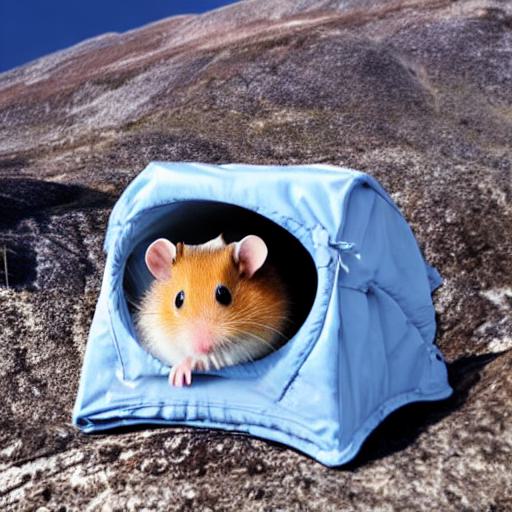} &
            \includegraphics[width=0.15\linewidth]{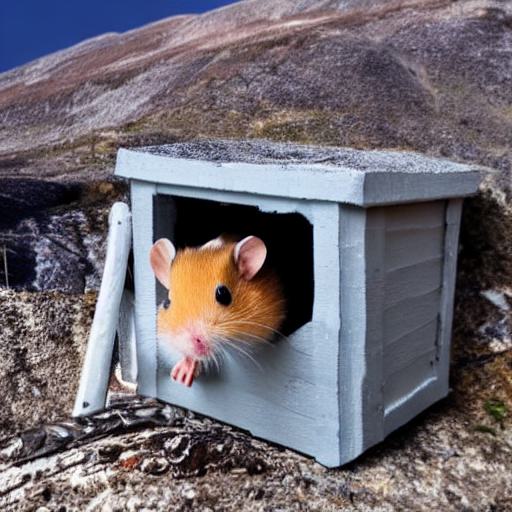} &
            \includegraphics[width=0.15\linewidth]{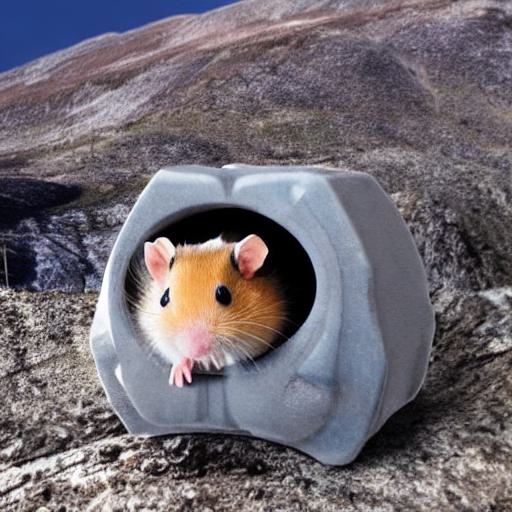} &
            \includegraphics[width=0.15\linewidth]{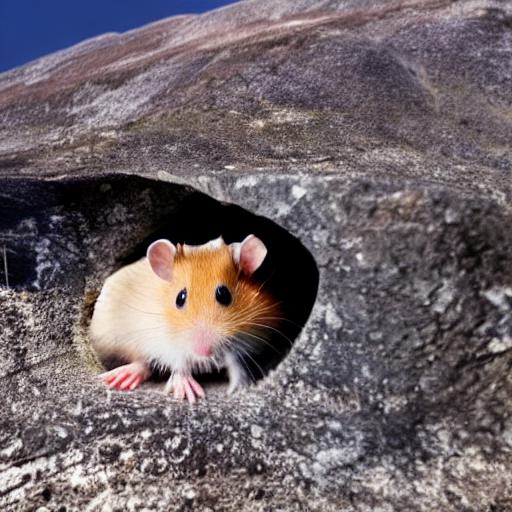} \\
            \raisebox{16pt}{\rotatebox{90}{ I-Pix2Pix }} & {} &&
            \includegraphics[width=0.15\linewidth]{images/comparison_tent_mountains/tent.jpg} &
            {  } &
            \includegraphics[width=0.15\linewidth]{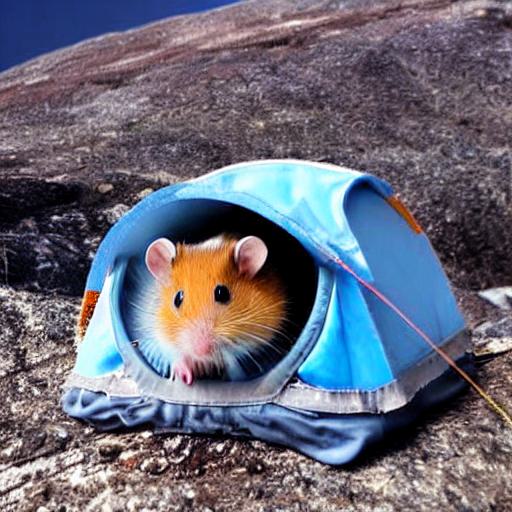} &
            \includegraphics[width=0.15\linewidth]{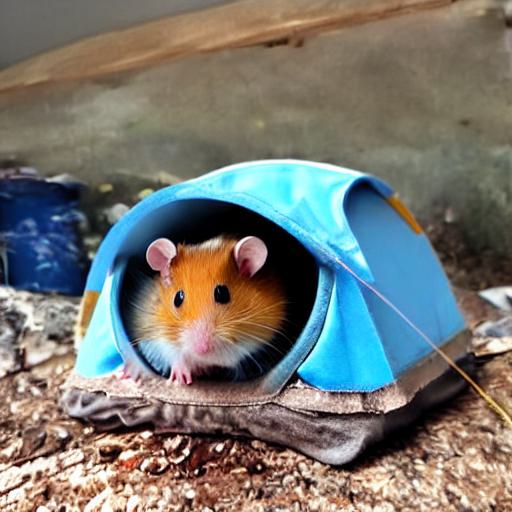} &
            \includegraphics[width=0.15\linewidth]{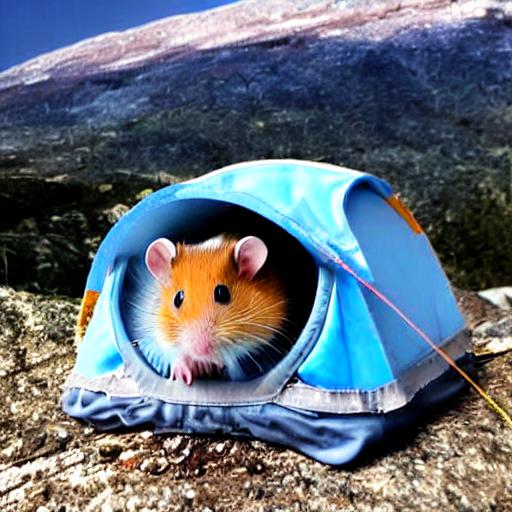} &
            \includegraphics[width=0.15\linewidth]{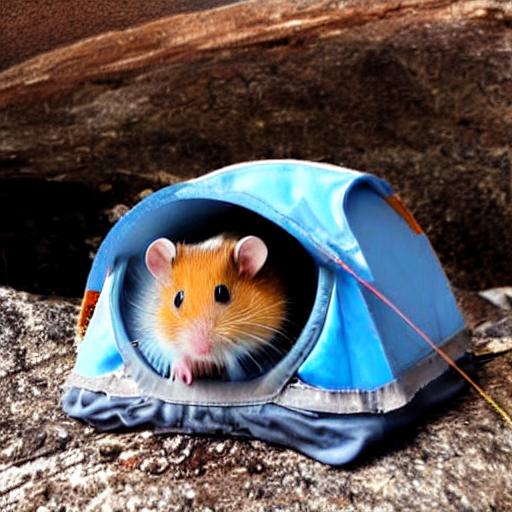}  \\
            \raisebox{26pt}{\rotatebox{90}{ PnP }} & {} &&
            \includegraphics[width=0.15\linewidth]{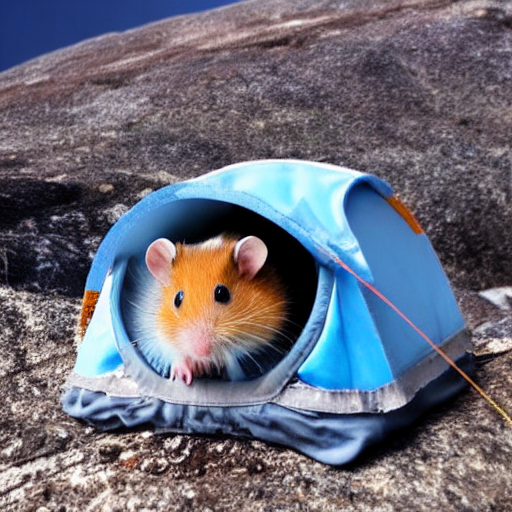} &
            {  } &
            \includegraphics[width=0.15\linewidth]{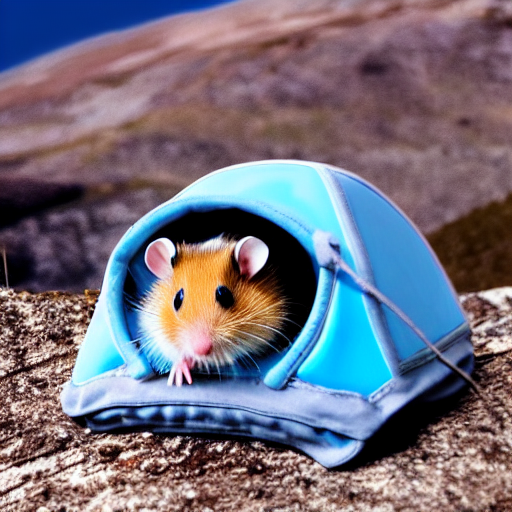} &
            \includegraphics[width=0.15\linewidth]{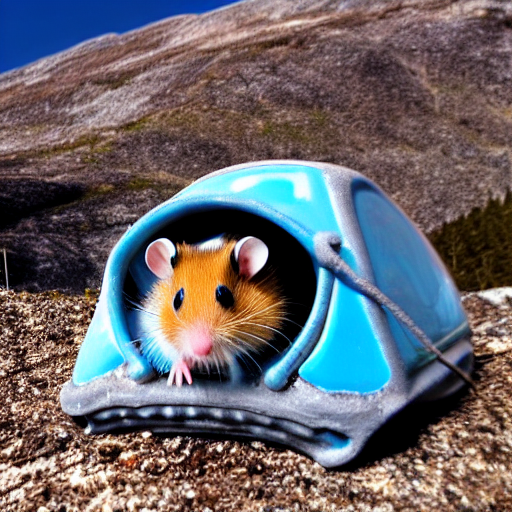} &
            \includegraphics[width=0.15\linewidth]{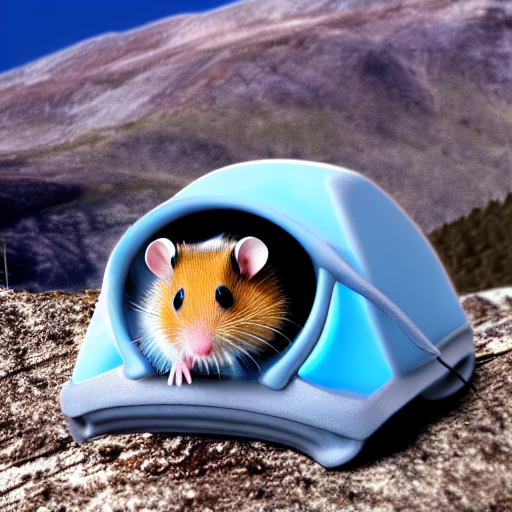} &
            \includegraphics[width=0.15\linewidth]{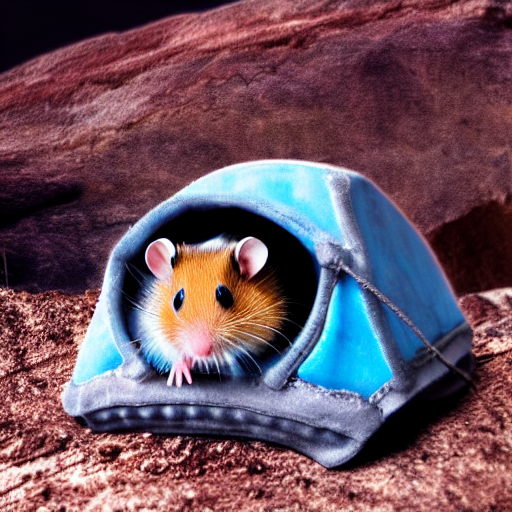} \\
            \raisebox{8pt}{\rotatebox{90}{ ZeroShotI2I }} & {} &&
            \includegraphics[width=0.15\linewidth]{images/comparison_tent_mountains/tent.jpg} &
            {  } &
            \includegraphics[width=0.15\linewidth]{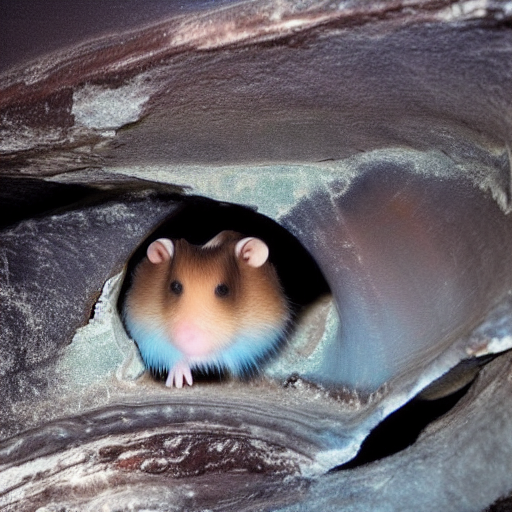} &
            \includegraphics[width=0.15\linewidth]{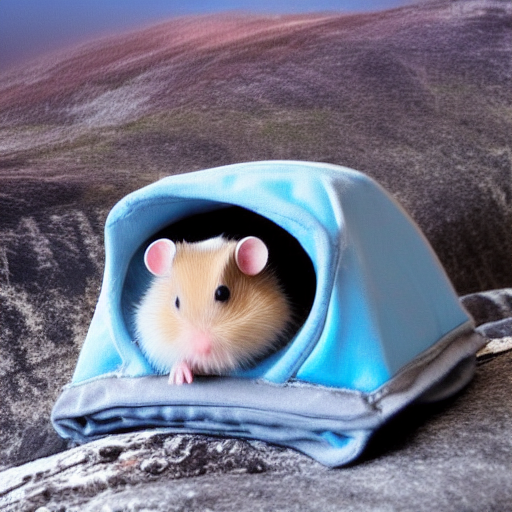} &
            \includegraphics[width=0.15\linewidth]{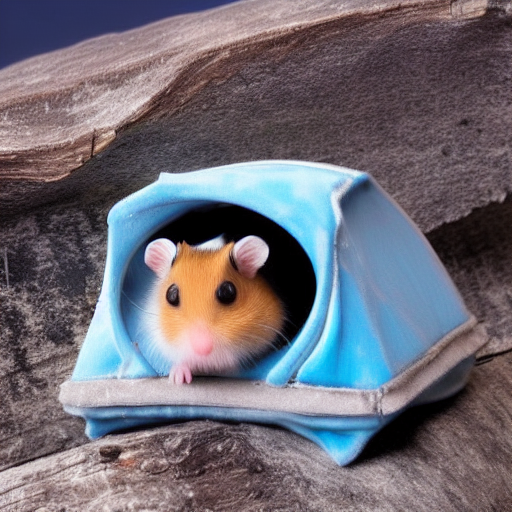} &
            \includegraphics[width=0.15\linewidth]{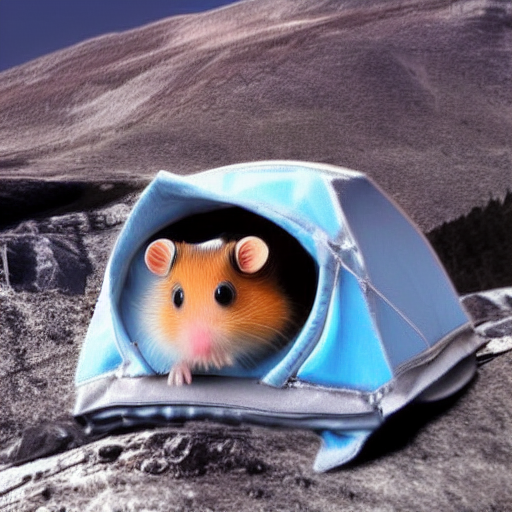} \\
            \raisebox{22pt}{\rotatebox{90}{ Imagic} } & {} &&
            \includegraphics[width=0.15\linewidth]{images/comparison_tent_mountains/tent.jpg} &
            {  } &
            \includegraphics[width=0.15\linewidth]{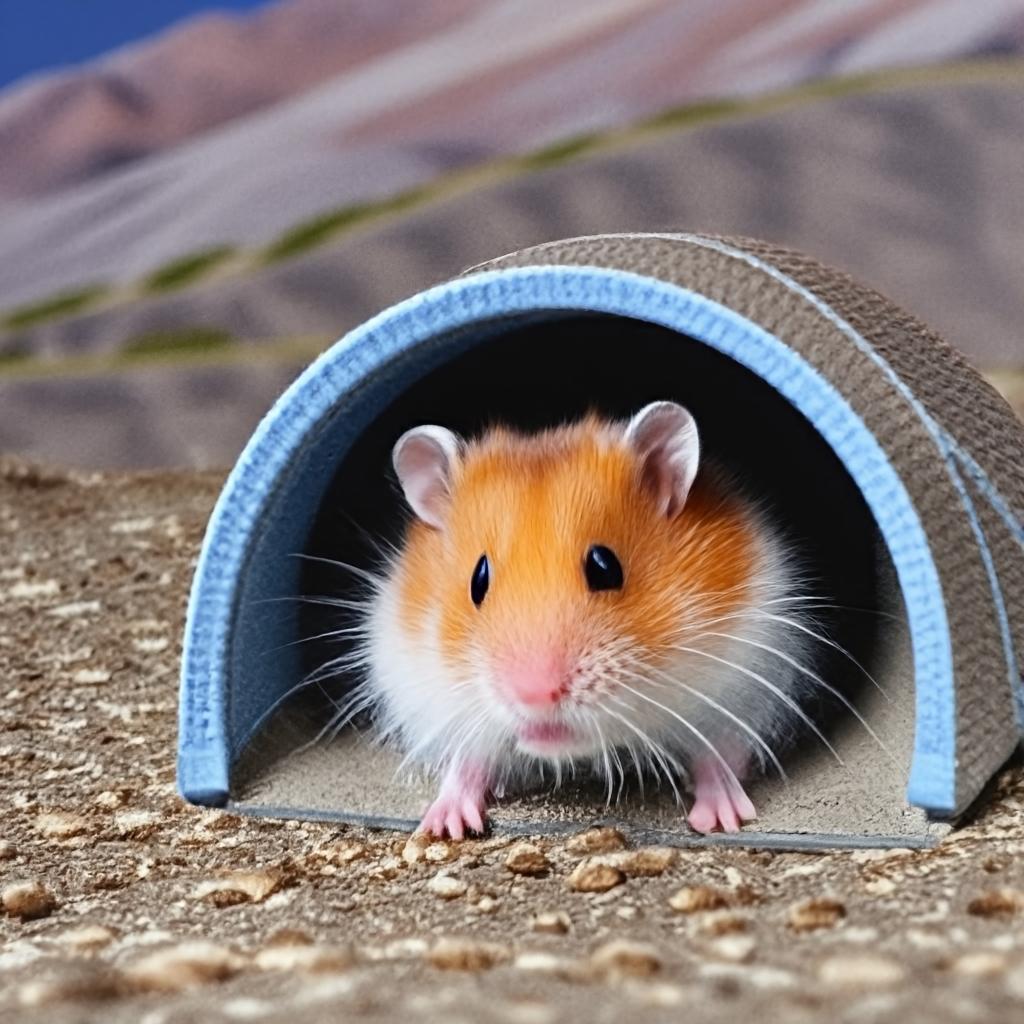} &
            \includegraphics[width=0.15\linewidth]{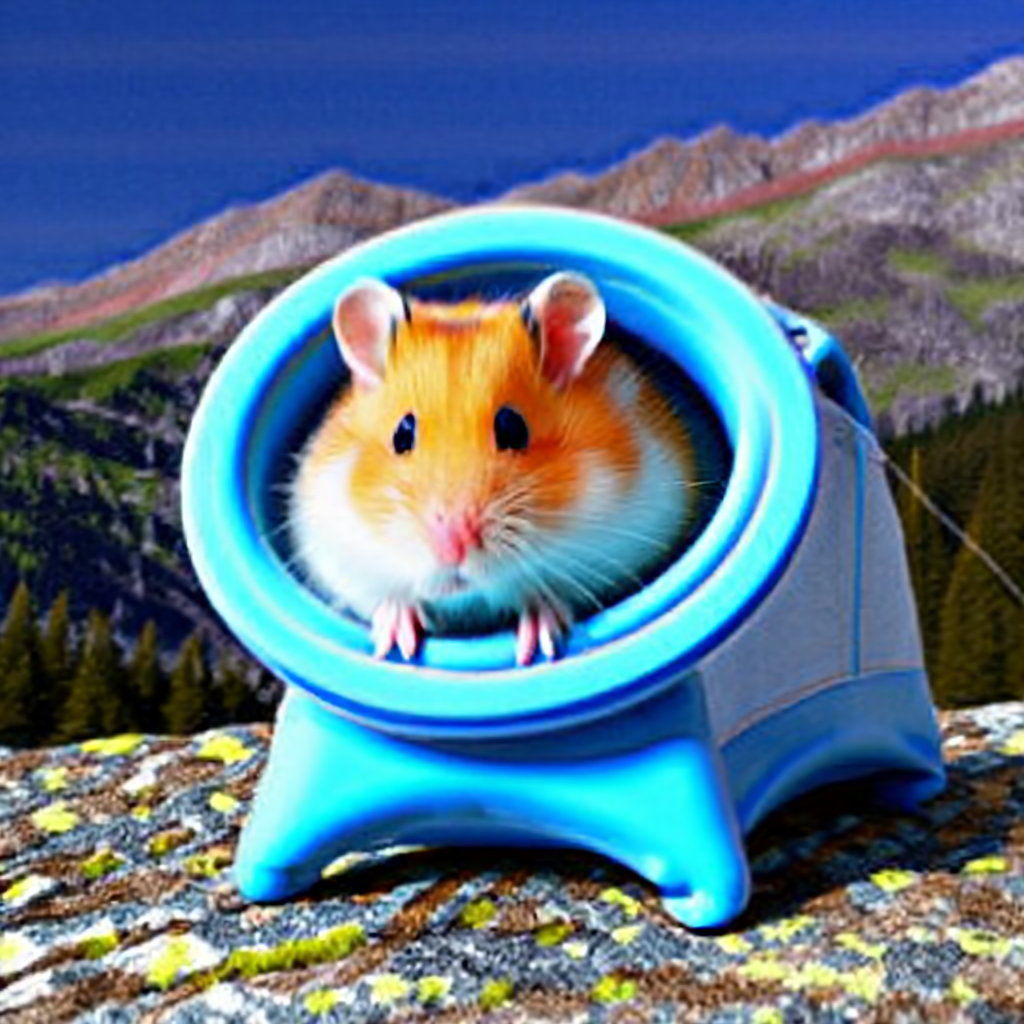} &
            \includegraphics[width=0.15\linewidth]{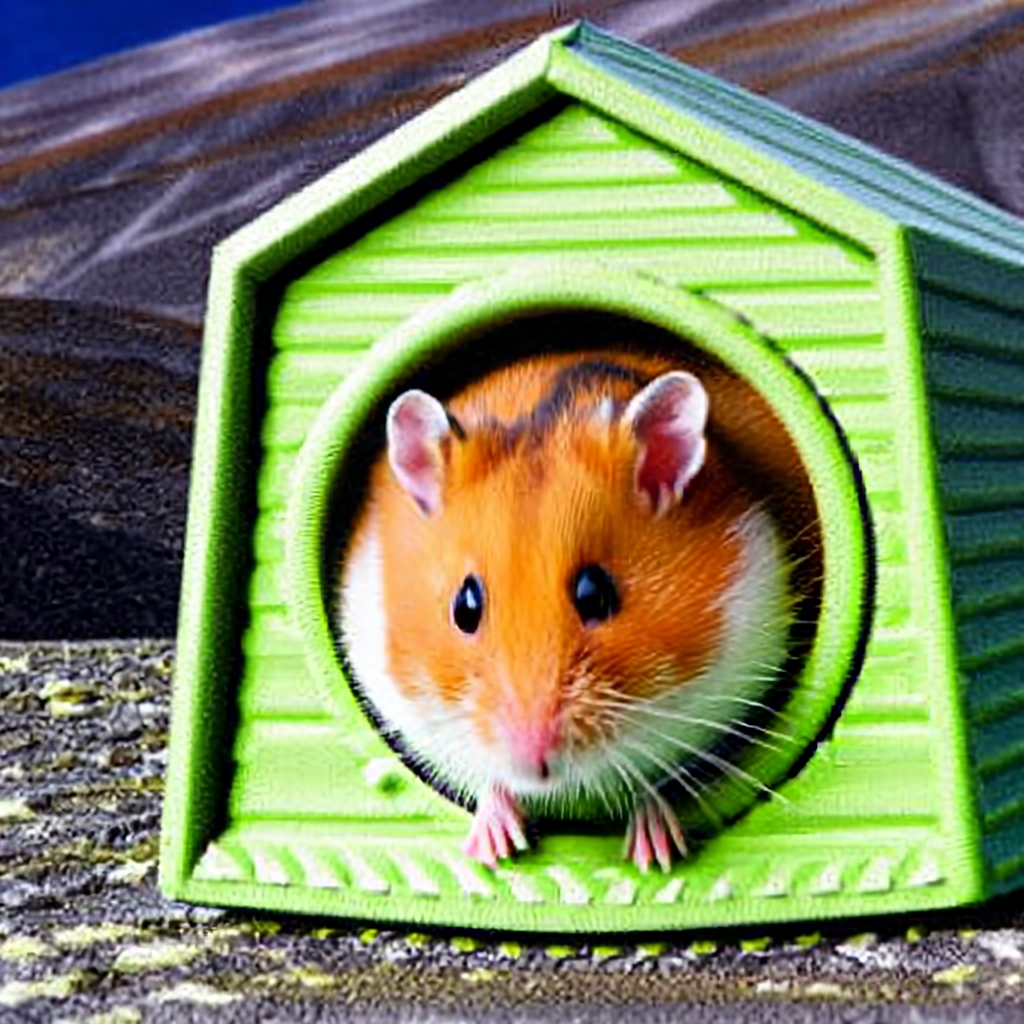} &
            \includegraphics[width=0.15\linewidth]{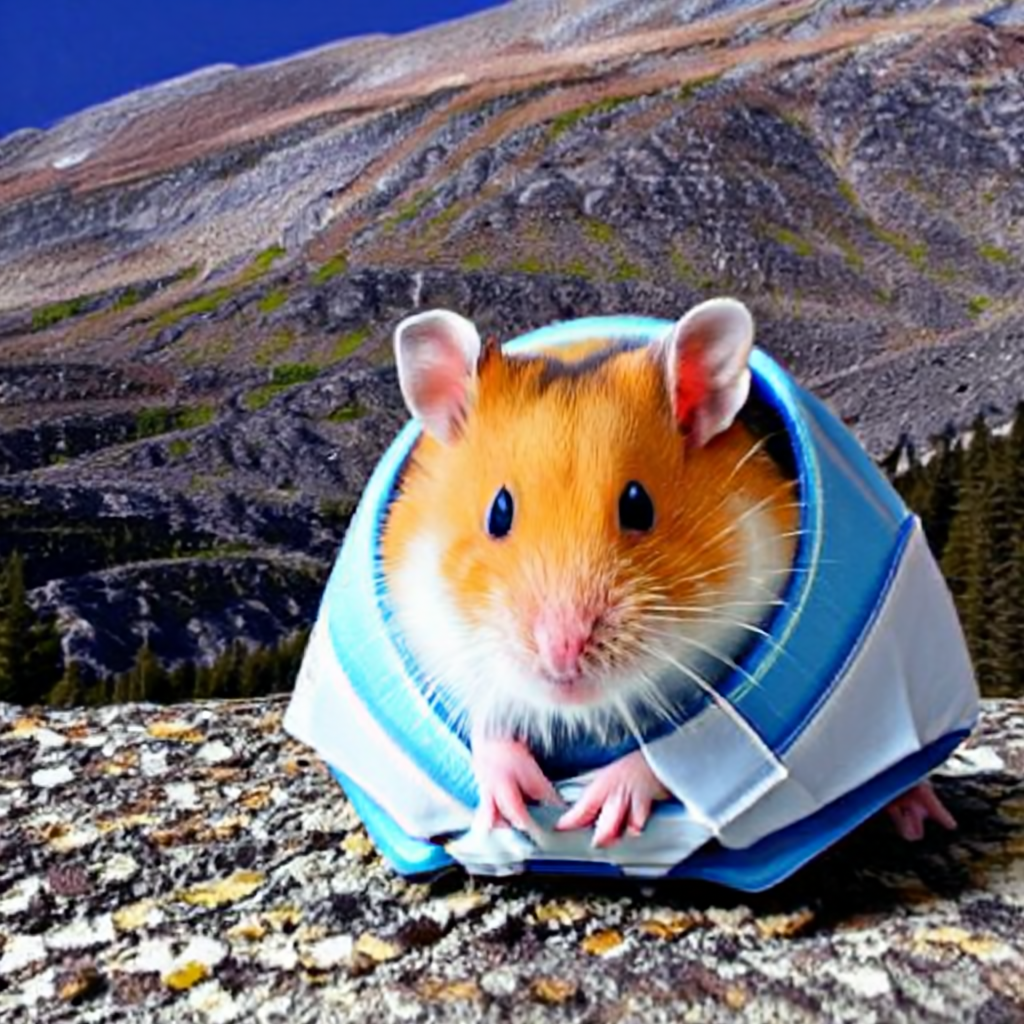} \\
            
            & {} &&
            { ``Tent'' } &
            {  } &
            { ``Cot'' } &
            { ``Shed'' } &
            { ``Something'' } &
            { ``Cave'' } 
            \\

        \end{tabular}
    \vspace{1mm}
    \captionof{figure}{
    Comparisons to text-guided editing methods. In each column, we show the results of a different word that replaces the original word ``tent'' in the prompt.}
    \label{fig:comparision_sup2}
\end{figure*}

In Figure~\ref{fig:comparision_sup2}, we extend Figure~\ref{fig:our-results} by showing another example. Our method produces diverse results and remains faithful to the class of tents while preserving the rest of the image. Plug-n-Play\cite{pnpDiffusion2022} and Instruct-Pix2Pix\cite{brooks2022instructpix2pix} make only minor texture changes to the tent and fail to preserve the background of the image. P2P\cite{hertz2022prompt} makes some noticeable texture changes when injecting self-attention during 40\% for the steps, or geometric shape changes when injecting self-attention during 10\% of the steps, but drifting away from the tents domain. 
Zero-shot Image2Image Translation struggles at changing the shape of the tent. Imagic struggles at preserving the hamster and the background and drifts away from the tents domain.

In Figure~\ref{fig:our-results-supp}, we show our result for the example in Figure~\ref{fig:inpainting}. As can be seen, our method produces significant shape changes while preserving the bananas and the background.

\subsection{Our Method Results}

\begin{figure}[h]
    \centering
    {\scriptsize
    \setlength{\tabcolsep}{0pt}
    \begin{tabular}{cc c c c }
        Original & { } &  
        \multicolumn{3}{c}{$\longleftarrow$ Object level variations $\longrightarrow$} 
 \\
        
        \raisebox{18pt}{\multirow{2}{*}{\includegraphics[width=0.24\linewidth]{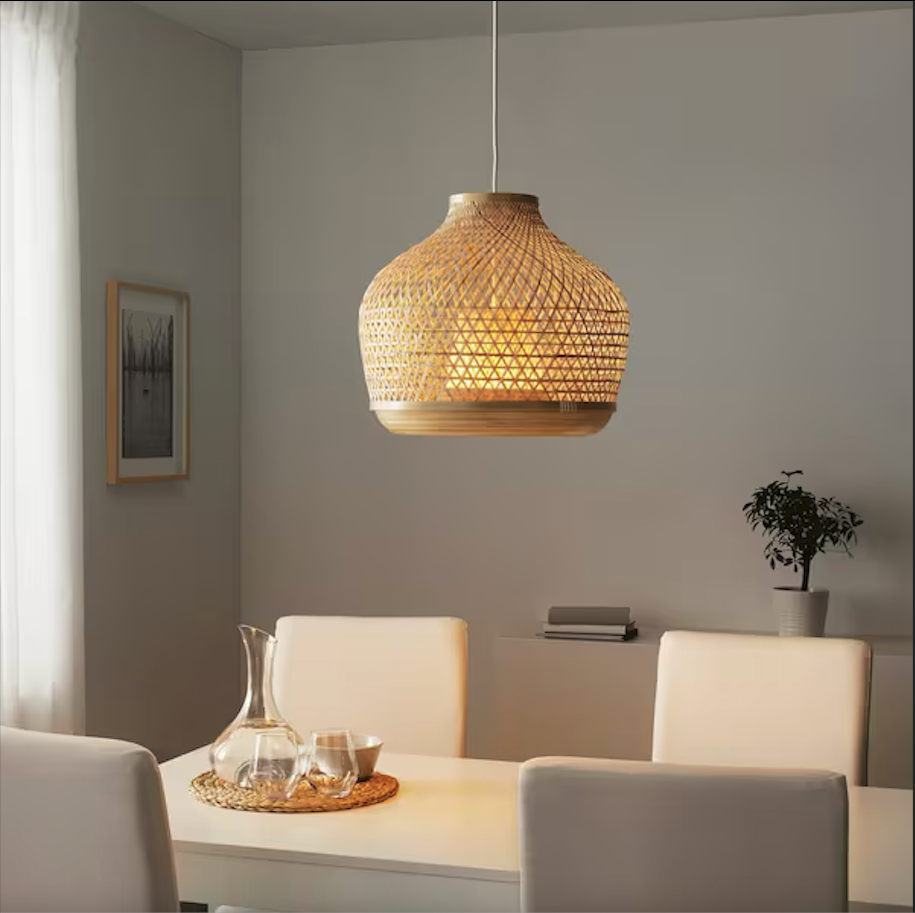}}} &
        { } &
        \includegraphics[width=0.24\linewidth]{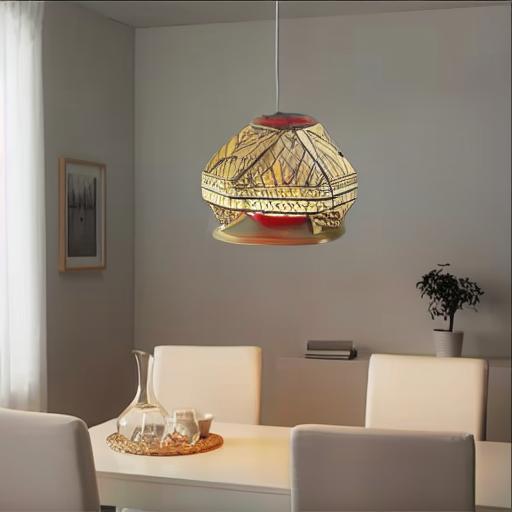} & 
        \includegraphics[width=0.24\linewidth]{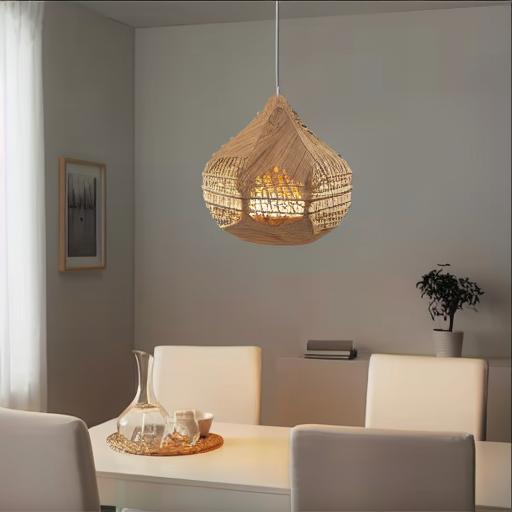} & 
        \includegraphics[width=0.24\linewidth]{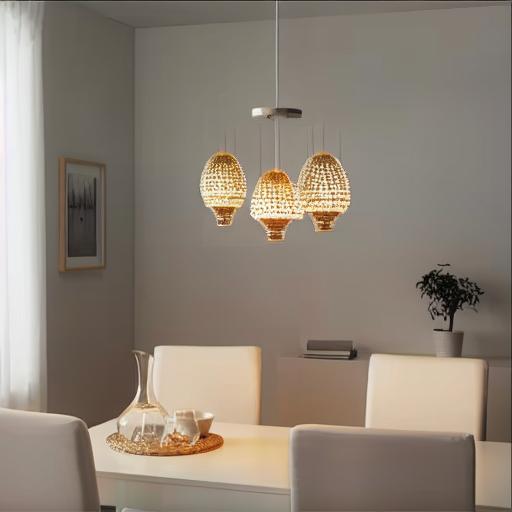} \\
        & &
        \includegraphics[width=0.24\linewidth]{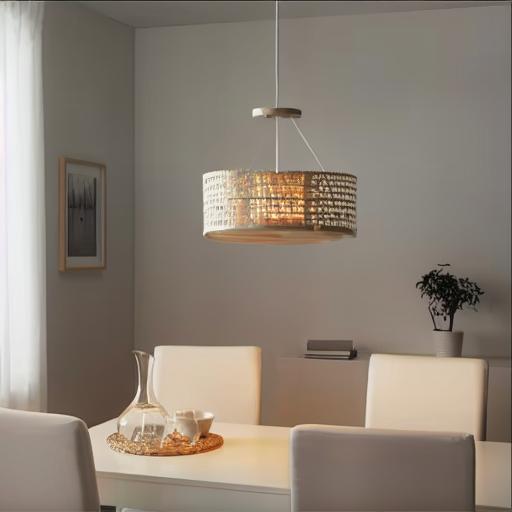} &
        \includegraphics[width=0.24\linewidth]{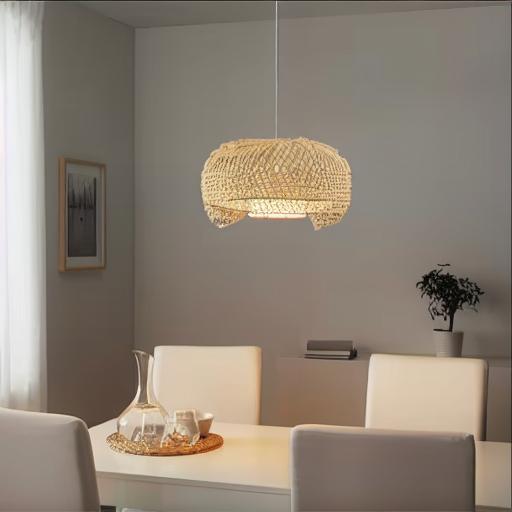} & 
        \includegraphics[width=0.24\linewidth]{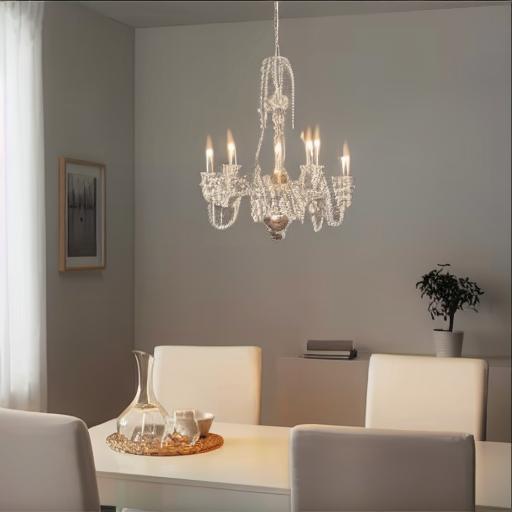} \\

        && \multicolumn{3}{c}{``A table below a \emph{lamp}''} \\
        \raisebox{18pt}{\multirow{2}{*}{\includegraphics[width=0.24\linewidth]{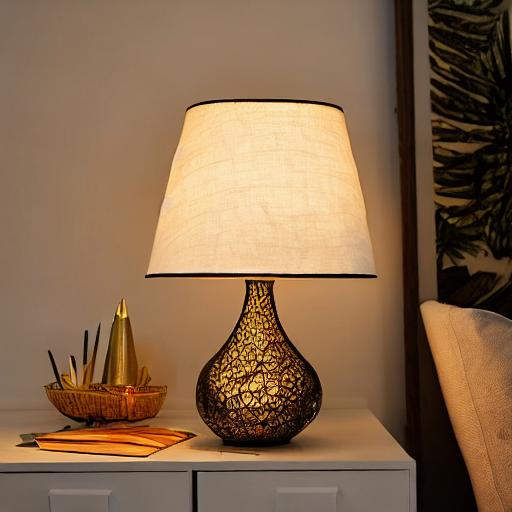}}} &
         { } & 
        \includegraphics[width=0.24\linewidth]{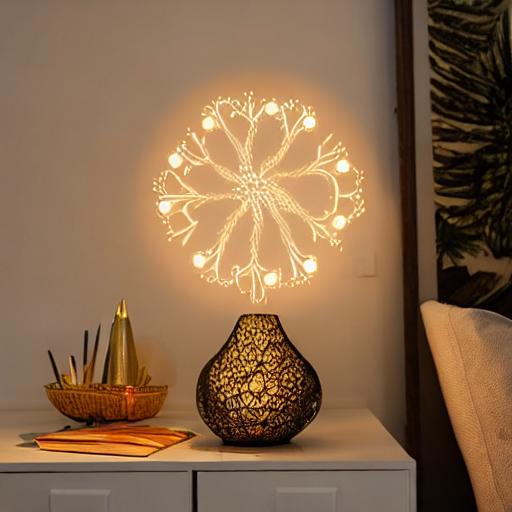} & 
        \includegraphics[width=0.24\linewidth]{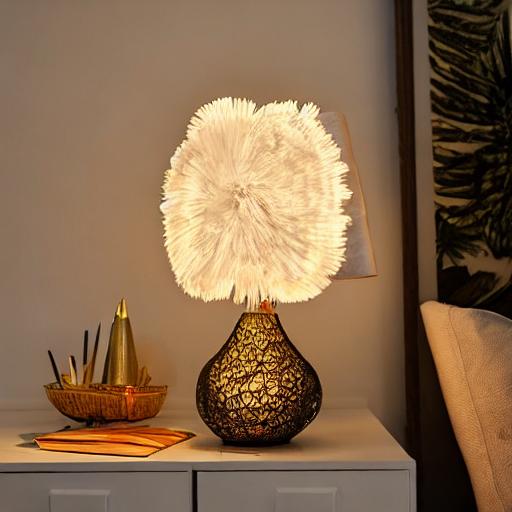} & 
        \includegraphics[width=0.24\linewidth]{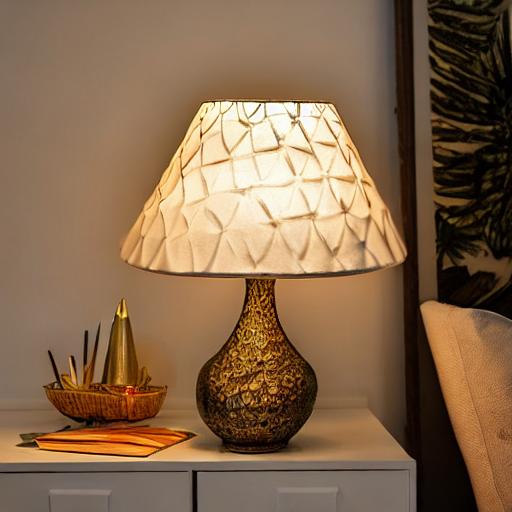} \\
        & &
        \includegraphics[width=0.24\linewidth]{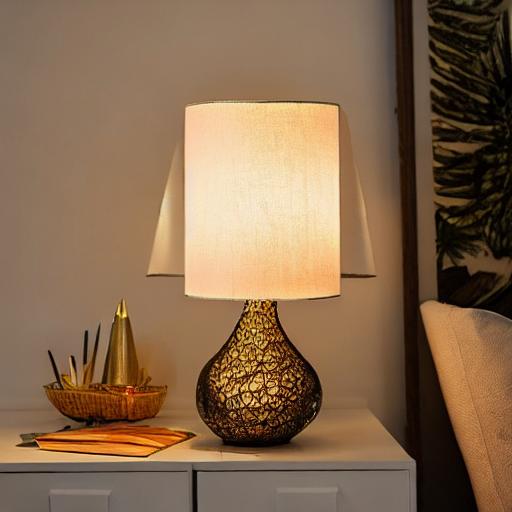} &
         
        \includegraphics[width=0.24\linewidth]{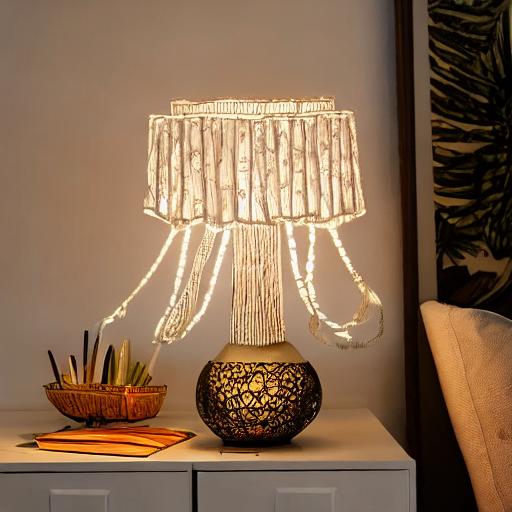} & 
        \includegraphics[width=0.24\linewidth]{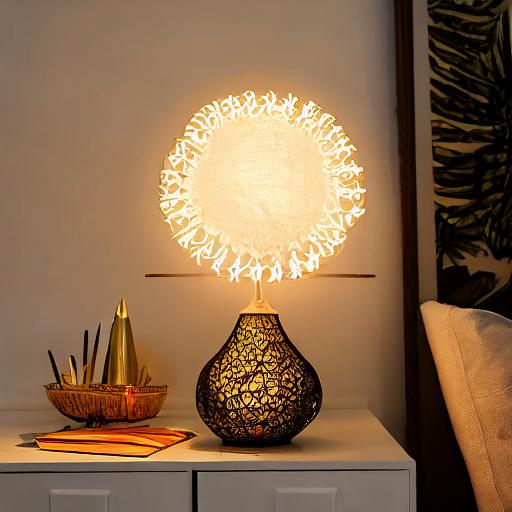}  \\
        && \multicolumn{3}{c}{``A decorated \emph{lamp} in the living room''}
    \end{tabular}
    \vspace{1mm}}
    \caption{
    Object-level variations: Real image vs Synthetic. On the top, we show object variations for a real image, where the edited object is a lamp. On the bottom, we edit the lamp but in a different synthetic scene. 
    Our method successfully generates different shape variations for both real and synthetic images.
    }
    
    \label{fig:our-extended-results-supp}
\end{figure}

\begin{figure*}
    \centering
    \setlength{\tabcolsep}{0pt}
    \begin{tabular}{cc ccc cc cc ccc}
        Original &&
        \multicolumn{3}{c}{$\longleftarrow$ Object level variations $\longrightarrow$} &&
        Original &&
        \multicolumn{3}{c}{$\longleftarrow$ Object level variations $\longrightarrow$} \\
        \includegraphics[width=0.12\textwidth]{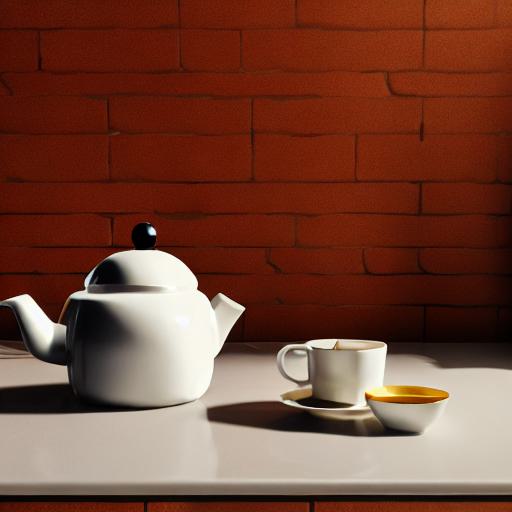} &
        { } &
        \includegraphics[width=0.12\textwidth]{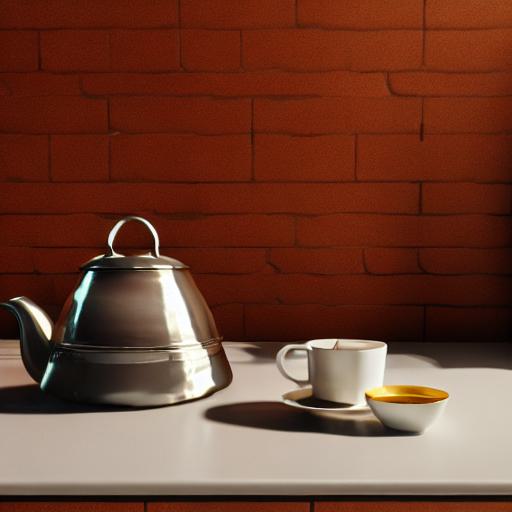} & 
        \includegraphics[width=0.12\textwidth]{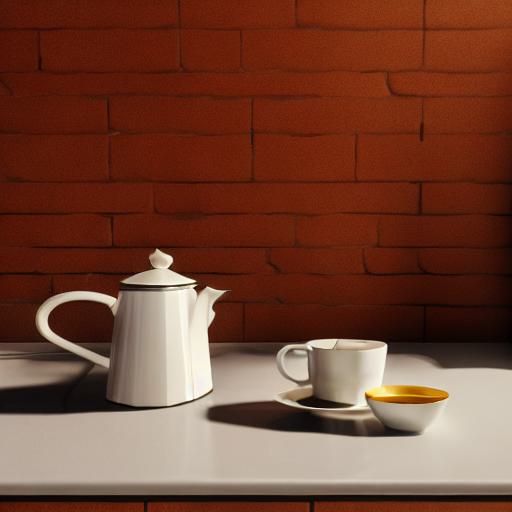} & 
        \includegraphics[width=0.12\textwidth]{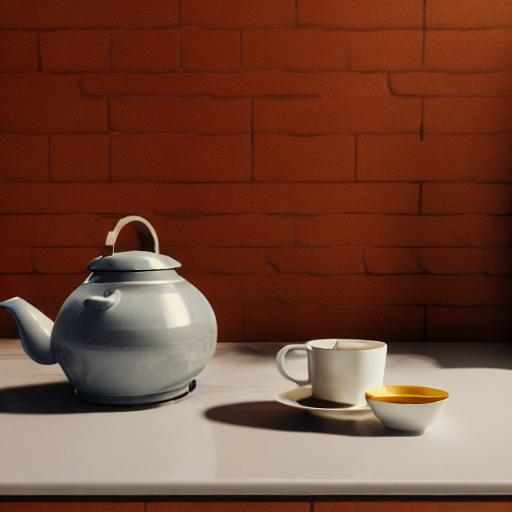} &
        {  } &
        \includegraphics[width=0.12\textwidth]{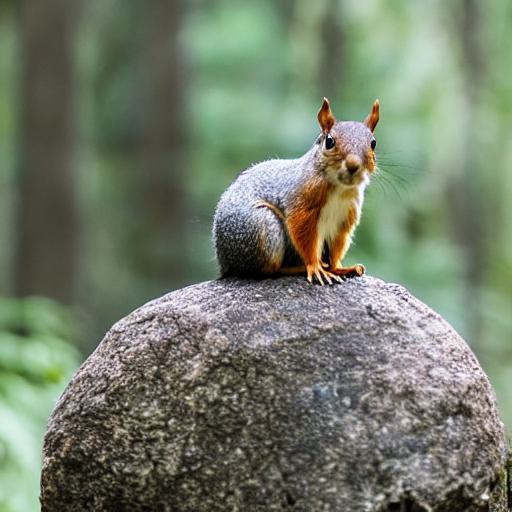} &
        { } &
        \includegraphics[width=0.12\textwidth]{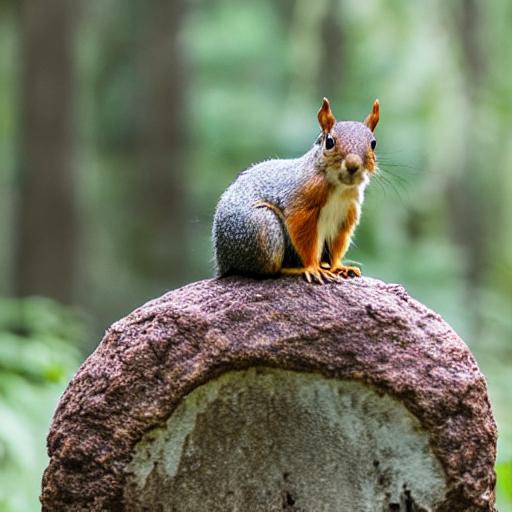} & 
        \includegraphics[width=0.12\textwidth]{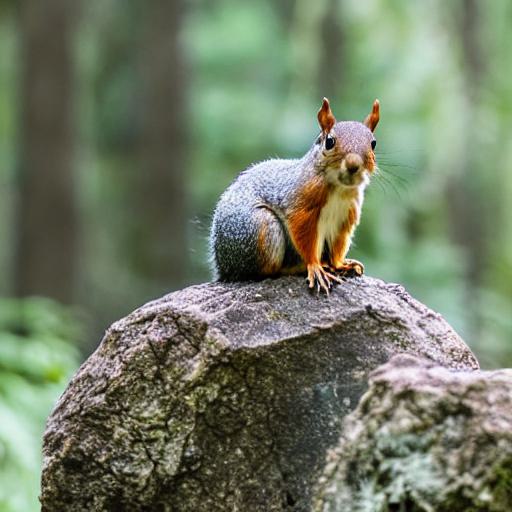} & 
        \includegraphics[width=0.12\textwidth]{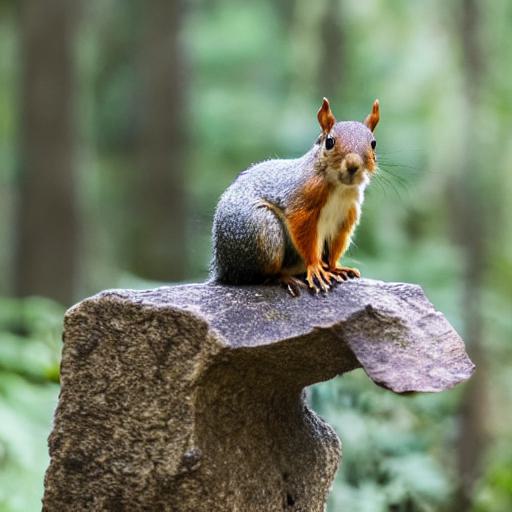} \\
        && \multicolumn{3}{c}{``a \emph{teapot} on the kitchen counter''} & &
        && \multicolumn{3}{c}{``A squirrel on a \emph{rock} in the forest''} \\
        \includegraphics[width=0.12\textwidth]{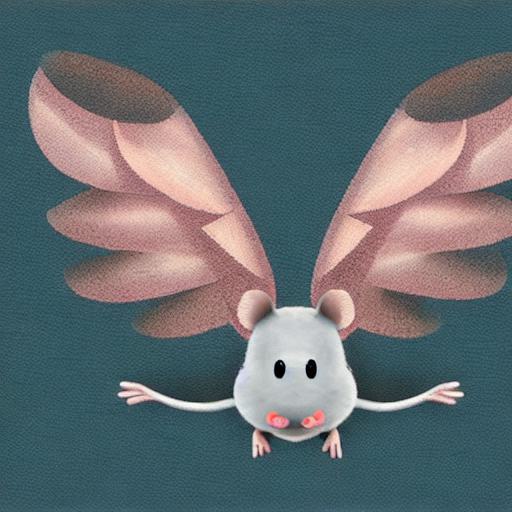} &
        { } &
        \includegraphics[width=0.12\textwidth]{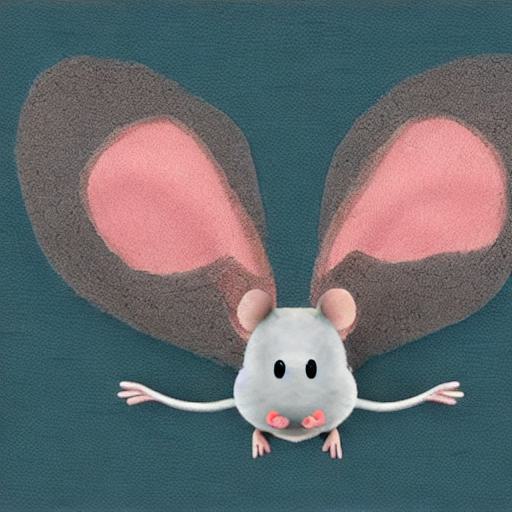} & 
        \includegraphics[width=0.12\textwidth]{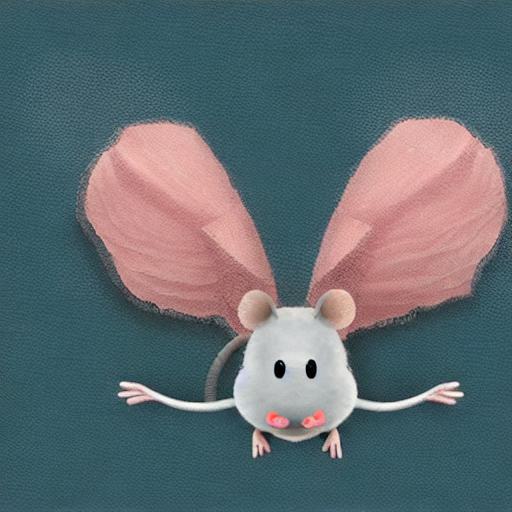} & 
        \includegraphics[width=0.12\textwidth]{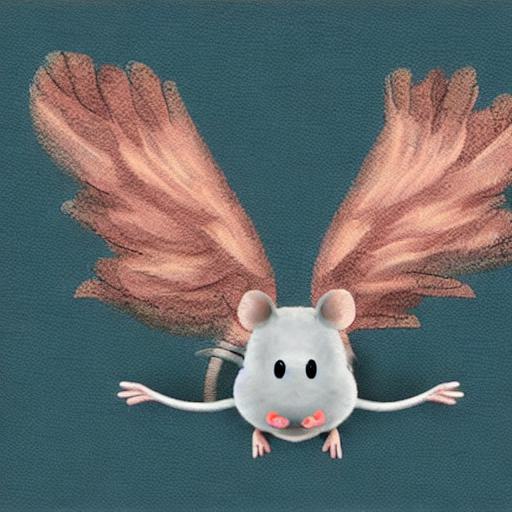} &
        { } &
        \includegraphics[width=0.12\textwidth]{images/our_results/fish/fish.jpg} &
        { } &
        \includegraphics[width=0.12\textwidth]{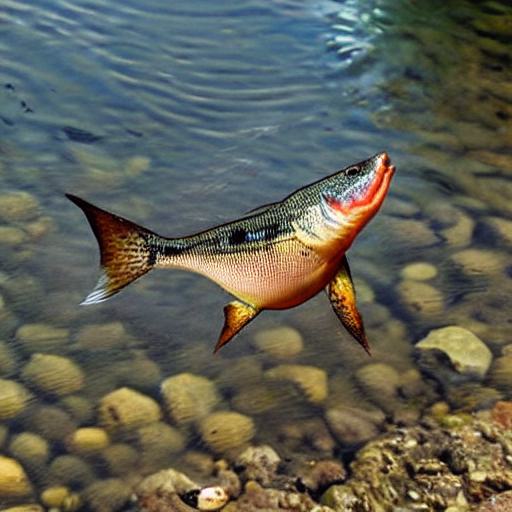} & 
        \includegraphics[width=0.12\textwidth]{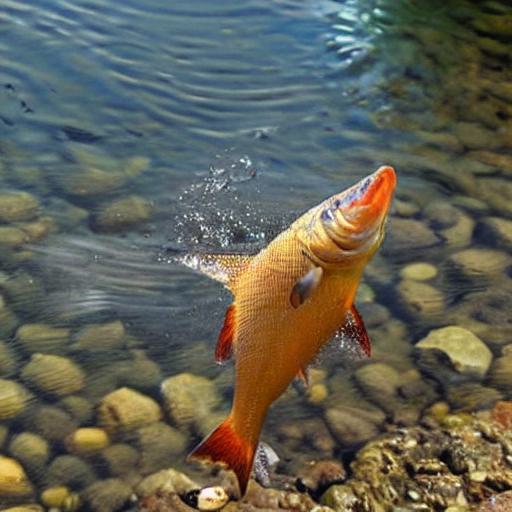} & 
        \includegraphics[width=0.12\textwidth]{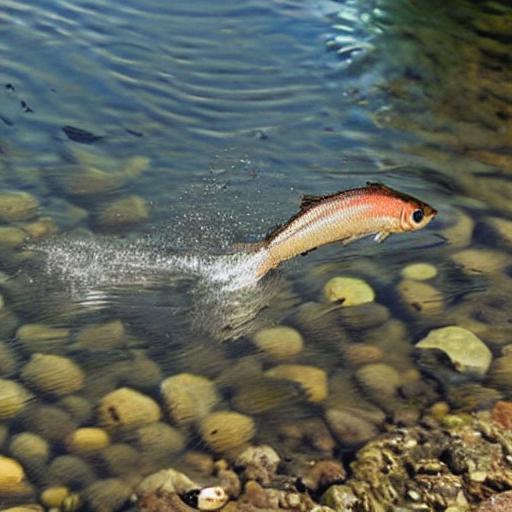} \\
        && \multicolumn{3}{c}{``A mouse with \emph{wings}''} & &
        && \multicolumn{3}{c}{``A \emph{fish} is jumping in a river''} \\
        \includegraphics[width=0.12\textwidth]{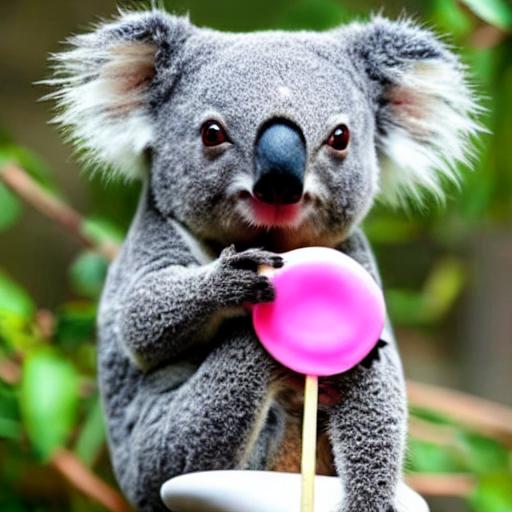} &
        { } &
        \includegraphics[width=0.12\textwidth]{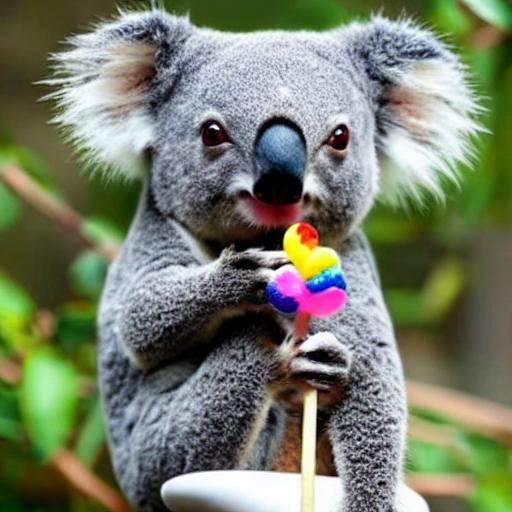} & 
        \includegraphics[width=0.12\textwidth]{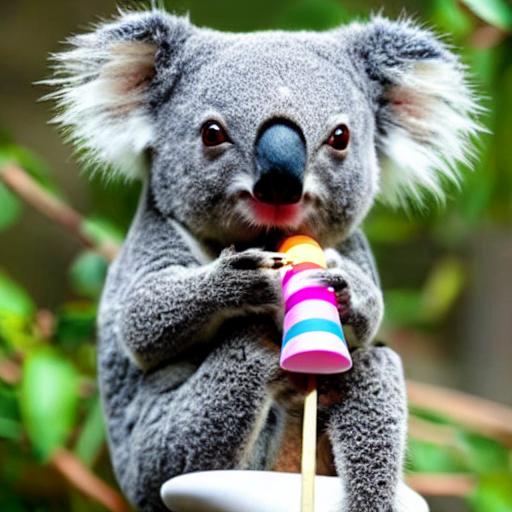} & 
        \includegraphics[width=0.12\textwidth]{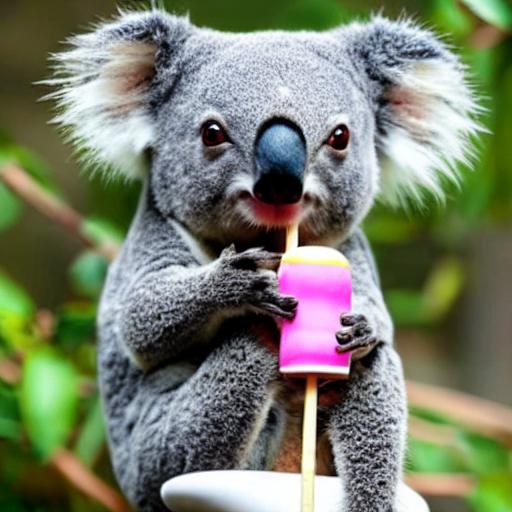} &
        { } &
        \includegraphics[width=0.12\textwidth]{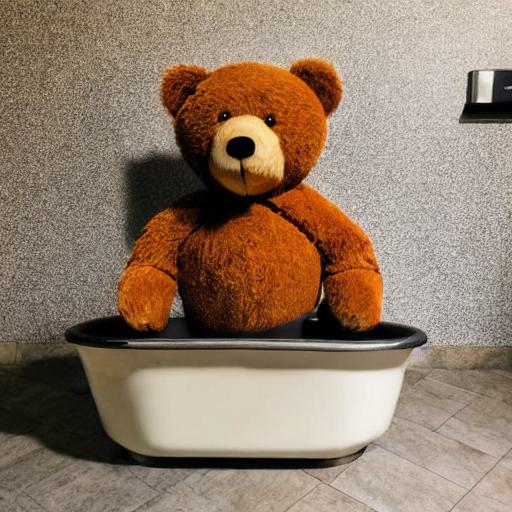} &
        { } &
        \includegraphics[width=0.12\textwidth]{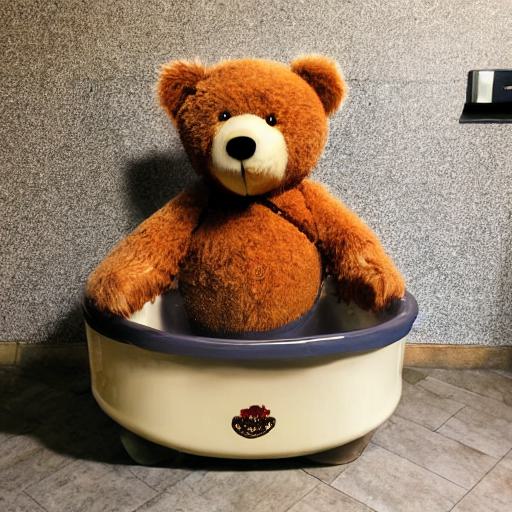} & 
        \includegraphics[width=0.12\textwidth]{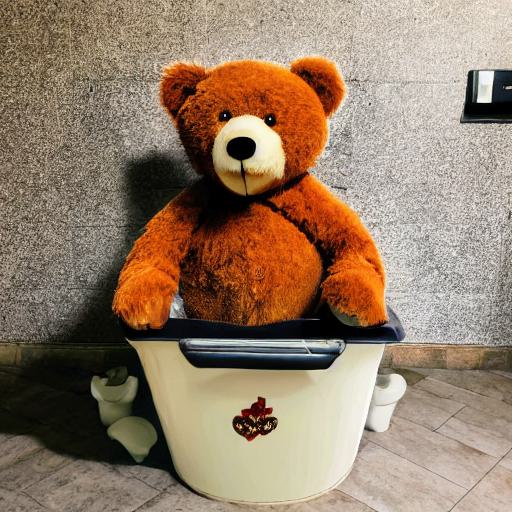} & 
        \includegraphics[width=0.12\textwidth]{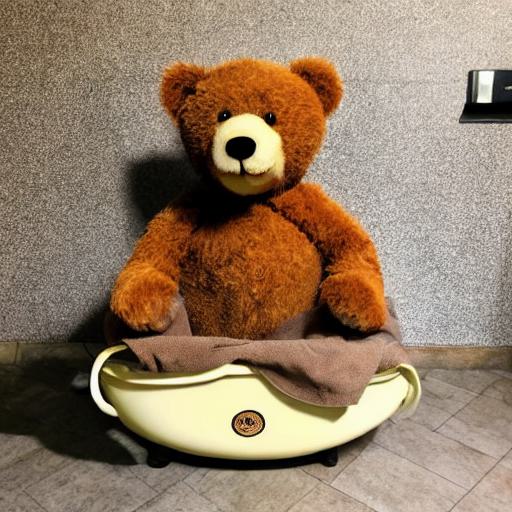} \\
        && \multicolumn{3}{c}{``A koala holding a \emph{lollipop}'' } & &
        && \multicolumn{3}{c}{``A teddybear in a \emph{tub}''} \\
        \includegraphics[width=0.12\textwidth]{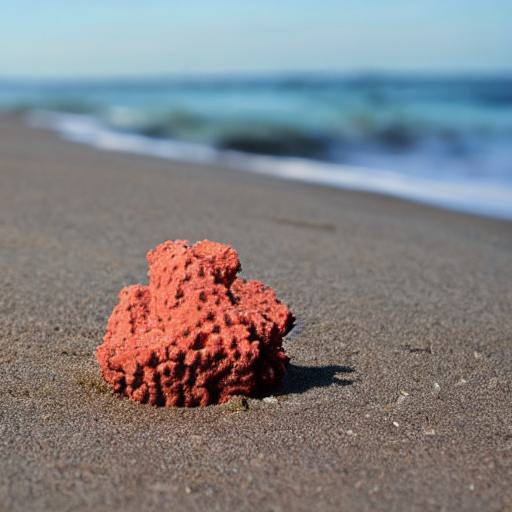} &
        { } &
        \includegraphics[width=0.12\textwidth]{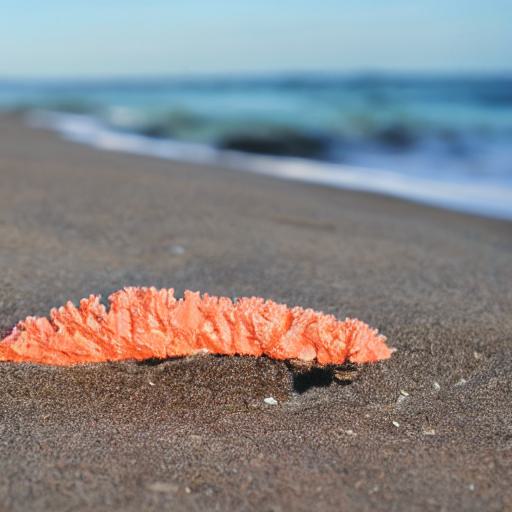} & 
        \includegraphics[width=0.12\textwidth]{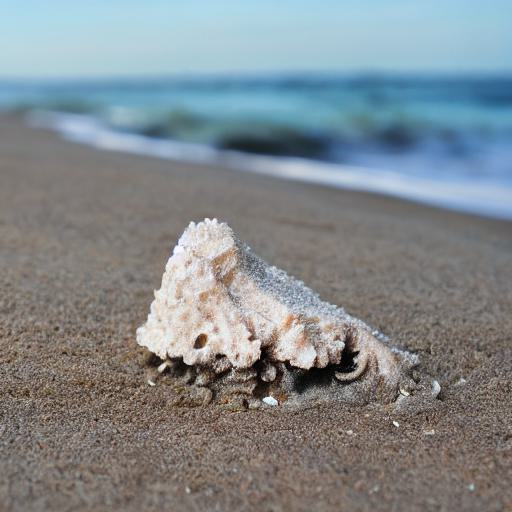} & 
        \includegraphics[width=0.12\textwidth]{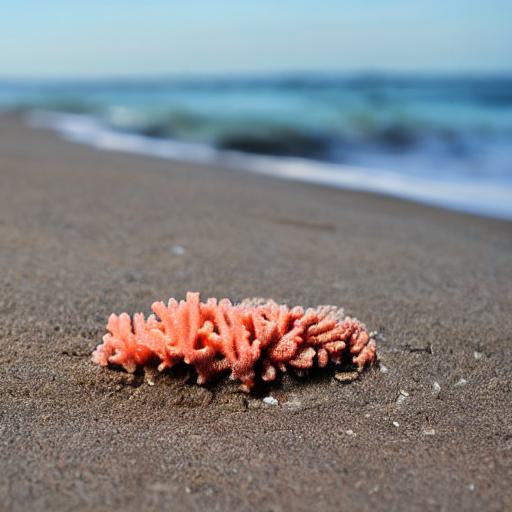} &
        { } &
        \includegraphics[width=0.12\textwidth]{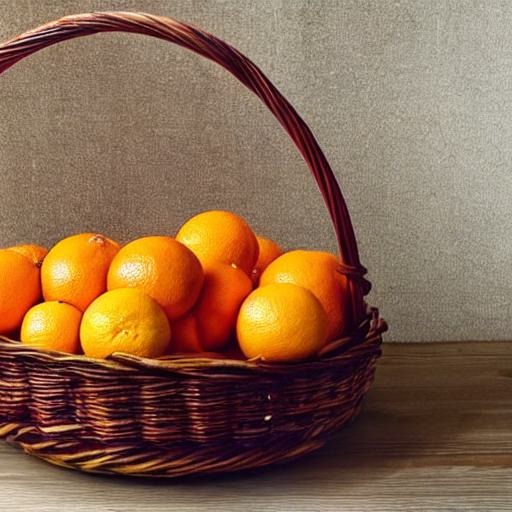} &
        { } &
        \includegraphics[width=0.12\textwidth]{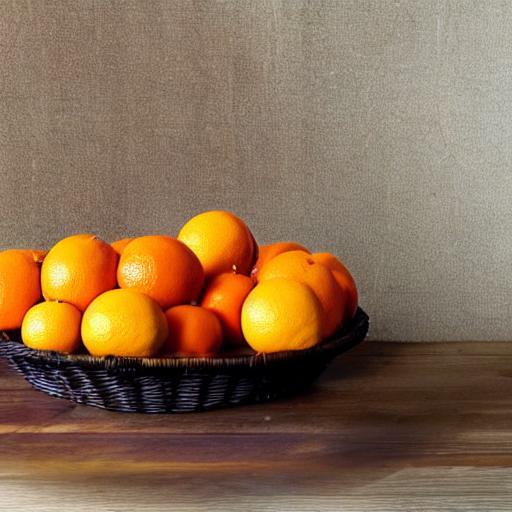} & 
        \includegraphics[width=0.12\textwidth]{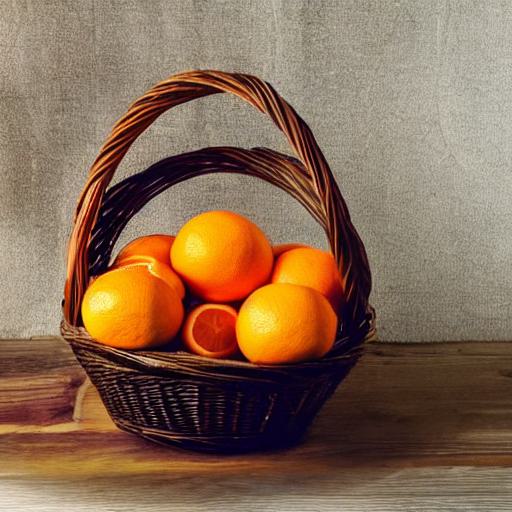} & 
        \includegraphics[width=0.12\textwidth]{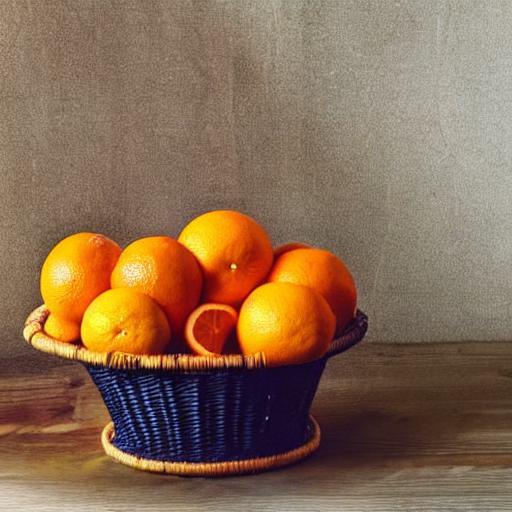} \\
        && \multicolumn{3}{c}{``A piece of \emph{coral} on the beach'' } & &
        && \multicolumn{3}{c}{``A \emph{basket} with oranges on a table''} \\
        \includegraphics[width=0.12\textwidth]{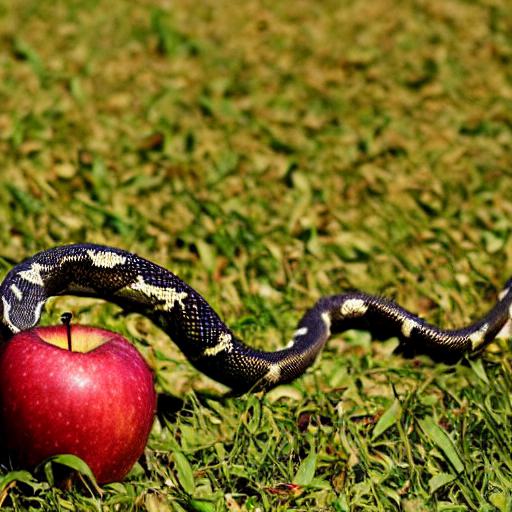} &
        { } &
        \includegraphics[width=0.12\textwidth]{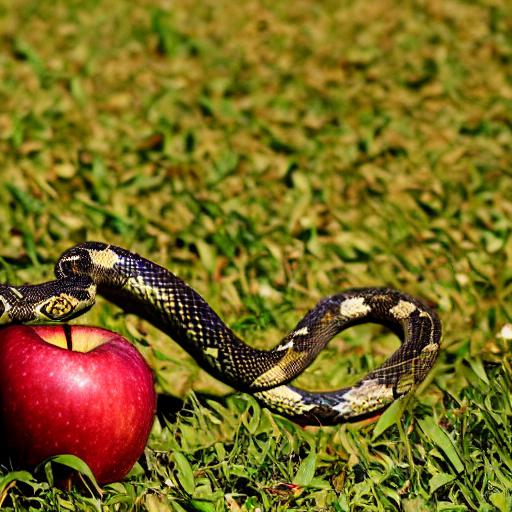} & 
        \includegraphics[width=0.12\textwidth]{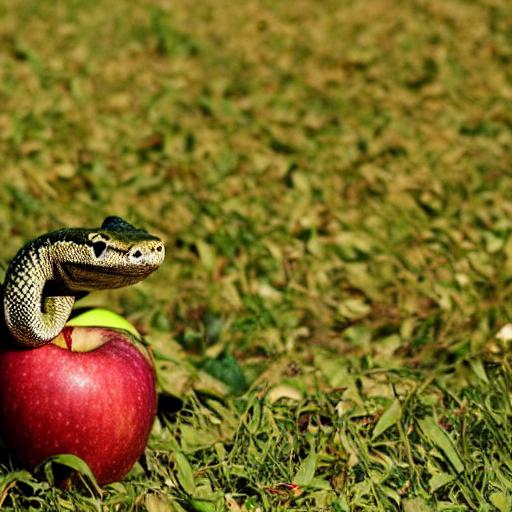} & 
        \includegraphics[width=0.12\textwidth]{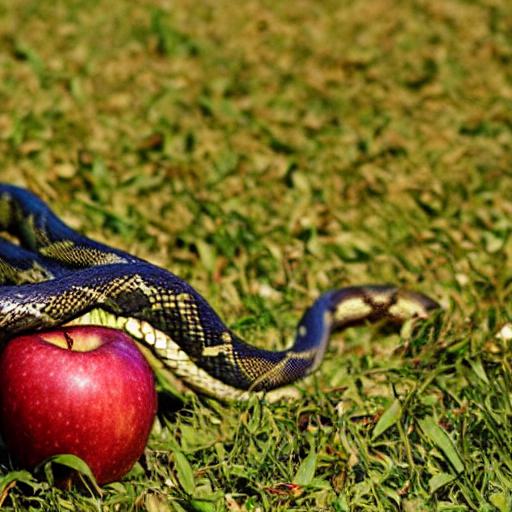} &
        { } &
        \includegraphics[width=0.12\textwidth]{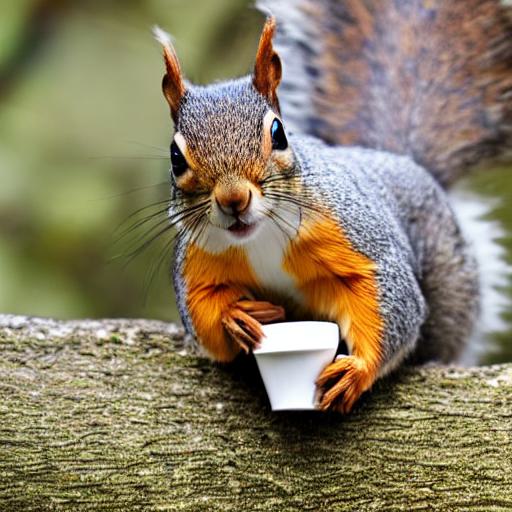} &
        { } &
        \includegraphics[width=0.12\textwidth]{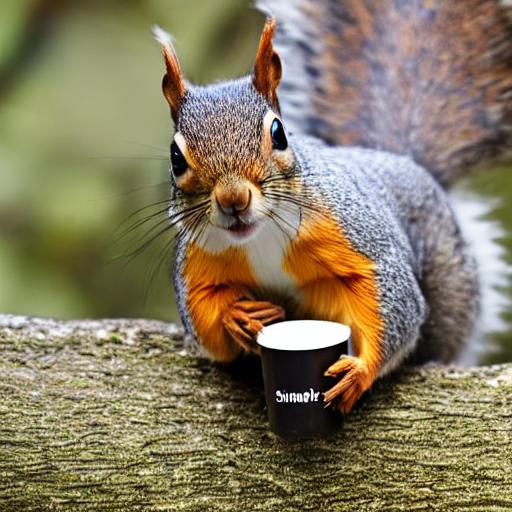} & 
        \includegraphics[width=0.12\textwidth]{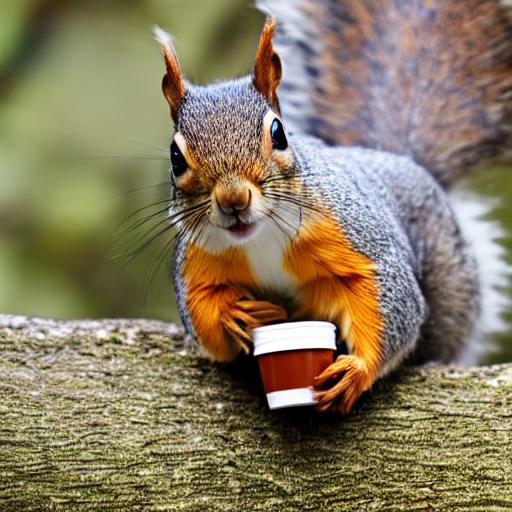} & 
        \includegraphics[width=0.12\textwidth]{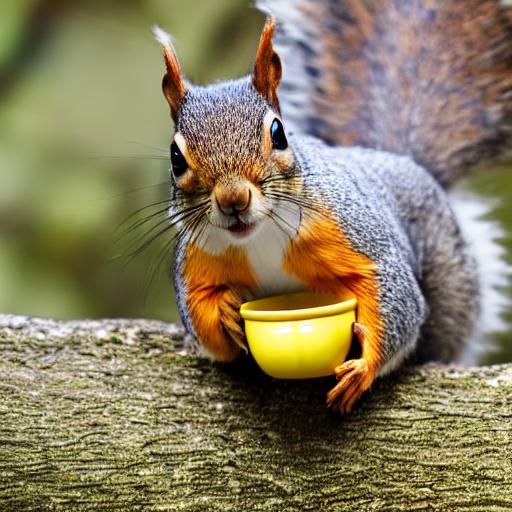} \\
        && \multicolumn{3}{c}{``A \emph{snake} in the field eats an apple''} & &
        && \multicolumn{3}{c}{``A squirrel holding a \emph{cup}''} \\
        \includegraphics[width=0.12\textwidth]{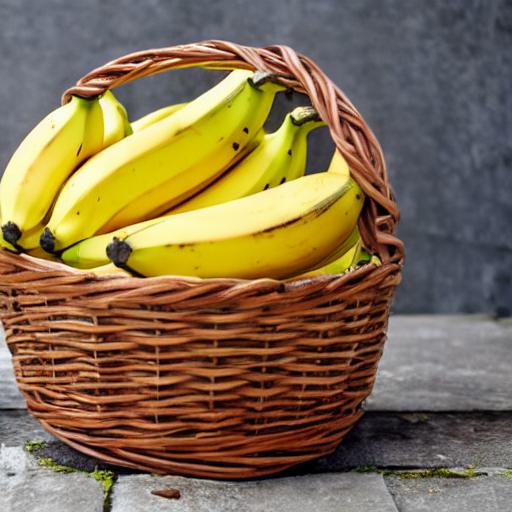} & { } &
			\includegraphics[width=0.12\textwidth]{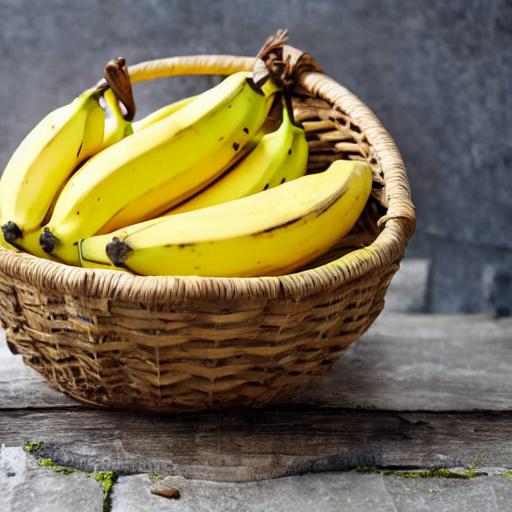} & 
			\includegraphics[width=0.12\textwidth]{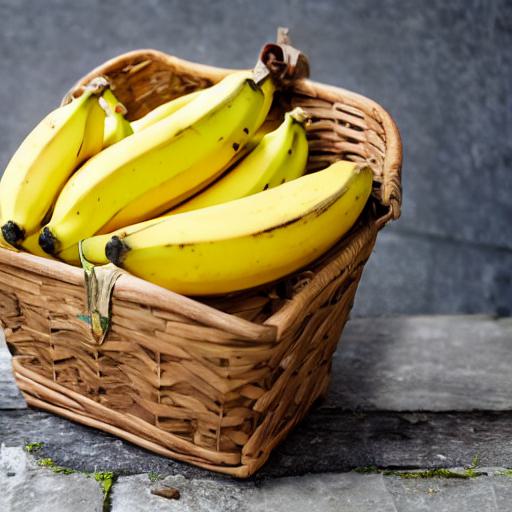} & 
			\includegraphics[width=0.12\textwidth]{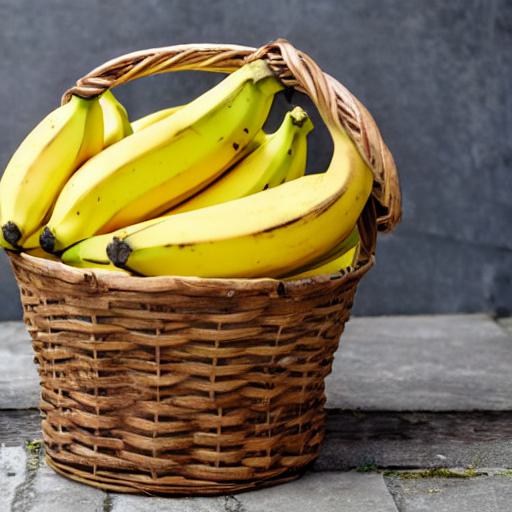} &
            { } &
   		\includegraphics[width=0.12\textwidth]{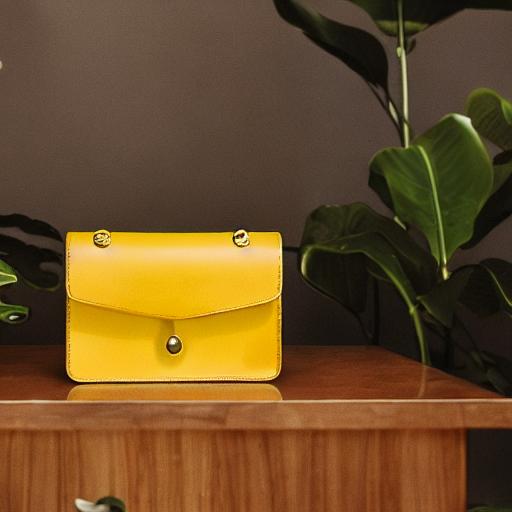} & { } &
			\includegraphics[width=0.12\textwidth]{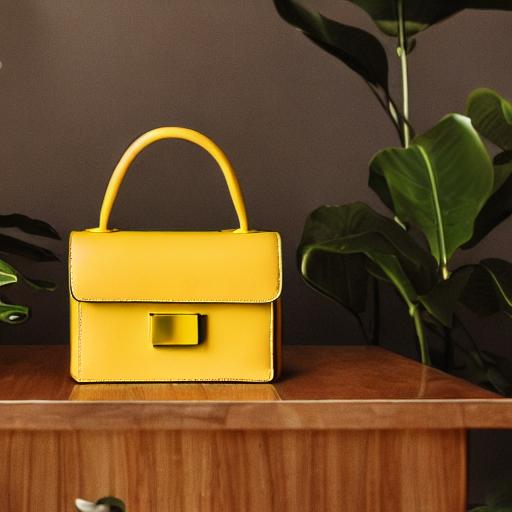} & 
			\includegraphics[width=0.12\textwidth]{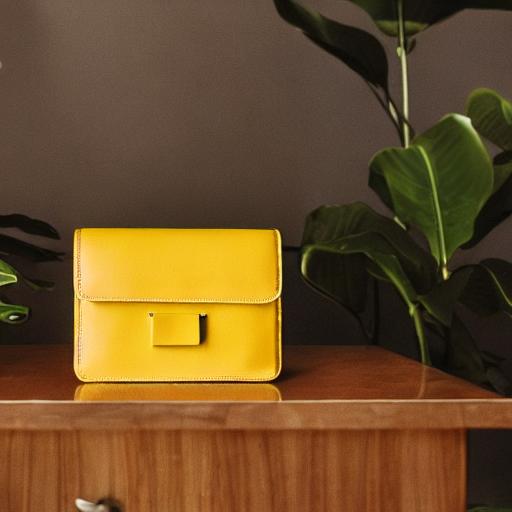} & 
			\includegraphics[width=0.12\textwidth]{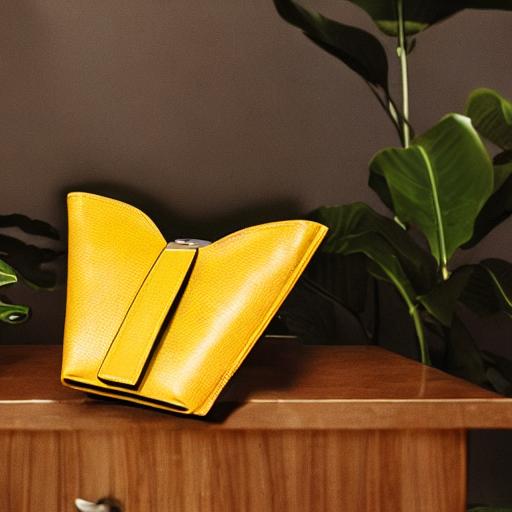}
			\\
           && \multicolumn{3}{c}{``A \emph{basket} with bananas''} & &
        && \multicolumn{3}{c}{``Luxury yellow \emph{purse} on a table''} \\
    \end{tabular}
    \vspace{5pt}
    \caption{Object-level variations for various scenes. For each scene, the leftmost image is the original sampled one. The \emph{emphasized} word corresponds to the explored object. As can be seen, our method generates different shape variations for each explored object.}
    \vspace{-7pt}
    \label{fig:our-results-supp}
\end{figure*}

In Figures~\ref{fig:our-extended-results-supp} and ~\ref{fig:our-results-supp}, we show additional results of our object shape variations pipeline.

\vspace{-8pt}

\end{appendices}

\end{document}